\documentclass[11pt]{article}

\usepackage[preprint]{acl}

\usepackage{times}
\usepackage{latexsym}

\usepackage[T1]{fontenc}
\usepackage[utf8]{inputenc}

\usepackage{microtype}

\usepackage{inconsolata}

\usepackage{hyperref}

\usepackage{etoc}

\usepackage{url}
\usepackage{graphicx}

\usepackage{inconsolata}
\usepackage{array}
\usepackage{pifont}
\usepackage{tabularx}
\usepackage{adjustbox}
\usepackage{multirow}
\usepackage{enumitem}
\usepackage{xspace}
\usepackage{tcolorbox}
\usepackage{booktabs,subcaption,amsfonts,dcolumn}
\usepackage{url}
\usepackage{amsmath,amsthm,amsfonts,amssymb,bm,stmaryrd, bbm}
\usepackage{color,xcolor,colortbl}
\usepackage{CJKutf8}
\usepackage{makecell}
\usepackage{booktabs}
\usepackage{graphicx}
\usepackage{listings}
\usepackage{float}
\usepackage{wrapfig}
\usepackage{rotating}
\usepackage{amssymb}
\usepackage{wasysym}

\newtoks\appBfloats \newtoks\appCfloats
\newcommand{\deferB}[1]{\global\appBfloats\expandafter{\the\appBfloats #1}}
\newcommand{\deferC}[1]{\global\appCfloats\expandafter{\the\appCfloats #1}}
\definecolor{warning}{HTML}{C80000}
\definecolor{mygreen}{rgb}{0.64, 0.76, 0.68}
\definecolor{myyellow}{rgb}{0.98, 0.94, 0.75}
\definecolor{mygreen}{rgb}{0.68, 0.85, 0.9}
\definecolor{myblue}{RGB}{33,95,154}
\definecolor{mypurple}{RGB}{224, 65, 245}
\definecolor{myorange}{RGB}{209, 136, 17}
\definecolor{Mycolor1}{HTML}{BAD8F2}
\definecolor{Mycolor2}{HTML}{DDEEFA}
\definecolor{carminepink}{rgb}{0.92, 0.3, 0.26}
\definecolor{jade}{rgb}{0.0, 0.66, 0.42}

\title{Tracing LLM Bias across Training: an Empirical Study on\\Bias Emergence, Training Configurations, and Mechanisms}

\author{
Omar Yahia\thanks{Equal contribution.} \quad Xuran Zhou\footnotemark[1] \quad Aaryan Gupta\footnotemark[1] \quad Julian McAuley \quad Xin Xu\thanks{Corresponding Author.}
\\ 
University of California, San Diego \\
\texttt{\{oyahia, xuranzhou, aag026, jmcauley, xinxucs\}@ucsd.edu}
}

\providecommand{\hfd}[1]{{\urlstyle{tt}\def\UrlBigBreaks{\do\/\do\-}\href{https://huggingface.co/datasets/#1}{\nolinkurl{#1}}}}
\providecommand{\hfl}[1]{{\urlstyle{tt}\def\UrlBigBreaks{\do\/\do\-}\href{https://huggingface.co/#1}{\nolinkurl{#1}}}}
\providecommand{\hfm}[2]{\href{https://huggingface.co/#1}{#2}}

\begin{document}
\maketitle
\begin{abstract}
Large language models (LLMs) have been widely reported to exhibit social biases. 
While substantial work has studied how to evaluate or mitigate bias after training, how different biased behaviors arise, evolve, and are mechanistically shaped during LLM training remains underexplored. 
In this work, we present a systematic empirical study on LLM biases across pre-training, mid-training, and post-training, examining bias emergence, the effects of training configurations, and underlying mechanisms.
Our experiments cover over 780 model checkpoints, including 10 random training seeds and 25 pre-training data recipes, across multiple model families and bias types.
We also apply cross-model activation patching to explore how post-training reshapes the mechanistic representations of LLM biases established during pre-training.
The extensive results and analysis show that LLM biases emerge very early during pre-training, while the overall bias remains stable through mid-training and post-training, with only some bias types shifting slightly. 
Post-training largely preserves the causal contribution patterns associated with LLM bias and slightly changes stereotypical preferences varying across bias types and models.
These findings suggest that effective debiasing requires interventions during pre-training, rather than relying on post-training alignment. 
We hope our findings can inspire more effective training-time debiasing methods for LLMs\footnote{Code and data: \url{https://github.com/notrandomath/bias-across-training}}.

\textcolor{red}{Warning: This paper explicitly contains the statement of stereotypes that may be offensive.}

\end{abstract}

\begin{figure}[htbp]
    \centering
    \includegraphics[width=\columnwidth]{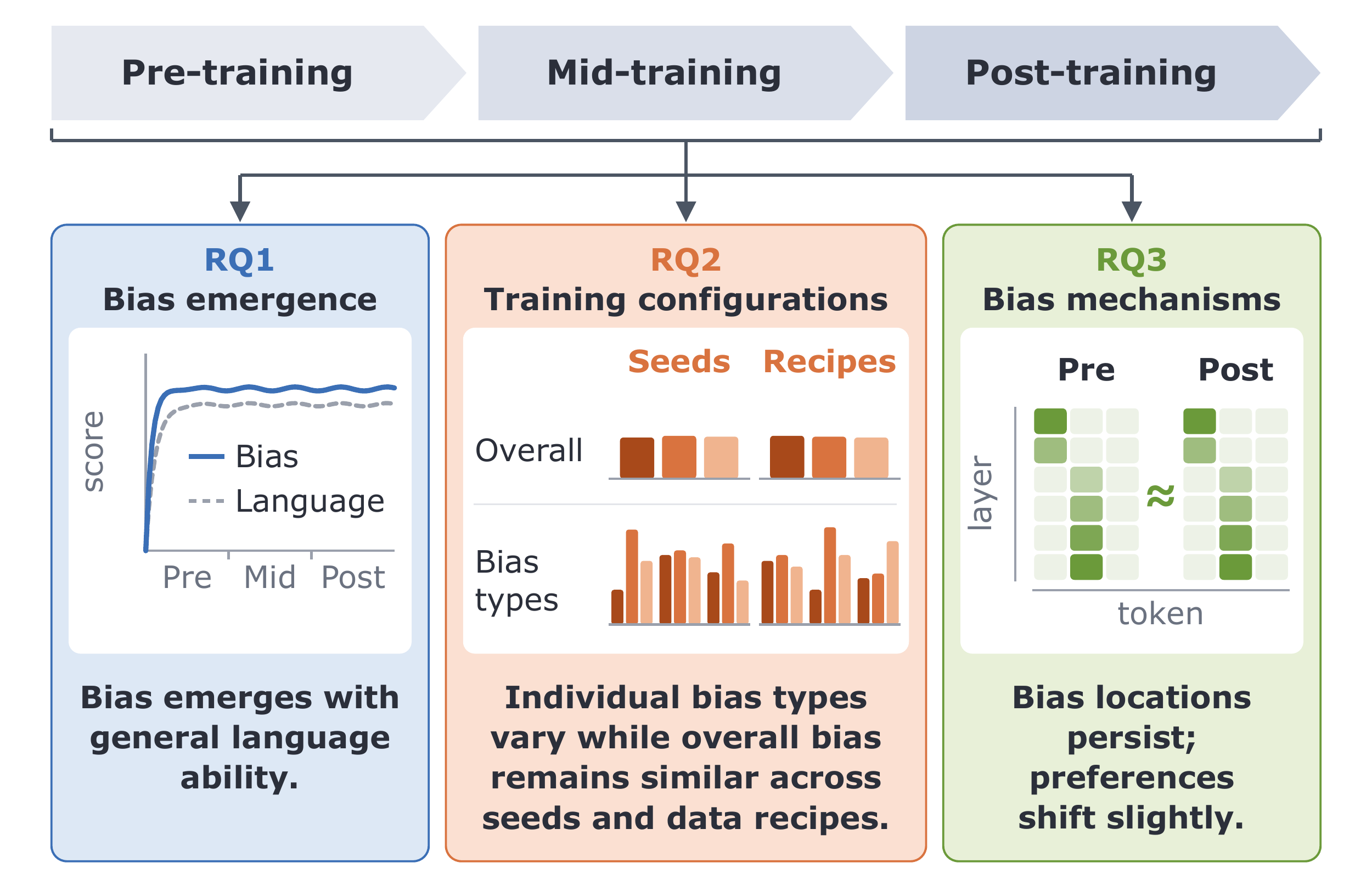}
    \caption{Three research questions explored in this work and main findings.}
    \label{fig:teaser}
\end{figure}

\section{Introduction}
Extensive prior work has shown that large language models (LLMs) exhibit socially biased behavior \citep{rajwal2025do,fan2025fairmtbench} and explored how to debias LLMs, often focusing on models after training is complete \citep{bias_edit, gallegos-etal-2025-self, xu2026biasfreebench}.
However, LLMs undergo multiple training stages, including pre-training, mid-training, and post-training, each involving different data and training objectives. 
Studying only the final models provides limited insight into how bias first emerges, how it evolves throughout training, and which training factors influence its development~\citep{pmlr-v306-biderman26a}, which is essential for identifying opportunities to mitigate bias during training and benefit LLM deployment.

A few recent studies have examined the origins of model biases during training \citep{swayed_by_finetuning,marcuzzi-etal-2026-quantization}. 
For instance, Pythia \citep{pythia} and PolyPythias \citep{polypythias} investigate how the gendered term frequency in pre-training data and random seeds used for initialization and training affect the gender bias of pretrained language models, respectively, but are limited to gender bias and pre-training.
\citet{swayed_by_finetuning} conducts a causal analysis of cognitive biases.
Nevertheless, no prior work investigates how training shapes the mechanistic representations that causally drive biased behaviors in LLMs (The related work is in Appendix \ref{app:related_work}).

Motivated by this gap, we conduct a comprehensive empirical study of how LLM bias emerges and evolves throughout training. 
Specifically, we investigate three research questions (RQ):
\begin{itemize}
    \item \textbf{RQ1}: How do LLM biases emerge during LLM training?
    \item \textbf{RQ2}: How do pre-training configurations affect learned bias?
    \item \textbf{RQ3}: How does training shape the mechanisms of LLM biases?
\end{itemize}
Figure~\ref{fig:teaser} provides an overview of our study. Our main contributions are summarized as follows:
\begin{itemize}
    \item \textbf{Comprehensive investigation of LLM bias during training.} We systematically study how LLM bias emerges, evolves, and is shaped across all training stages, including pre-training, mid-training, and post-training, covering training dynamics, pre-training configurations, and internal causal mechanisms.
    \item \textbf{Large-scale experiments}. We evaluate over 780 training checkpoints spanning pre-training, mid-training, and post-training, including 10 random seeds and 25 pre-training data recipes. Our experiments cover multiple model families and sizes, 11 bias types, different bias benchmarks, together with three general language capability benchmarks. We further conduct corpus-level association analysis and causal tracing to investigate LLM biases from the perspectives of both LLM behavior and mechanism.
    \item \textbf{Key empirical findings.} We find that LLM bias emerges early in pre-training alongside general language capability acquisition and remains relatively stable thereafter. Training configurations introduce varying effects across bias types, while post-training preserves the causal contribution patterns associated with LLM bias and slightly changes stereotypical preferences, which provides insights into how LLM training shapes bias and future training-time bias mitigation research.
\end{itemize}

\begin{figure*}[htbp] % 't' attempts to place it at the top of the page
     \centering
    \includegraphics[width=0.95\textwidth]{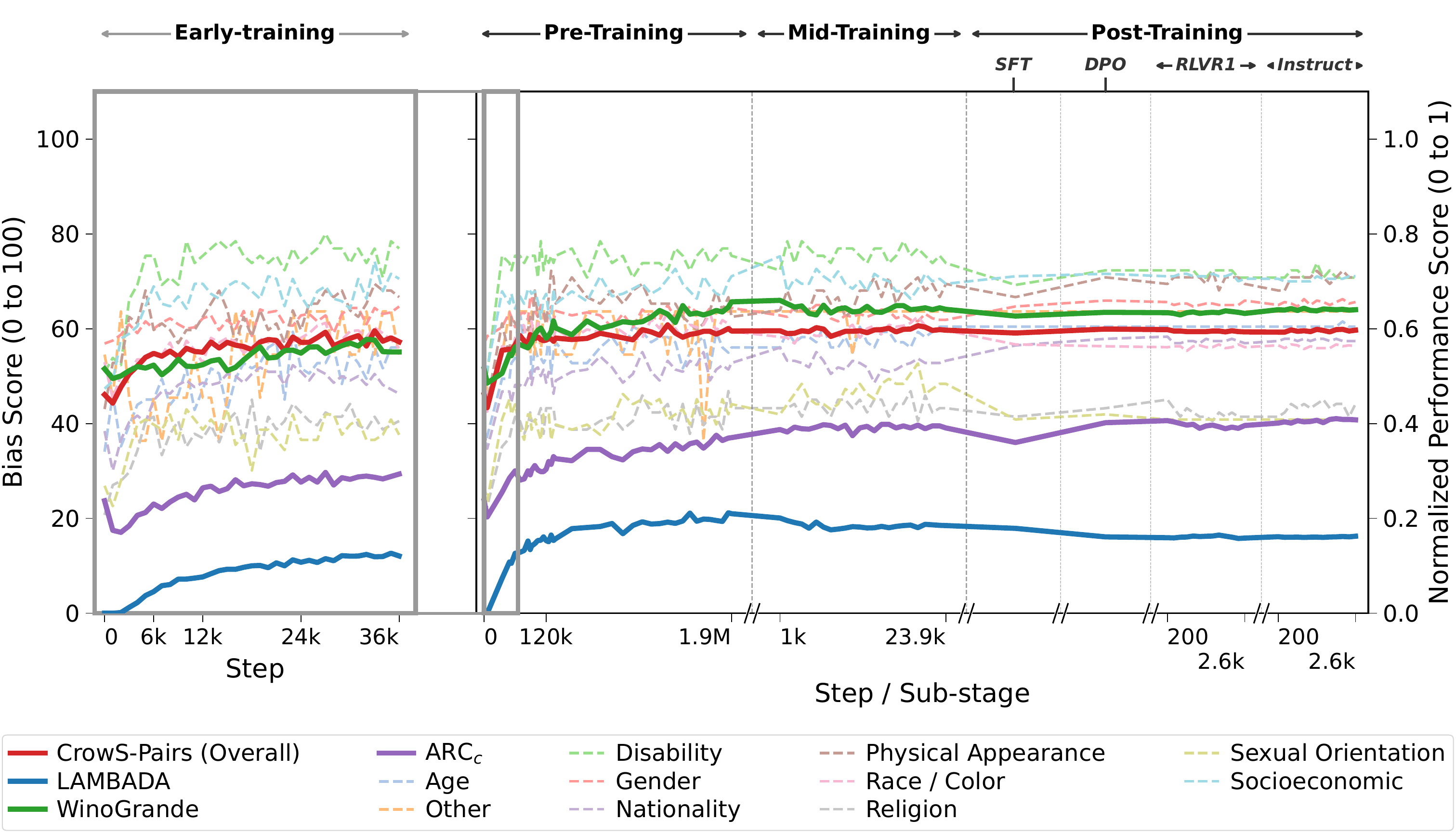}

     \caption{Evolution of bias score ($>$0.5 stereotyped, $<$0.5 anti-stereotyped) and performance scores ($\uparrow$) across OLMo 2 1B training. Task performance ($\uparrow$) on LAMBADA is reported as inverse perplexity (1/PPL) ranging from 0 to 1. Steps are shown independently for each sub-stage. The right panel: the main run. The left panel: the first 36k steps of the early-training run described in \S \ref{sec:pretraining_setup}. Metric formulations are described in Appendix~\ref{app:RQ1_benchmark}.
     95\% confidence intervals are in Figure \ref{fig:olmo1b_ci}. Detailed scores are in Table~\ref{tab:steps_olmo_2_1b_crows_pairs_pre_training_bias}.
     }
     \label{fig:OLMo_1b_line}
\end{figure*}

\section{How Do LLM Biases Emerge During LLM Training?}
\label{sec:rq1}

\subsection{Experimental Setups}
\label{sec:pretraining_setup}

\paragraph{Models.} 
We use OLMo 2 \citep{OLMo2} and Pythia \citep{pythia} with open-sourced intermediate training checkpoints. 
Because there are many checkpoints, we select a representative subset. 
Based on our early findings that bias acquisition concentrates in earlier pre-training (\S \ref{sec:rq1_analysis}), we select a subset that includes the first 20 checkpoints, followed by periodic sampling at 5\% intervals from all training checkpoints at each substage. 
% For instance, this procedure yields 40 checkpoints during pre-training, 24 checkpoints for mid-training, 1 checkpoint for SFT, 1 checkpoint for DPO, 13 checkpoints for RLVR1, and 13 checkpoints for RLVR2 for a total of 92 checkpoints.
We explore $i)$ 92, 61, and 122 checkpoints from OLMo 2 1B, 7B, and 13B, respectively, which cover pre-training, mid-training, supervised fine-tuning (SFT), direct preference optimization (DPO), and reinforcement learning with verifiable rewards (RLVR); $ii)$ 40 pre-training checkpoints from Pythia 1B.
More details are in Appendix \ref{app:RQ1_model}.
Because there are only 6 checkpoints provided before 30,000 steps for the main OLMo 2 1B run, we also evaluate its additional early-training checkpoints officially released with the same initialization, which provides checkpoints every 1k steps up to step 36k\footnote{\url{https://huggingface.co/allenai/OLMo-2-0425-1B-early-training}} to examine the emergence of LLM bias at a finer granularity (Figure~\ref{fig:OLMo_1b_line}).

\paragraph{Benchmarks.}
We adopt two commonly used bias benchmarks, including revised English CrowS-Pairs from \citet{crows-pairs-french} with 10 bias types and StereoSet \citep{stereoset} with 4 bias types \citep{meade-etal-2022-empirical, li-etal-2025-fairsteer}.
In each sample, there is a stereotyped context $x_\text{stereo}$ and a corresponding anti-stereotyped context $x_\text{anti}$ that share the same bias attribute words (the demographic subject, e.g., girl, scientist, Asian) but use different bias attribute terms that describe the biased meaning.
For example, there is a stereotyped context $x_\text{stereo}$, such as `\textit{Italy is known for being full of lazy people.}' with the bias attribute word $u$, `\textit{Italy}', and the bias attribute term $v_\text{stereo}$, `\textit{lazy}', and its anti-stereotyped context $x_\text{anti}$, such as `\textit{Italy is known for being full of diligent people.}' with the bias attribute term $v_\text{anti}$ `\textit{diligent}'.
We also consider generation-based bias evaluation with BBQ \cite{bbq}.
\begin{figure*}[ht!]
  \centering
  \begin{minipage}[t]{0.44\textwidth}
    \centering
    \includegraphics[width=\linewidth]{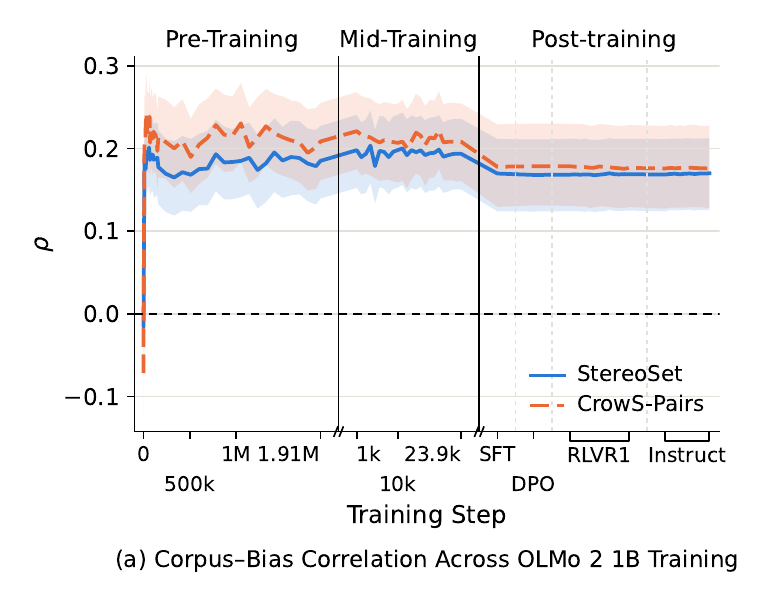}
  \end{minipage}\hfill
  \vspace{-2mm}
  \begin{minipage}[t]{0.43\textwidth}
    \centering
    \includegraphics[width=\linewidth]{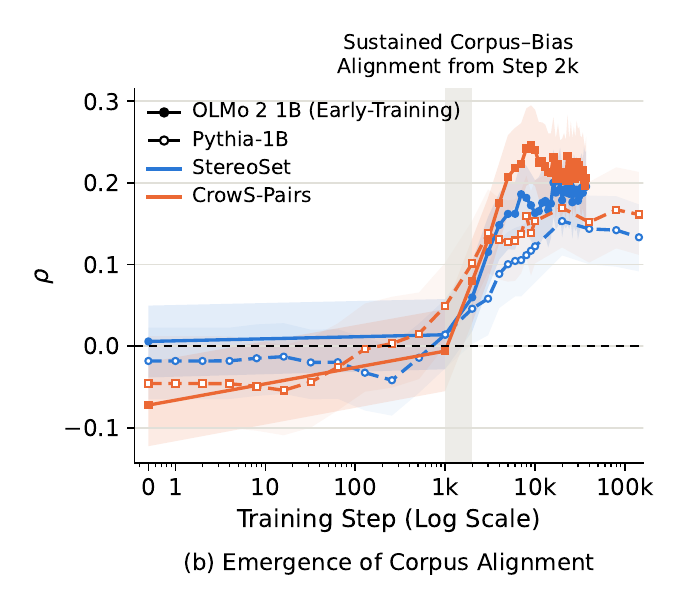}
  \end{minipage}

  \vspace{-1mm}

  \begin{minipage}[t]{0.43\textwidth}
    \centering
    \includegraphics[width=\linewidth]{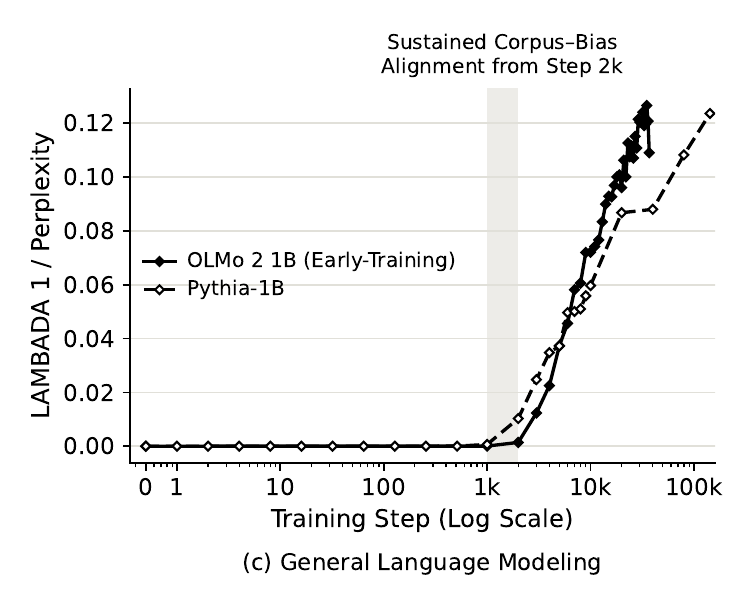}
  \end{minipage}\hfill
  \begin{minipage}[t]{0.46\textwidth}
    \centering
    \includegraphics[width=\linewidth]{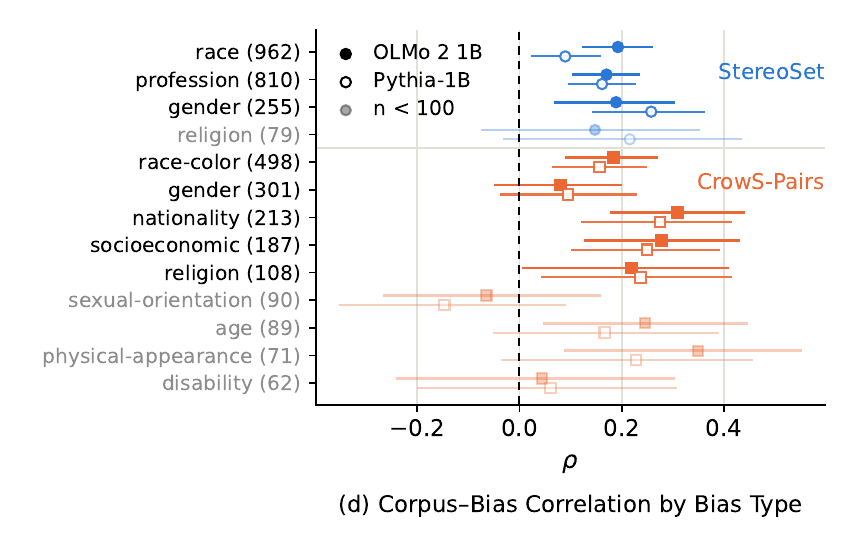}
  \end{minipage}
\vspace{-2mm}
  \caption{Corpus-bias co-occurrence analysis. Shading and error bars indicate 95\% confidence intervals
from 1,000 bootstrap resamples of the bias benchmarks. 
(b)(c) 38 OLMo 2 1B early-training checkpoints (every 1k step in 0 - 37k steps) and 25 Pythia 1B checkpoints (densely sampled early from 0 - 143k steps) on logarithmic step axes. (d) The parentheses indicate sample counts.
Light colors indicate the sample count for that bias type is less than 100.}
  \label{fig:cooccurrence}
\end{figure*}
However, early-training checkpoints, particularly before mid-training, often fail to generate valid answers in either option-letter or answer-text form, even with 5-shot prompting following HELM \citep{helm} and OLMo 2 evaluation (Appendix \ref{app:RQ1_benchmark}).
Following the technical reports of Pythia \citep{pythia} and OLMo 2 \citep{OLMo2}, general language benchmarks, including LAMBADA \citep{lambada}, WinoGrande \citep{winogrande} and ARC$_c$ \citep{arcc}, are used to explore bias emergence alongside general language capability acquisition.
More details are in Appendix \ref{app:RQ1_benchmark}.

\subsection{Results and Analysis}
\label{sec:rq1_analysis}

\textbf{LLM biases emerge at very early pre-training steps alongside general language capability acquisition and then plateau overall during mid-training and post-training, with only minor shifts in some bias types}.
As shown in Figure \ref{fig:OLMo_1b_line}, \ref{fig:cross_model_model_architectures_crows_pairs}--\ref{fig:cross_model_model_sizes_stereoset}, across all surveyed LLMs, bias begins to be acquired at very early stages of pre-training, around the point where improvements on general language benchmarks begin to plateau. On each bias benchmark, a higher bias score means that the model favors the stereotype more often (Appendix~\ref{app:RQ1_benchmark}). We define the acquisition of bias as the span of pre-training over which this score rises most sharply. For OLMo 2 1B, this happens within the first 130k steps, or around 6.82\% of the total pre-training process. This indicates that bias acquisition is closely coupled with the emergence of fundamental language capabilities.
The fine-grained results of checkpoints, including from early-training for OLMo 2, are shown in the left panel of Figure \ref{fig:OLMo_1b_line} and Table ~\ref{tab:steps_olmo_2_1b_crows_pairs_early_training_bias}--\ref{tab:steps_pythia_1b_crows_pairs_pre_training_general}.
We conjecture that stereotypical associations are themselves high-frequency co-occurrence patterns in corpora, and are therefore acquired through the same distributional learning signal that underlies general language modeling capability \citep{doi:10.1126/science.aal4230,feng-etal-2023-pretraining}. 
Once the model has internalized the corpus-level co-occurrence statistics sufficient to support fluent language modeling, the demographic asymmetries embedded in those same statistics are necessarily encoded as well.

To further validate our explanation, we examine whether model bias correlates with stereotypical word co-occurrence in each model’s own pre-training corpus.
We examine 92 OLMo 2 1B checkpoints spanning pre-training, mid-training, and post-training, supplemented by 38 checkpoints from a separate early-training run with the same initialization and 25 Pythia 1B checkpoints. 
Corpus statistics are estimated from a random sample of 280 DCLM shards from OLMo-mix-1124\footnote{\url{https://huggingface.co/datasets/allenai/olmo-mix-1124}} for OLMo and the first 10K training steps of the Pile \citep{Pile} ($\approx$21B tokens) for Pythia. 
For each sample from StereoSet and the revised English CrowS-Pairs, we quantify bias-related word associations using a pointwise mutual information (PMI) \citep{PMI}.
Higher PMI values indicate that the two words co-occur more often than expected if their occurrences were independent.
We contrast the PMI-based associations corresponding to the stereotypical and anti-stereotypical sentences $\Delta_{\text{PMI}}$.
% $\Delta_{\text{PMI}} = \text{PMI}(u, v_\text{stereo}) - \text{PMI}(u, v_\text{anti})$.
For each checkpoint, we compute the partial Spearman correlation $\rho$ \citep{rho} between $\Delta_{\text{PMI}}$ 
and the bias score, controlling for word-frequency covariates and differences in sentence length, which are described in Appendix \ref{app:cooccurrence}.
We additionally track LAMBADA perplexity to compare the development of this correlation with language modeling performance, and repeat the correlation analysis separately for each bias type using the released base models.
Figure \ref{fig:cooccurrence}(a) tracks $\rho$ across OLMo 2 1B training stages.
Figure \ref{fig:cooccurrence}(b) examines $\rho$ across early-training of OLMo 2 1B and pre-training of Pythia 1B.
Figure \ref{fig:cooccurrence}(c) tracks LAMBADA perplexity at the same checkpoints. 
Figure \ref{fig:cooccurrence}(d) reports the partial Spearman correlation $\rho$ by bias type for both released base models.

\begin{figure*}[htbp]
\centering
\includegraphics[width=0.9\textwidth]{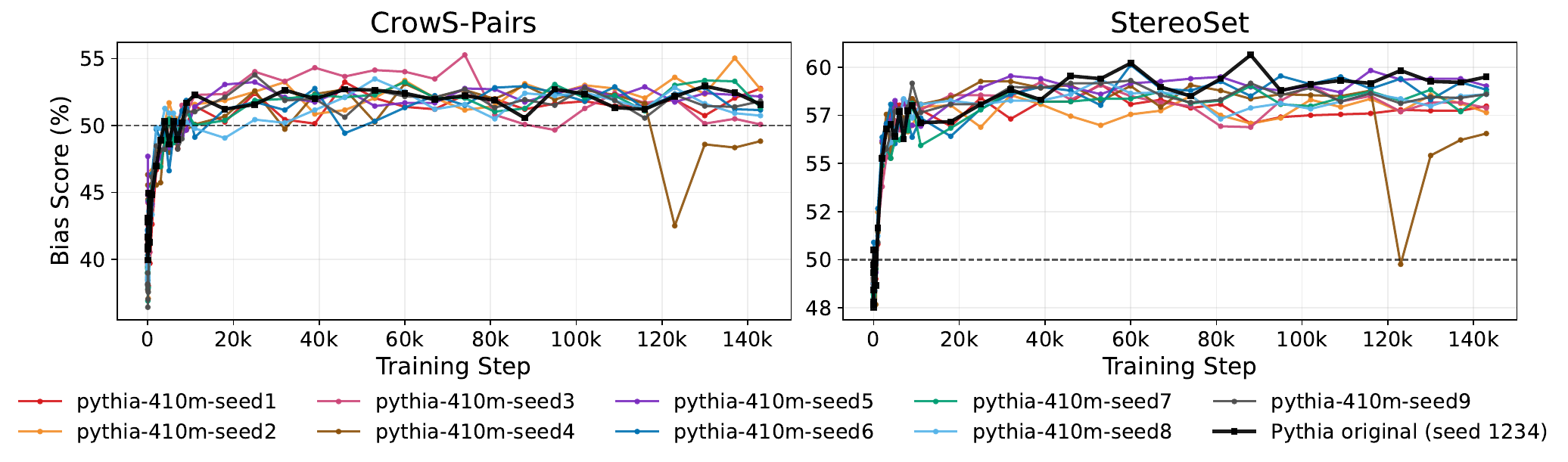}
\caption{
Bias scores during pre-training of the PolyPythias 410M seeds and the standard Pythia 410M (seed 1234), with final-checkpoint 95\% confidence intervals for the cross-seed mean scores in Table~\ref{tab:pythia_seed_bias_type_statistics}.
}
\label{fig:pretraining_config_seeds}
\end{figure*}

Figure \ref{fig:cooccurrence}(a) shows that $\rho$ rises sharply early in OLMo 2 1B pre-training and remains relatively stable through the rest of pre-training and mid-training, and the samples with stronger corpus associations favoring stereotypical wording also tend to receive stronger stereotypical preferences from the LLM.
These results align with our suggestion that model biases emerge early alongside a rapid increase in their association with corpus bias co-occurrence, while subsequent post-training largely preserves this association.
Figure \ref{fig:cooccurrence}(b) further locates the emergence of this relationship.
The positive correlations are absent at initialization but become consistently significant from step 2k onward on both benchmarks and LLMs. 
Figure \ref{fig:cooccurrence}(c) shows that LAMBADA inverse perplexity improves during the same early-training period in which corpus–bias correlations increase. 
Their parallel development is consistent with our suggestion that stereotypical associations are acquired as part of learning the corpus co-occurrence patterns underlying language modeling ability.
Figure \ref{fig:cooccurrence}(d) shows positive correlations whose 95\% confidence intervals exclude zero for the most bias types in both released base models.
This consistency across bias types and model families suggests that no single bias type drives the overall positive correlation.

\section{How Do Pre-training Configurations Affect Learned Bias?}
\label{sec:pretraining_config}
Having observed that LLM biases emerge early in pre-training and plateau throughout subsequent training stages, we next investigate how pre-training configurations influence bias acquisition.

\subsection{Experimental Setups}
\label{sec:pretraining_config_setup}

\paragraph{Pre-training random seeds.}
To investigate how training seeds affect LLM bias acquisition, we use Pythia 410M models from PolyPythias \citep{polypythias}, trained with different random seeds while keeping all other training configurations fixed. 
Specifically, we consider 10 seeds in $\{1,2,\dots,9, 1234\}$, where seed 1234 is for the standard Pythia 410M\footnote{\url{https://huggingface.co/EleutherAI/pythia-410m}}, and track LLM biases of revised CrowS-Pairs and StereoSet across 40 pre-training checkpoints (same in \S \ref{sec:pretraining_setup}) for each model from \texttt{step0} to the final \texttt{step143000}.
Their bias scores across pre-training are shown in Figure \ref{fig:pretraining_config_seeds}.

\paragraph{Pre-training data recipes.}
To investigate the effect of pre-training data on bias types, we use DataDecide \citep{datadecide}, which varies data recipes while keeping other training configurations fixed.
Specifically, we use 25 1B DataDecide models, whose recipes cover different source corpora, Dolma1.7 source ablations, quality filtering, and mixtures of the DCLM-Baseline \citep{dclm} and Dolma1.7 \citep{soldaini-etal-2024-dolma} corpora, as listed in Table~\ref{tab:datadecide_data_mixture_models}.
We compare the bias score at the final-checkpoints (\texttt{step69369-seed-default}) for each bias type to analyze how pre-training data recipes affect biased behavior.
For readability, we arrange the 25 recipes into six comparison groups: cross-corpus baselines, Dolma1.7 source ablations, Falcon \citep{DBLP:conf/nips/PenedoMHCACPAL23} and Falcon+CC quality filtering, DCLM classifier quality filtering, DCLM FineWeb-Edu classifier quality filtering, and mixtures of DCLM-Baseline and Dolma1.7.

\subsection{Results and Analysis}
\begin{figure}[ht!]
\centering
\includegraphics[width=0.8\columnwidth]{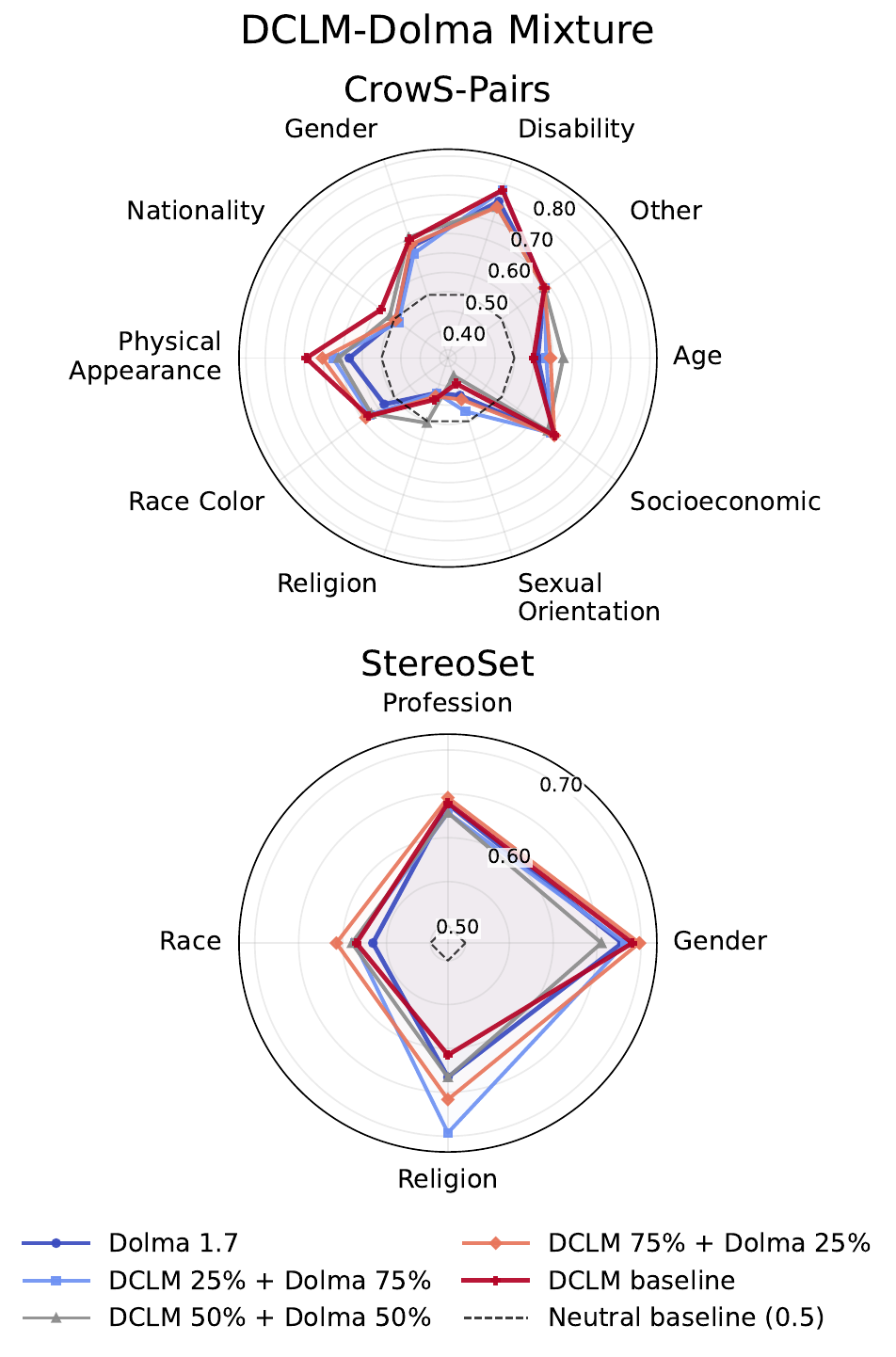}
\caption{
Bias scores by bias types for the DataDecide mixtures of DCLM-Baseline and Dolma1.7, with benchmark-level 95\% confidence intervals in Table~\ref{tab:datadecide_score_ci}.
}
\label{fig:pretraining_config_mixtures}
\end{figure}

\begin{figure}[ht]
\centering
\includegraphics[width=0.8\columnwidth]{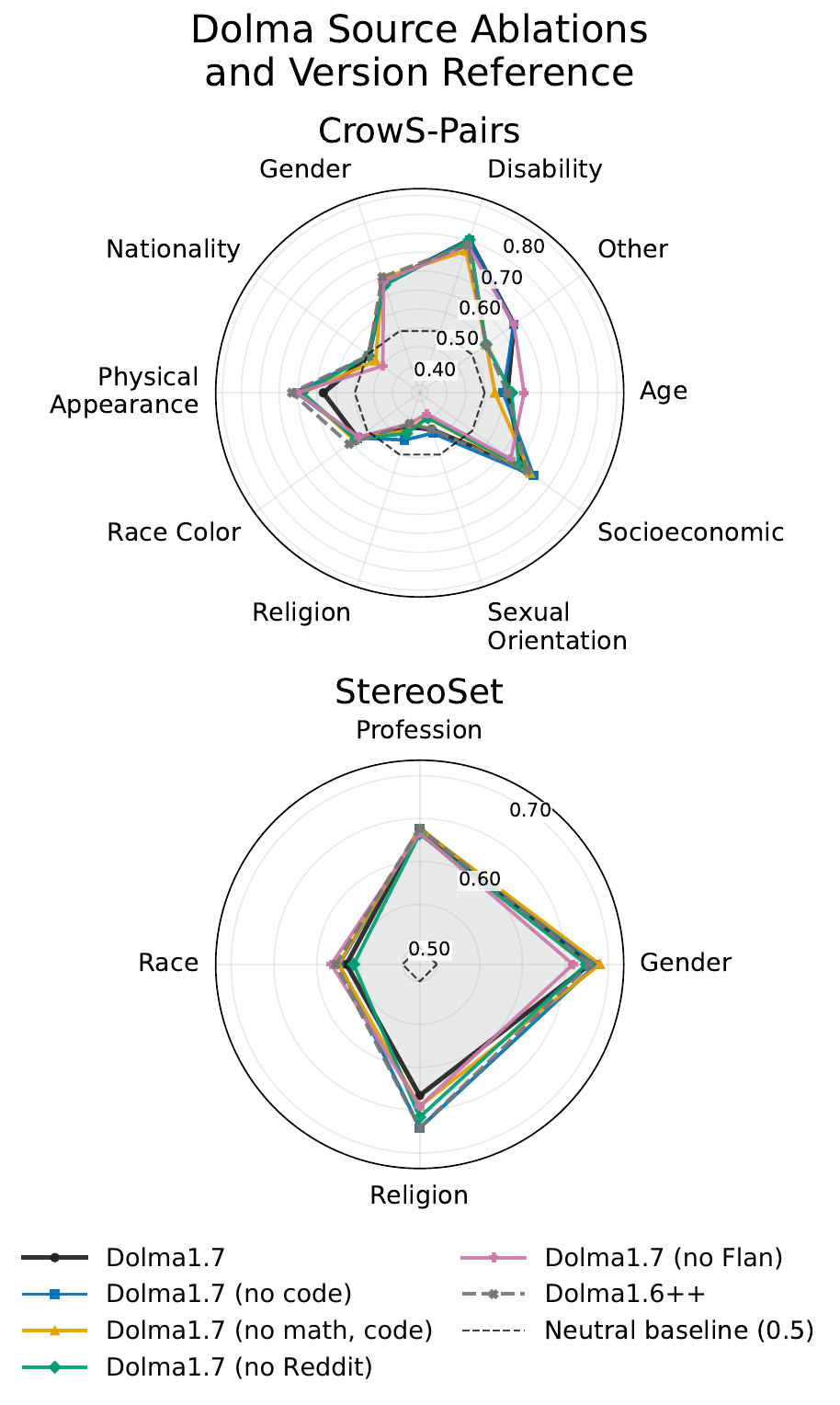}
\caption{
Bias scores by bias types for the DataDecide Dolma1.7 source ablations, with benchmark-level 95\% confidence intervals in Table~\ref{tab:datadecide_score_ci}. The dashed line at 0.5 corresponds to a bias score of 50\%.
}
\label{fig:pretraining_config_dolma_ablations}
\end{figure}

Figures \ref{fig:pretraining_config_mixtures} and \ref{fig:pretraining_config_dolma_ablations} show the results of the mixture ablation and Dolma1.7 source ablation, and Appendix \ref{app:more_datadecide_results} shows the results of the other four groups.
Additional details are provided in Appendix \ref{app:rq2}. Based on the LLMs and benchmarks considered in this work, we provide analysis on training seeds and data recipes as follow.

% \subsubsection{Effects of Random Seeds}
\paragraph{Different training seeds exhibit similar bias evolution patterns throughout pre-training but lead to substantial variation in certain biases.}
As shown in Figure \ref{fig:pretraining_config_seeds}, models trained with different random seeds exhibit similar bias evolution patterns, with bias scores increasing rapidly during early pre-training and remaining relatively stable thereafter. At the final checkpoint, the overall bias scores show relatively small variations across seeds, with standard deviations of 1.21 and 0.84 percentage points on CrowS-Pairs and StereoSet, respectively (Table \ref{tab:pythia_seed_bias_type_statistics}). However, certain bias types exhibit substantially larger variations. 
For example, bias scores for physical appearance on CrowS-Pairs vary by 19.44 percentage points across seeds, compared to only 3.94 percentage points for the overall benchmark. This suggests that aggregate bias scores can mask larger differences within individual bias types.
Moreover, the seeds yielding scores closest to 50\% differ across benchmarks (Table \ref{tab:pythia_seed_trajectory_scores}), and their relative rankings fluctuate throughout training (Figure \ref{fig:pretraining_config_seeds}). 
This suggests that the effects of training seeds on LLM bias are uncontrollable, making seed selection unlikely to be a reliable bias mitigation strategy.

\paragraph{Different pre-training data recipes largely preserve similar bias profiles, but their effects vary across bias types and corpora.}
Figure~\ref{fig:pretraining_config_dolma_ablations} presents the bias scores for the Dolma1.7 source ablations.
Removing Reddit brings the gender scores closer to 50\% on both benchmarks but changes race scores in opposite directions on CrowS-Pairs and StereoSet.
Similarly, removing FLAN \citep{DBLP:conf/icml/LongpreHVWCTZLZ23} reduces the overall CrowS-Pairs score while increasing the physical appearance score.
All source ablations increase physical appearance scores on CrowS-Pairs and religion scores on StereoSet.
Data mixing also produces non-monotonic changes (Figure~\ref{fig:pretraining_config_mixtures}).
StereoSet scores do not consistently rise or fall as the proportion of DCLM-Baseline increases.
The mixture with 75\% DCLM-Baseline scores 63.49\%, higher than the models trained only on Dolma1.7 (60.97\%) or only on DCLM-Baseline (61.92\%), while the mixtures with 25\% and 50\% DCLM-Baseline score 61.82\% and 61.40\%, respectively, falling between these two models.
% These scores do not change monotonically with the mixture proportion, and all three mixtures have higher StereoSet race scores than either single-corpus model.
% For quality filtering, retaining the highest-scoring 10\% of documents under the reproduced DCLM quality classifier increases the CrowS-Pairs score by 3.70 percentage points for Falcon+CC but decreases it by 1.85 percentage points for DCLM-Baseline.
% StereoSet scores increase for both corpora, by 1.33 and 1.03 percentage points, respectively, as reported in Table~\ref{tab:datadecide_score_ci}.
Quality filtering exhibits corpus-dependent effects: retaining the top 10\% of documents ranked by the replicated DCLM classifier increases the CrowS-Pairs score by 3.70 percentage points for Falcon+CC but decreases it by 1.85 percentage points for DCLM-Baseline. StereoSet scores increase by 2.00 and 1.00 percentage points, respectively (Table \ref{tab:datadecide_score_ci}).
These results suggest that the effects of data composition and filtering on biased behavior depend on the underlying corpus and bias type.
% Lower overall scores can conceal stronger stereotype preferences within individual bias types.

\section{How Does Training Shape the Mechanisms of LLM Bias?}
\label{sec:rq3}

\subsection{Experimental Setup}
\label{sec:causal_tracing_setup}

\paragraph{Models and data.}To analyze the mechanistic representation of LLM bias across training, following
\citet{DBLP:conf/nips/VigGBQNSS20, bias_edit}, we conduct bias causal tracing on gender, race, and profession bias types of StereoSet intrasentence (\S \ref{sec:pretraining_setup}, Appendix~\ref{app:rq3_setup}).
We trace LLMs from 4 model families: OLMo~2 1B~\citep{OLMo2}, Qwen2.5 1.5B~\citep{qwen2.5report}, Llama~3.2 1B~\citep{llama32} and Pythia 1B~\citep{pythia}, each as a pre-trained model (Pre, released as base) and, except for Pythia, as a post-trained model (Post, released as Instruct). For OLMo~2, Pre is the last mid-training checkpoint, which the Instruct model is post-trained from, rather than the released base model (Appendix~\ref{app:rq3_setup}). This gives 21 pairs of model and bias type (4 Pre and 3 Post models, 3 bias types), and 9 Pre--Post comparisons (3 model families, 3 bias types). We also trace the intermediate training checkpoints of OLMo 2 and Pythia (Appendix~\ref{app:rq3_setup}). We call the tokens of the bias attribute word $u$ (\S\ref{sec:pretraining_setup}) the \textbf{subject}, the tokens of the bias attribute term $v_\text{stereo}$ or $v_\text{anti}$ the \textbf{target}, and the token right before the target the \textbf{pre-target} token. For example, in `\textit{Italy} is known for being full of \textit{lazy}/\textit{diligent} people.' ($x^\text{stereo}$/$x^\text{anti}$), `\textit{Italy}' is the subject, `\textit{of}' is the pre-target token, and `\textit{lazy}' or `\textit{diligent}' is the target.

\paragraph{Bias causal tracing protocol.} There are three sequential runs for $x^\text{stereo}$ and $x^\text{anti}$ in each sample: (1) a \textit{clean} run records the activations of every token position $t$ at each layer $\ell$ for each input; (2) a \textit{corrupted} run adds Gaussian noise to the input embedding of each subject to disrupt the cues that can trigger bias; (3) a \textit{patched} run restores the corrupted activation to the clean activation at each ($t$, $\ell$) position across Transformer blocks (\textit{single state}) or MLPs or attention layers over a window of layers (\textit{MLP window}, \textit{Attn window}; 5 layers for the 16-layer models and 7 for the 28-layer one). To investigate whether post-training shapes the mechanistic representation of LLM biases, we apply \textbf{cross-model patching} between each base model (Pre) and its Instruct one (Post)~\citep{cross_patching}. The clean activation from one LLM patches the corrupted activation of the corresponding LLM in both directions, i.e., Pre$\to$Post and Post$\to$Pre, in comparison to within-model patching, where a model is patched with its own clean activations.
All details are shown in Appendix~\ref{app:single_patching_additional_details}.

\paragraph{Quantifying Bias and Effects.}
Given an LLM with parameters $\theta$, we measure its preference between $x_i^\text{stereo}$ and $x_i^\text{anti}$ from each sample $i$ over $N$ samples for a bias type by the signed log-probability difference \textit{\textbf{LPD}}:
\begin{equation}
\label{eq:lpd}
    \textit{LPD} = \frac{1}{N}\sum_{i=1}^{N}[\log P_\theta(x_i^{\text{stereo}}) - \log P_\theta(x_i^{\text{anti}})]
\end{equation}
where $\log P_\theta(x)$ is the length-normalized log probability the model assigns
to $x$.
We compute \textit{LPD} in each of the three forward passes by averaging the per-sample values of Appendix Equations~\ref{eq:lpdclean}--\ref{eq:lpdrestored}. $\textit{LPD}^\text{clean}$ and $\textit{LPD}^\text{corrupted}$ are its values in the clean and corrupted passes. 
$\textit{LPD}^\text{restored}$ is the preference after restoring the clean activation at one token position and layer in the corrupted.
Corrupting the subjects across the dataset is used to remove the bias triggers, which is measured by \textit{\textbf{Effect Gap}},
$
\textit{Effect Gap} = \textit{LPD}^\text{clean} - \textit{LPD}^\text{corrupted}.
\label{eq:effect_gap}
$
To evaluate the causal effect of a patched activation on the model bias, we use the normalized indirect effect (\textbf{\textit{NIE}}).
\textit{NIE} quantifies how much restoring the clean activation recovers the model’s original preference between stereotypical and anti-stereotypical sentences after subject embeddings are corrupted.
At each token position and each layer, \textbf{NIE} is:
\begin{equation}
\textit{NIE} = \frac{\textit{LPD}^\text{restored} - \textit{LPD}^\text{corrupted}}{\textit{LPD}^\text{clean} - \textit{LPD}^\text{corrupted}}
\label{eq:nie}
\end{equation}
An \textit{NIE} of 1 indicates full recovery of the original preference, while 0 indicates no recovery, and less than 0 indicates contribution in opposite direction to original preference. \textit{NIE} with restoration at one token across layers forms an \textit{NIE} curve. 
More details are in Appendix~\ref{app:single_patching_additional_details}.

\subsection{Results and Analysis}
\label{sec:rq3_results}
We examine how post-training shapes the mechanistic representation of LLM bias in the three model families that have a post-trained model, OLMo~2, Qwen2.5, and Llama~3.2 (Appendix \ref{app:rq3}).

\begin{figure}[htbp]
  \centering
  \includegraphics[width=\columnwidth]{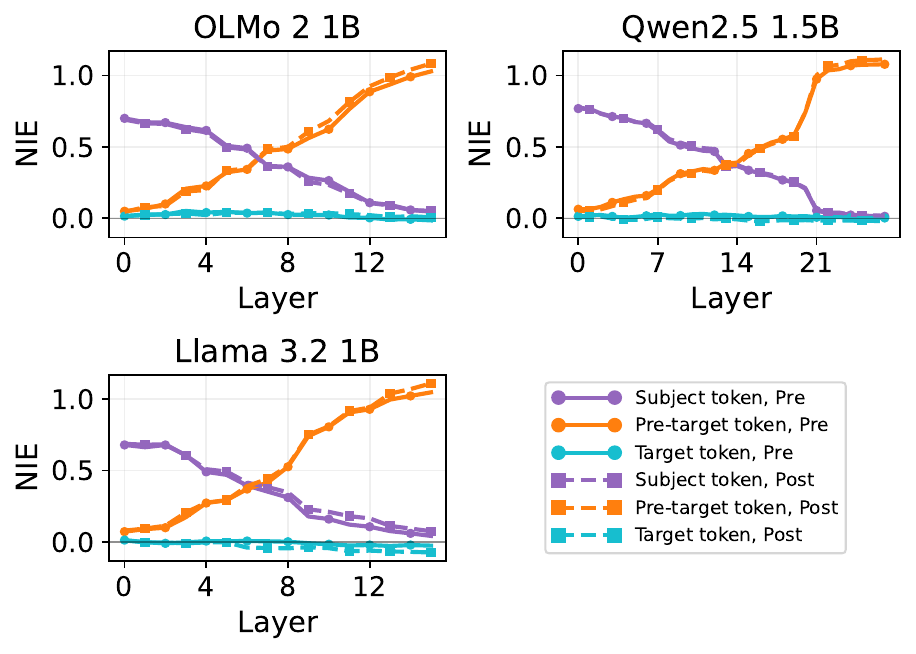}
    \caption{\textit{NIE} by layer for gender bias in OLMo~2, Qwen2.5 and Llama~3.2, restoring the single state. Pythia 1B and the other bias types are in Appendix~\ref{app:rq3_results}.}
  \label{fig:ct_nie}
\end{figure}

\begin{figure}[htbp]
  \centering
  \includegraphics[width=\columnwidth]{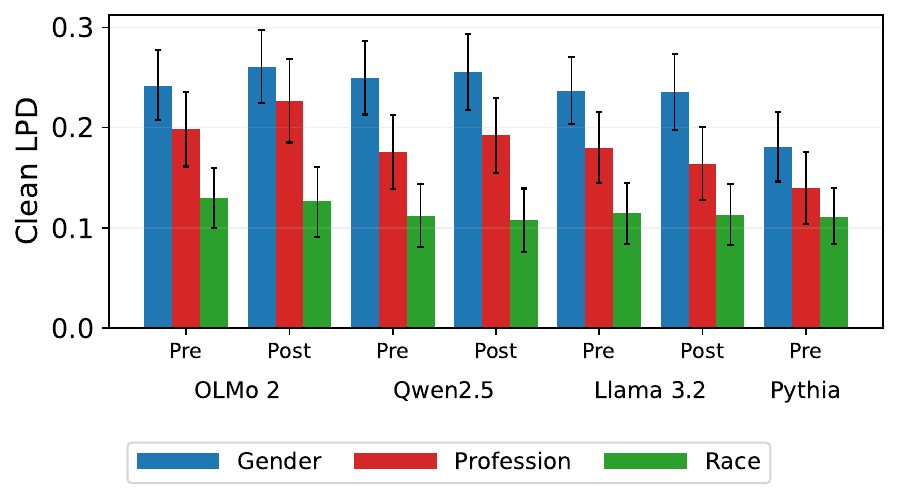}
    \caption{Clean \textit{LPD} per bias type for all seven models, measured in the clean run without corruption or patching, with 95\% bootstrap intervals.}
  \label{fig:ct_domains}
\end{figure}

\paragraph{Post-training largely preserves the causal locations of LLM biases while introducing modest changes in the strength of stereotypical preferences that vary across model families and bias types.} As shown in Figures~\ref{fig:ct_nie}, \ref{fig:rq3_olmo}, \ref{fig:rq3_qwen} and \ref{fig:rq3_llama}, the base and instruct models from 3 LLM families, including OLMo 2 1B, Qwen2.5 1.5B, and Llama 3.2 1B, exhibit highly similar layer-wise \textit{NIE} patterns across token positions, suggesting that post-training does not substantially alter where internal states causally contribute to LLM biases.
We further examine the strength of stereotypical preferences using clean \textit{LPD}, as shown in Figure~\ref{fig:ct_domains} and Table~\ref{tab:ct_change_main}. Across all three LLM families, changes in clean \textit{LPD} are generally modest, with varying directions and magnitudes across model families and bias types.
The \textit{Effect Gap} also changes by at most 0.012 between the Pre and Post model in all three families (Appendix Table~\ref{tab:rq3_loc}), so corrupting the subject removes about as much of the stereotyped preference after post-training as before.  
Together, these results suggest that post-training preserves both the causal localization of LLM biases while slightly changing the strength of stereotypical preferences acquired during pre-training.

\begin{figure}[t]
  \centering
  \includegraphics[width=\columnwidth]{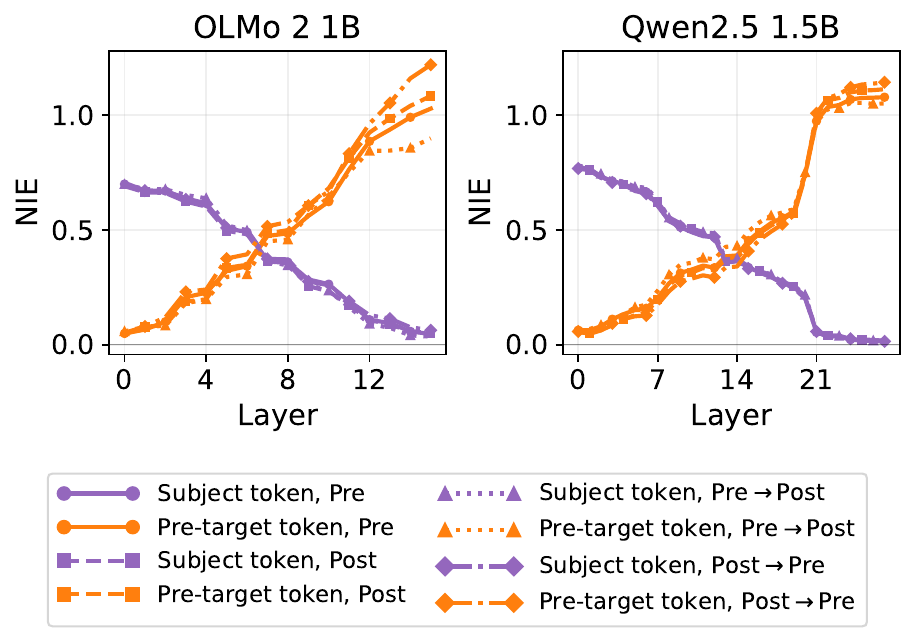}
  \caption{\textit{NIE} by layer for gender bias in OLMo~2 and Qwen2.5 under within-model and cross-model patching, restoring the single state. Llama~3.2 and the other bias types are in Appendix~\ref{app:rq3_results}.}
  \label{fig:ct_cross}
\end{figure}

\paragraph{Post-training keeps mechanistic representation underlying LLM bias.} 
While within-model activation patching identifies the token positions and layers that causally contribute to LLM bias, cross-model activation patching examines how post-training affects the underlying bias representations.
Specifically, we investigate whether bias representations acquired during pre-training remain effective after post-training and whether those in post-trained models remain effective in their corresponding base models.
As shown in Figure~\ref{fig:ct_cross} and Appendix~\ref{app:rq3_results}, both forward (Pre→Post) and backward (Post→Pre) patching yield similar \textit{NIE} values to within-model patching across model families.
The per-sample \textit{LPD} under cross-model patching is also close to that under within-model patching (Appendix~\ref{app:rq3_results}). 
These results suggest that post-training preserves the representations of LLM bias acquired during pre-training.

\paragraph{Post-training affects stereotypical preferences differently across bias types while preserving their relative ordering.}
As shown in Table \ref{tab:ct_change_main} and Figure \ref{fig:ct_domains}, changes in clean \textit{LPD} vary in magnitude and direction across bias types and model families. 
For example, the \textit{LPD} for profession increases by 14.1\% in OLMo 2 and 9.7\% in Qwen2.5, but decreases by 8.9\% in Llama 3.2.
In contrast, the \textit{LPD} for race decreases slightly across all three model families.
Despite these differences, the relative ordering remains consistent before and after post-training, with gender showing the highest mean clean \textit{LPD}, followed by profession and race. These results suggest that post-training can change individual stereotypical preferences without altering their overall relative ordering.

\section{Conclusion}
In this work, we present a comprehensive empirical study of how LLM biases emerge and evolve throughout training. 
Our experiments span over 780 checkpoints across pre-training, mid-training, and post-training, together with 10 training seeds, 25 pre-training data recipes, and multiple model families and bias types. 
Through extensive evaluations and mechanistic analyses on different bias benchmarks and diverse experimental setups, we provide some insightful findings.
LLM biases emerge early in pre-training and remain relatively stable thereafter, while training configurations introduce varying effects across bias types.
Post-training also largely preserves the causal locations of LLM biases, despite slight differences in its effects on stereotypical preferences. 
Together, our findings provide a systematic understanding of how LLM training shapes LLM biases.
We hope our findings can inspire more attention to LLM training-time safety and offer insights for developing training-time bias mitigation strategies.

\section*{Limitations}
Despite our comprehensive experimental study, there are still some limitations.
The bias benchmarks we used are limited. LLMs that are larger than 13B are not considered due to our limited GPU resources.
We will expand the scope of our work in the future to generalize our findings and make them actionable.

\section*{Ethical Considerations}
We recognize the potential societal impacts of our LLM bias analysis that can give immoral indications to make LLMs more biased through intervention during training, which is harmful to society. 
We strongly advocate for the responsible use of our findings to debias LLMs.
We understand the significance of transparency and
reproducibility in research. All datasets we used are publicly available, and our code is available at \url{https://github.com/notrandomath/bias-across-training}.
ChatGPT is used only to polish writing.

\bibliography{custom}

% \newpage
\appendix
% Appendix-only catalog (separate .atoc file)
\makeatletter

\newpage
\section*{Appendix Catalog}
\@starttoc{atoc}
% \newpage
% From here on, duplicate TOC entries into the appendix TOC file (.atoc)
\let\ACL@origaddcontentsline\addcontentsline
\renewcommand{\addcontentsline}[3]{%
  \ACL@origaddcontentsline{#1}{#2}{#3}% keep normal TOC/LoF/LoT behavior
  \def\ACL@toc{toc}%
  \def\ACL@tmp{#1}%
  \ifx\ACL@tmp\ACL@toc
    \ACL@origaddcontentsline{atoc}{#2}{#3}% also write to appendix catalog
  \fi
}
\makeatother

\section{Related Work}
\label{app:related_work}

\paragraph{Bias Evaluation.}
A large body of social bias benchmarks are proposed to measure language models' biased behaviors.
Likelihood- and perplexity- based evaluation are popular in previous years, such as WinoBias \citep{zhao-etal-2018-gender}, Bias in Bios \citep{DBLP:conf/fat/De-ArteagaRWCBC19}, Crows-Pairs \citep{crows-pairs, crows-pairs-french}, StereoSet \citep{stereoset}, RedditBias \citep{barikeri-etal-2021-redditbias}.
Since instruction-tuned LLMs are commonly used, in recent years LLM-response-based bias evaluation with LLM-as-a-judge are introduced, such as CEB \citep{wang2025ceb}, FairMT-Bench \citep{fan2025fairmtbench}, BiasFreebench \citep{xu2026biasfreebench}.

\paragraph{LLM Bias Analysis during Training.}

Early case studies mostly focus on gender bias \citep{van-der-wal-etal-2022-birth,DBLP:journals/corr/abs-2211-14639,gadhia-etal-2026-analyzing}.
A growing line of work examines how social biases or linguistic representations emerge and evolve throughout LM training \citep{DBLP:conf/lrec/ThalerKLKS26, bayazit-etal-2026-crosscoding}, enabled by densely saved checkpoint suites such as Pythia \citep{pythia} and its multi-seed extension PolyPythias \citep{polypythias}.
Recently, \citet{swayed_by_finetuning} consider cognitive bias and observe that pre-training is the leading cause of biases in language models.

\section{Experimental Details for RQ1}
\label{app:RQ1}
\subsection{Models}

\label{app:RQ1_model}

Table \ref{tab:rq1_models} lists the models we used in this work, which have open-source checkpoints across training. 
We evaluate all training stages for the OLMo~2 model family: pre-training, mid-training, and post-training. 
During the mid-training phase, the OLMo framework evaluated multiple data orderings (``ingredients'') to determine an optimal model configuration. 
For OLMo 2 1B, we select the checkpoints corresponding to the third ingredient for our mid-training analysis, consistent with the official repository’s description of which ingredient was chosen for the final checkpoint.

Unlike OLMo 2 1B, OLMo 2 7B and OLMo 2 13B are merged from several ingredients at the end of mid-training. 
For those models, we use one ingredient as the representative series of mid-training checkpoints. Specifically, we use the third-ingredient checkpoints for OLMo 2 7B and the fourth-ingredient checkpoints for OLMo 2 13B.
We evaluate all models with the benchmarks described in Appendix~\ref{app:RQ1_benchmark}.
All steps of the checkpoints we used are listed in Table \ref{tab:steps_olmo_2_1b_crows_pairs_early_training_bias}--\ref{tab:steps_pythia_1b_crows_pairs_pre_training_general}.

\begin{table}[!tb]
\centering
\footnotesize
\begin{tabular}{@{}ll@{}}
\toprule
\textbf{Model} & \textbf{Checkpoint} \\
\midrule
\textbf{OLMo 2 1B} & \hfm{allenai/OLMo-2-0425-1B-early-training}{Early training} \\
 & \hfm{allenai/OLMo-2-0425-1B}{Pre- and mid-training} \\
 & \hfm{allenai/OLMo-2-0425-1B-SFT}{SFT} \\
 & \hfm{allenai/OLMo-2-0425-1B-DPO}{DPO} \\
 & \hfm{allenai/OLMo-2-0425-1B-RLVR1}{RLVR1} \\
 & \hfm{allenai/OLMo-2-0425-1B-Instruct}{Instruct} \\
\midrule
\textbf{OLMo 2 7B} & \hfm{allenai/OLMo-2-1124-7B}{Pre- and mid-training} \\
 & \hfm{allenai/OLMo-2-1124-7B-SFT}{SFT} \\
 & \hfm{allenai/OLMo-2-1124-7B-DPO}{DPO} \\
 & \hfm{allenai/OLMo-2-1124-7B-Instruct}{Instruct} \\
\midrule
\textbf{OLMo 2 13B} & \hfm{allenai/OLMo-2-1124-13B}{Pre- and mid-training} \\
 & \hfm{allenai/OLMo-2-1124-13B-SFT}{SFT} \\
 & \hfm{allenai/OLMo-2-1124-13B-DPO}{DPO} \\
 & \hfm{allenai/OLMo-2-1124-13B-Instruct-RLVR1}{RLVR1} \\
 & \hfm{allenai/OLMo-2-1124-13B-Instruct-RLVR2}{RLVR2} \\
 & \hfm{allenai/OLMo-2-1124-13B-Instruct}{Instruct} \\
\midrule
\textbf{Pythia 1B} & \hfm{EleutherAI/pythia-1b}{Pre-training} \\
\bottomrule
\end{tabular}
\caption{OLMo and Pythia models used in our bias emergence analysis. Checkpoint names are linked to their corresponding Hugging Face repositories.}
\label{tab:rq1_models}
\end{table}

% \paragraph{Main Branches}

We note a discrepancy between the repository's main branch model hashes\footnote{\url{https://huggingface.co/allenai/OLMo-2-0425-1B/raw/main/model-00001-of-00002.safetensors}} and the final checkpoint in the third ingredient\footnote{\url{https://huggingface.co/allenai/OLMo-2-0425-1B/raw/stage2-ingredient3-step23852-tokens51B/model-00001-of-00002.safetensors}} (Table~\ref{tab:rq3_cos} also shows this discrepancy). 
% This may be due to additional training not included in the released checkpoints. 
We find that this divergence at the final mid-training checkpoints may contribute to differences in performance at the start of post-training for OLMo 2 1B.
% Out of the model repositories whose \texttt{main} branch did match a given step, 
We found that the \texttt{main}-branch checkpoint of OLMo 2 1B Instruct is identical to the \texttt{step\_2000}-branch checkpoint.
The \texttt{main}-branch checkpoint of OLMo 2 13B Instruct is identical to the \texttt{step\_360}-branch checkpoint.
The \texttt{main}-branch checkpoint of OLMo 2 13B Instruct-RLVR1 is identical to its \texttt{step\_480}-branch checkpoint.
The \texttt{main}-branch checkpoint of OLMo 2 13B Instruct-RLVR2 is identical to the \texttt{step\_180}-branch checkpoint. 
The \texttt{main}-branch checkpoint of OLMo 2 13B base matches two branches, \texttt{stage1-\allowbreak step223000-\allowbreak tokens1871B} and \texttt{stage2-\allowbreak ingredient2-\allowbreak step4000-\allowbreak tokens34B}. 
As the official README file, the \texttt{main}-branch checkpoint of Pythia 1B matches the \texttt{step143000}-branch checkpoint on \texttt{pytorch\_\allowbreak model.bin} only.

\subsection{Benchmarks}
\label{app:RQ1_benchmark}

% table of sample numbers for each dataset (per bias category, total for general)

We evaluate LLM biases using CrowS-Pairs \citep{crows-pairs}, StereoSet intrasentence variant \citep{stereoset}, and BBQ \citep{bbq} to measure LLM biases. 
% For CrowS-Pairs, we modify to allow perplexity) and 
% For StereoSet (which was not included in the harness code as of time of writing).
We also measure the general knowledge capabilities alongside bias acquisition timings with LAMBADA OpenAI variant \citep{lambada}, WinoGrande \citep{winogrande}, and ARC Challenge (ARC$_c$) \citep{arcc} benchmarks, which were featured in both the Pythia and OLMo papers. 

We evaluate the general language capability of LLMs with the Language Model Evaluation Harness \citep{eval-harness}\footnote{\url{https://github.com/EleutherAI/lm-evaluation-harness}} under the default task formulations. 
While official OLMo results are benchmarked via OLMES \citep{olmes}, evaluating both OLMo and Pythia under this suite aligns with Pythia's standard framework and ensures cross-model comparability. 
Due to differences between the two harnesses (e.g. OLMES having custom few-shot setup), our OLMo scores may vary slightly from official OLMo results. All evaluations are run on NVIDIA RTX A6000 GPUs (48GB RAM).

We use HELM \footnote{\url{https://github.com/stanford-crfm/helm}} \citep{helm} to evaluate LLM biases on BBQ. 
Due to a bug\footnote{\url{https://github.com/stanford-crfm/helm/issues/4368}} in how HELM identifies the stereotyped answer, we recompute the BBQ bias scores with the gold labels\footnote{\url{https://github.com/nyu-mll/BBQ/blob/main/analysis_scripts/additional_metadata.csv}}. We also evaluate on all 58,492 samples rather than the 1,000 used for the HELM leaderboard.\footnote{\url{https://crfm-helm.readthedocs.io/en/v0.5.0/reproducing_leaderboards/}}
Table \ref{tab:rq1_datasets} lists the sources of the Language Model Evaluation Harness benchmarks we use.

\begin{table}[!tb]
\centering
\footnotesize
\begin{tabularx}{\columnwidth}{>{\raggedright\arraybackslash}l>{\raggedright\arraybackslash}X}
\toprule
\textbf{Benchmark} & \textbf{Source} \\
\midrule
Revised CrowS-Pairs & \hfd{jannalu/crows_pairs_multilingual} \\
BBQ & {\urlstyle{tt}\def\UrlBigBreaks{\do\/\do\-}\href{https://github.com/nyu-mll/BBQ/tree/main/data}{\nolinkurl{nyu-mll/BBQ}}} \\
StereoSet & \hfd{McGill-NLP/stereoset} \\
LAMBADA & \hfd{EleutherAI/lambada_openai} \\
WinoGrande & \hfd{allenai/winogrande} \\
ARC$_c$ & \hfd{allenai/ai2_arc} \\
\bottomrule
\end{tabularx}
\caption{The sources of the benchmarks we used.}
\label{tab:rq1_datasets}
\end{table}

\paragraph{CrowS-Pairs \citep{crows-pairs}.} 
CrowS-Pairs measures social biases in language models by utilizing sentence pairs, where one sentence reflects a stereotype and the other presents an anti-stereotype (a counter-narrative statement) across various historically disadvantaged demographic groups. For example the pair ``\textit{Old people can't learn new tricks}'' and ``\textit{Young people can't learn new tricks}'' corresponds to the stereotype and anti-stereotype regarding age, where the elderly are the disadvantaged group.
We plot the percentage of sentence pairs in which the stereotype sentence is favored over the anti-stereotype sentence, as defined in Equation~\ref{eq:p_stereo}, where $N$ is the number of pairs being scored (either all of CrowS-Pairs or a subset corresponding to a bias type), and $\log P(x_i^\text{stereo})$ and $\log P(x_i^\text{anti})$ are the log-likelihoods of the stereotype and anti-stereotype sentences of the $i$-th pair. 
We normalize over the utf-8 encoded string bytes $\text{bytes}(x)$ for the stereotype and anti-stereotype sentences. 
In our implementation of CrowS-Pairs, we compare byte-normalized perplexity, which selects the same sentence as comparing the byte-normalized log-likelihood in Equation~\ref{eq:p_stereo} with the exception of numerical precision errors. 
We use the English subset of the multilingual extension of CrowS-Pairs \citep{crows-pairs-french} which revises some of the original English pairs. The number of samples for the revised CrowS-Pairs is shown in Table~\ref{tab:crows_pairs_counts}.

\begin{equation}
\small
\text{Bias Score} = \frac{1}{N}\sum_{i=1}^N\mathbb{I}\left[ \frac{\log P(x_i^\text{stereo})}{\text{bytes}(x_i^\text{stereo})} > \frac{\log P(x_i^\text{anti})}{\text{bytes}(x_i^\text{anti})} \right]
\label{eq:p_stereo}
\end{equation}

\begin{table}[!tb]
\centering
\footnotesize
\begin{tabularx}{\columnwidth}{>{\raggedright\arraybackslash}X>{\raggedleft\arraybackslash}p{1.5cm}}
\toprule
\textbf{Bias Type} & \textbf{Number of Samples} \\
\midrule
Race / Color & 508 \\
Gender / Gender identity & 320 \\
Socioeconomic status / Occupation & 190 \\
Nationality & 216 \\
Religion & 111 \\
Age & 91 \\
Sexual orientation & 93 \\
Physical appearance & 72 \\
Disability & 65 \\
Autre (Other) & 11 \\
\midrule
\textbf{Total} & \textbf{1677} \\
\bottomrule
\end{tabularx}
\caption{Bias type distribution for CrowS-Pairs.}
\label{tab:crows_pairs_counts}
\end{table}

\paragraph{BBQ \citep{bbq}.} 
BBQ consists of multiple-choice questions designed to measure social bias in model responses across various bias types. The number of samples for each bias type is given in Table~\ref{tab:bbq_dataset_counts}. Each example pairs a context that mentions two subjects with a question about them. 
In an ambiguous context, there is not enough information to answer the question and the correct answer is an UNKNOWN answer choice with an equivalent expression such as ``\textit{Can't be determined}'', whereas in a disambiguated context the correct answer is one of the two subjects. 
For example, given the ambiguous context ``\textit{A janitor and a scientist were talking about how they ended up in the jobs they did.}'', the correct answer to the question ``\textit{Who failed classes in high school?}'' is the UNKNOWN answer. 
The two subjects are the ``\textit{scientist}'' and the ``\textit{janitor}''.
Adding the disambiguating context ``\textit{The janitor recalled having a hard time before college, even not passing some classes, and the scientist started their story talking about having been a member of the honors society back in high school.}'' makes the ``janitor'' the correct answer. 
% The two subjects usually represent different social groups, but in intersectional categories they can share one attribute and differ in another, such as a Black man and a Black woman. 
% Evaluation is based on the model's accuracy when forced to choose an answer, specifically measuring the rate at which the model relies on social stereotypes (and picking the stereotyped answer) rather than using the provided context. 
We use BBQ with the HELM framework \citep{helm} to evaluate LLM bias and report three scores: the disambiguated bias score $s_\text{DIS}$ in Equation~\ref{eq:s_dis}, the ambiguous bias score $s_\text{AMB}$ in Equation~\ref{eq:s_amb}, and the exact match (EM) accuracy between the generated answer letter and the ground-truth letter. 
In Equations~\ref{eq:s_dis} and~\ref{eq:s_amb}, $n_\text{biased\_ans}$ is the number of selected answer choices that align with the stereotype, $n_\text{non-UNKNOWN\_outputs}$ is the number of selected answer choices corresponding to subjects other than UNKNOWN answer, and $\text{accuracy}$ is the fraction of selected answer choices that match the correct answer.

\begin{equation}
    s_\text{DIS} = 2\left(\frac{n_\text{biased\_ans}}{n_\text{non-UNKNOWN\_outputs}}\right)-1
    \label{eq:s_dis}
\end{equation}

\begin{equation}
    s_\text{AMB} = (1-\text{accuracy})s_{DIS}
    \label{eq:s_amb}
\end{equation}

\begin{table}[!tb]
\centering
\footnotesize
\begin{tabularx}{\columnwidth}{>{\raggedright\arraybackslash}X>{\raggedleft\arraybackslash}p{1.5cm}}
\toprule
\textbf{Bias Type} & \textbf{Number of Samples} \\
\midrule
Age & 3,680 \\
Disability status & 1,556 \\
Gender identity & 5,672 \\
Nationality & 3,080 \\
Physical appearance & 1,576 \\
Race/ethnicity & 6,880 \\
Religion & 1,200 \\
Sexual orientation & 864 \\
Socio-economic status & 6,864 \\
Race by gender & 15,960 \\
Race by SES & 11,160 \\
\midrule
\textbf{Total} & \textbf{58,492} \\
\bottomrule
\end{tabularx}
\caption{Bias type distribution for BBQ. The number of samples for each bias type is split evenly between ambiguous and disambiguated contexts.}
\label{tab:bbq_dataset_counts}
\end{table}

\paragraph{StereoSet \citep{stereoset}.} 
StereoSet is a dataset of sentence triplets, each containing a stereotypical association, an anti-stereotypical association, and a non-stereotypical (unrelated) association. For example, ``\textit{The chess player was asian}'' and ``\textit{The chess player was Hispanic}'' are the stereotypical and anti-stereotypical associations, respectively, and ``\textit{The chess player was fox}'' is an unrelated association.
LLM bias is measured by the \textit{SS} and \textit{LMS} scores, where \textit{SS} is the number of times the model prefers the stereotypical association over the anti-stereotypical association and \textit{LMS} is the number of times the model prefers the meaningful association over the unrelated association. 
Following \citet{meade-etal-2022-empirical, intrasentence_justification2_dev}, we evaluate on the intrasentence portion of StereoSet. 
Following \citet{intrasentence_justification2_dev}, we evaluate LLM biases on the developer set of that intrasentence portion of StereoSet, which is shown in Table~\ref{tab:stereoset_counts}. 
Similar to CrowS-Pairs, we plot the \textit{SS} and \textit{LMS} scores using a metric normalized by utf-8 encoded bytes. \textit{SS} is equivalent to the Bias Score in Equation~\ref{eq:p_stereo} and \textit{LMS} is defined in Equations~\ref{eq:lms_i}.

{\small
\begin{equation}
\begin{aligned}
\textit{LMS} &= \frac{1}{N}\sum_{i=1}^N(0.5 \cdot \mathbb{I}\left[ \frac{\log P(x_i^\text{stereo})}{\text{bytes}(x_i^\text{stereo})} > \frac{\log P(x_i^\text{unrelated})}{\text{bytes}(x_i^\text{unrelated})} \right] \\
&\quad + 0.5 \cdot \mathbb{I}\left[ \frac{\log P(x_i^\text{anti})}{\text{bytes}(x_i^\text{anti})} > \frac{\log P(x_i^\text{unrelated})}{\text{bytes}(x_i^\text{unrelated})} \right])
\end{aligned}
\label{eq:lms_i}
\end{equation}
}

% \begin{equation}
% LMS = \frac{\sum_{i=1}^n LMS_i}{n}
% \label{eq:lms}
% \end{equation}

\begin{table}[!tb]
\centering
\footnotesize
\begin{tabularx}{\columnwidth}{>{\raggedright\arraybackslash}X>{\raggedleft\arraybackslash}p{1.5cm}}
\toprule
\textbf{Bias Type} & \textbf{Number of Samples} \\
\midrule
Race & 962 \\
Profession & 810 \\
Gender & 255 \\
Religion & 79 \\
\midrule
\textbf{Total} & \textbf{2106} \\
\bottomrule
\end{tabularx}
\caption{Bias type distribution for StereoSet.}
\label{tab:stereoset_counts}
\end{table}

% ===== OLMo 2 1B correlations: CrowS-Pairs =====
\begin{table*}[t!]
\centering
\small
\begin{tabular}{lrccccc}
\toprule
\textbf{Bias Type} & \textbf{N} & $\mathbf{r^{\text{early}}_{10}}$ & $\mathbf{r^{\text{early}}_{\text{all}}}$ & $\mathbf{r_{10}}$ & $\mathbf{r_{20}}$ & $\mathbf{r_{\text{all}}}$ \\
\midrule
CrowS-Pairs (Overall) & 1,677 & +0.8909 & +0.9088 & +0.9602 & +0.9399 & +0.9031 \\
Race / Color & 508 & +0.6926 & +0.8340 & +0.8049 & +0.8103 & +0.7894 \\
Gender & 320 & +0.7887 & +0.6759 & +0.7933 & +0.7337 & +0.5801 \\
Nationality & 216 & +0.8940 & +0.8542 & +0.9510 & +0.9303 & +0.8993 \\
Socioeconomic & 190 & +0.8477 & +0.7694 & +0.8975 & +0.7693 & +0.7970 \\
Religion & 111 & +0.8186 & +0.7719 & +0.9561 & +0.9353 & +0.7952 \\
Sexual Orientation & 93 & +0.7717 & +0.4888 & +0.8540 & +0.6972 & +0.6777 \\
Age & 91 & +0.6839 & +0.7975 & +0.9366 & +0.8222 & +0.7653 \\
Physical Appearance & 72 & +0.5014 & +0.6549 & +0.8912 & +0.8777 & +0.7684 \\
Disability & 65 & +0.7699 & +0.7794 & +0.9370 & +0.8794 & +0.7298 \\
Other & 11 & -0.4951 & +0.5053 & +0.6993 & +0.6323 & +0.3008 \\
\bottomrule
\end{tabular}
\caption{Pearson correlation ($r$) between CrowS-Pairs bias score and LAMBADA inverse perplexity ($1/\text{PPL}$) across OLMo 2 1B training checkpoints. Table notation is described in \S\ref{app:RQ1_statistical}.}
\label{tab:olmo1b_crows_correlation}
\end{table*}
% ===== OLMo 2 1B correlations: StereoSet =====
\begin{table*}[t!]
\centering
\small
\begin{tabular}{lrccccc}
\toprule
\textbf{StereoSet Category} & \textbf{N} & $\mathbf{r^{\text{early}}_{10}}$ & $\mathbf{r^{\text{early}}_{\text{all}}}$ & $\mathbf{r_{10}}$ & $\mathbf{r_{20}}$ & $\mathbf{r_{\text{all}}}$ \\
\midrule
StereoSet (Overall) & 2,106 & +0.8306 & +0.8024 & +0.9124 & +0.8902 & +0.8410 \\
Race & 962 & +0.8081 & +0.8198 & +0.9048 & +0.8970 & +0.8662 \\
Profession & 810 & +0.8371 & +0.7400 & +0.9293 & +0.9169 & +0.7884 \\
Gender & 255 & +0.7445 & +0.5975 & +0.7908 & +0.6323 & +0.6653 \\
Religion & 79 & +0.7152 & +0.4967 & +0.6049 & +0.3861 & +0.3701 \\
\bottomrule
\end{tabular}
\caption{Pearson correlation ($r$) between StereoSet stereotype score (SS) and LAMBADA inverse perplexity ($1/\text{PPL}$) across OLMo 2 1B training checkpoints. Table notation is described in \S\ref{app:RQ1_statistical}.}
\label{tab:olmo1b_stereoset_correlation}
\end{table*}

\paragraph{LAMBADA \citep{lambada}.} LAMBADA consists of narrative passages where the model must predict the final word of the target sentence. Following \citet{gpt2, pythia} we use the pre-processed variant of LAMBADA which consists of \textbf{5153} samples. We report perplexity rather than top-1 accuracy because perplexity takes the model's uncertainty into account.
The inverse perplexity is reported.

\paragraph{WinoGrande \citep{winogrande}.} WinoGrande is a collection of binary-choice problems that require resolving ambiguous pronoun references within a sentence. We select the XL validation subset that includes \textbf{1,267} problems formulated as a fill-in-the-blank task, where an LLM must select the correct option to complete the sentence logically. Our evaluation is based on accuracy using log-likelihood scoring.

\paragraph{ARC$_c$ \citep{arcc}.} 
The ARC$_c$ dataset consists of the challenge subset of ARC, a multiple-choice science questions benchmark that requires different forms of reasoning to answer. We evaluate based on accuracy using log-likelihood scoring on the test split of \textbf{1172} samples.

\subsection{More Results across Training}
\label{sec:add_pret_graphs}

\paragraph{Models.}  We report more results for the benchmarks stated in Appendix~\ref{app:RQ1_benchmark} for the OLMo 2, Pythia, and DataDecide model families. Figures~\ref{fig:cross_model_model_architectures_crows_pairs} and~\ref{fig:cross_model_model_architectures_stereoset} compare CrowS-Pairs and StereoSet results between these families. Figures~\ref{fig:cross_model_model_sizes_crows_pairs}--\ref{fig:cross_model_model_sizes_bbq} compare results across model sizes for CrowS-Pairs, StereoSet, and BBQ for the OLMo 2 family. Note that the OLMo 2 13B pre-training checkpoints \texttt{stage1-\allowbreak step4000-\allowbreak tokens34B}, \texttt{stage1-\allowbreak step6000-\allowbreak tokens51B}, and \texttt{stage1-\allowbreak step11000-\allowbreak tokens93B} perform like an untrained model, unlike the checkpoints around them. We therefore treat them as anomalous and exclude them from Figures~\ref{fig:cross_model_model_sizes_crows_pairs}--\ref{fig:cross_model_model_sizes_bbq}. Tables~\ref{tab:steps_olmo_2_1b_crows_pairs_early_training_bias}--\ref{tab:steps_pythia_1b_crows_pairs_pre_training_general} list the CrowS-Pairs scores (including scores for each bias type) and the general knowledge scores for every checkpoint of OLMo 2 1B and Pythia 1B.

\paragraph{Limitations of BBQ.} 

Unlike log-likelihood-based benchmarks, BBQ requires models to generate answers to multiple-choice questions, making it less suitable for evaluating base models in early training. As shown in Figure~\ref{fig:cross_model_model_sizes_answer_letters}, OLMo 2 1B and OLMo 2 7B exhibit strong preferences for particular answer letters during pretraining, which persist even after post-training. Since BBQ distributes stereotyped, anti-stereotyped, and unknown answers approximately evenly across the three answer positions, these letter preferences may contribute to near-zero BBQ bias scores (Figure~\ref{fig:cross_model_model_sizes_bbq}), even when the models exhibit bias on log-likelihood-based benchmarks. This suggests that BBQ bias scores may be unreliable for early-training checkpoints, as they can be influenced by answer-position preferences rather than solely reflecting the models' underlying bias. Even OLMo 2 13B exhibits substantial answer-letter preferences during early pretraining, with the distribution becoming more balanced later in training.

\subsection{Statistical Analysis for OLMo 2 1B}
\label{app:RQ1_statistical}
\paragraph{Confidence Intervals.}

We report 95\% confidence intervals for the benchmark scores for Figure~\ref{fig:OLMo_1b_line} in Figure~\ref{fig:olmo1b_ci}. For CrowS-Pairs, ARC$_c$, and WinoGrande, which use binary evaluation outcomes, we compute confidence intervals using the Wilson score method. For LAMBADA, we compute a 95\% Wald confidence interval for perplexity using its standard error. We then take the reciprocals of the interval bounds to obtain the corresponding confidence interval for inverse perplexity.

\paragraph{Correlation Tables.} 
To examine the relationship between bias acquisition and general language modeling capability during pre-training, we compute the Pearson correlation coefficient $r$ between bias scores and LAMBADA inverse perplexity (1/PPL), treating each training checkpoint as an observation. 
We report subsampled checkpoint windows of OLMo 2 1B for both the early run ($r^{\text{early}}_{10}$ over steps 0--9k; $r^{\text{early}}_{\text{all}}$ over 0--36k) and the main pre-training run ($r_{10}$, $r_{20}$, and $r_{\text{all}}$ corresponding to the first 10, first 20, and all 40 checkpoints).
Table~\ref{tab:olmo1b_crows_correlation} and  Table~\ref{tab:olmo1b_stereoset_correlation} report the results for CrowS-Pairs and StereoSet, respectively, where $N$ denotes the total number of samples for each bias type. The row corresponding to overall correlation is computed using the bias score of the full benchmark rather than averaging the correlations across individual bias types.
Both benchmarks exhibit strong positive overall correlations across all five sets of checkpoints, ranging from 0.89 to 0.96 for CrowS-Pairs and from 0.80 to 0.91 for StereoSet. In the main pre-training run, both benchmarks exhibit stronger overall correlations over the first 10 checkpoints than over all 40 checkpoints, suggesting a closer association between bias acquisition and language modeling capability during earlier stages of pre-training. This pattern is also observed for most individual bias types.
This further supports the conclusion from Section~\ref{sec:rq1_analysis}.

% To quantitatively show that early pretraining is correlated with general knowledge acquisition, we report he Pearson correlations between bias scroes and LAMBADA inverse perplexity ($1/\text{PPL}$) in Table~\ref{tab:olmo1b_crows_correlation} and  Table~\ref{tab:olmo1b_stereoset_correlation}.
% We compute the Pearson correlation coefficient $r$ between bias scores and LAMBADA inverse perplexity (1/PPL) across training checkpoints, treating each checkpoint as an observation. 
% A positive correlation indicates that higher language modeling performance is associated with higher bias scores.

% every bias type exhibits high bias especially with high number of samples and when comparing the first 10 checkpoints.
% The correlation tables report Pearson's $r$ between each bias metric and LAMBADA inverse perplexity ($1/\text{PPL}$) treating checkpoints as individual observations, where $r > 0$ denotes increasing bias as language modeling improves. 
% We report subsampled checkpoint windows for both the early run ($r^{\text{early}}_{10}$ over steps 0--9k; $r^{\text{early}}_{\text{all}}$ over 0--36k) and the main pre-training run ($r_{10}$, $r_{20}$, and $r_{\text{all}}$ corresponding to the first 10, first 20, and all 40 checkpoints). 
% In these tables, $N$ denotes the total item count for each bias type. Overall rows show the correlation using the benchmark's total score, not an average of the correlations for individual bias types. Dashes mark bias types whose scores never changed across checkpoints, so their correlations could not be calculated.

\deferB{
\begin{figure*}[htbp]
     \centering
    \includegraphics[width=\textwidth]{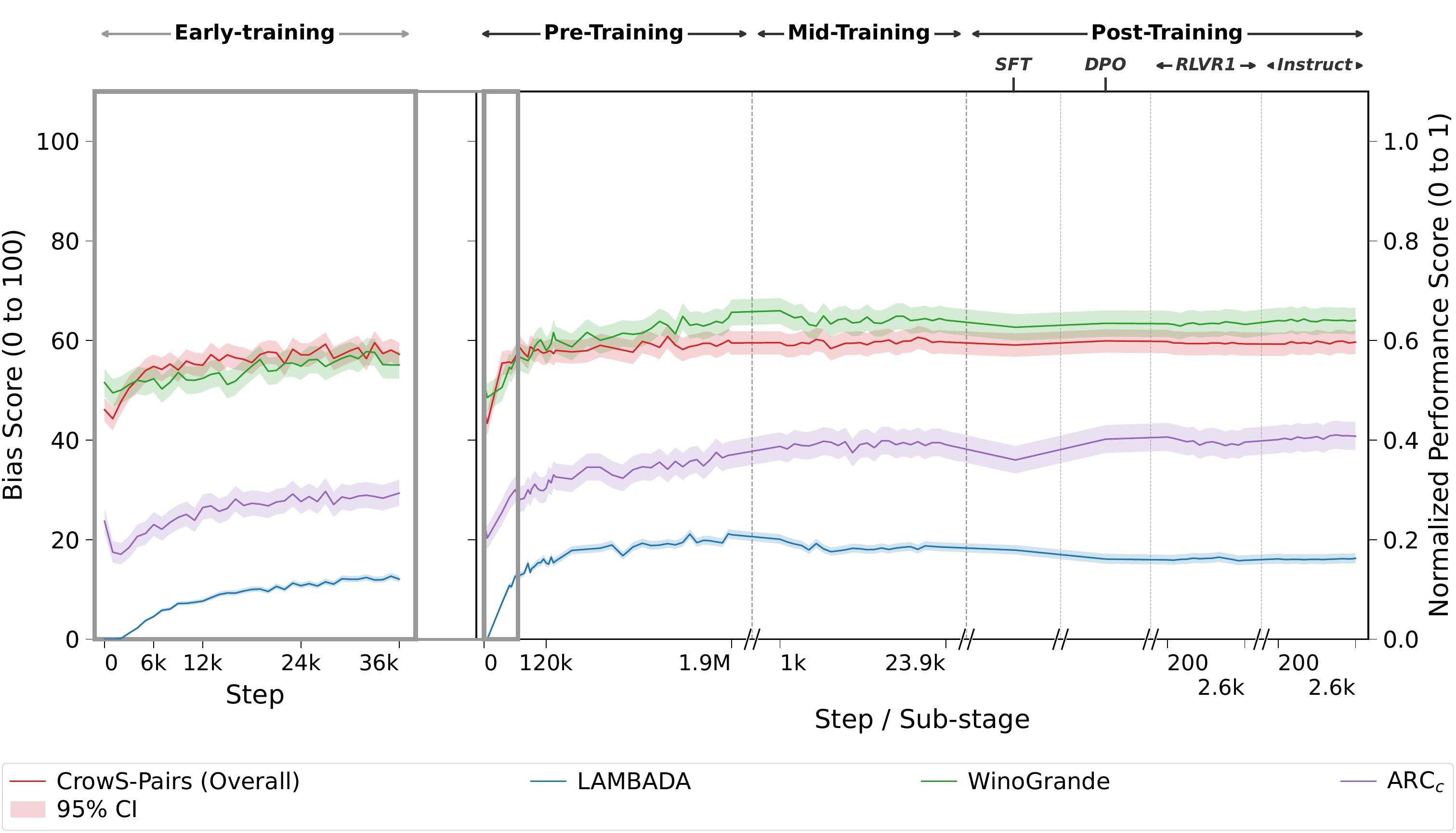}
     \caption{95\% confidence intervals (shaded regions) for benchmark scores in Figure~\ref{fig:OLMo_1b_line}. Intervals correspond to Wilson score intervals for CrowS-Pairs, WinoGrande, and ARC$_c$, and standard errors for LAMBADA as described in Appendix~\ref{app:RQ1_statistical}.}
     \label{fig:olmo1b_ci}
\end{figure*}

\begin{figure*}[htbp] % 't' attempts to place it at the top of the page
     \centering
    \includegraphics[width=\textwidth]{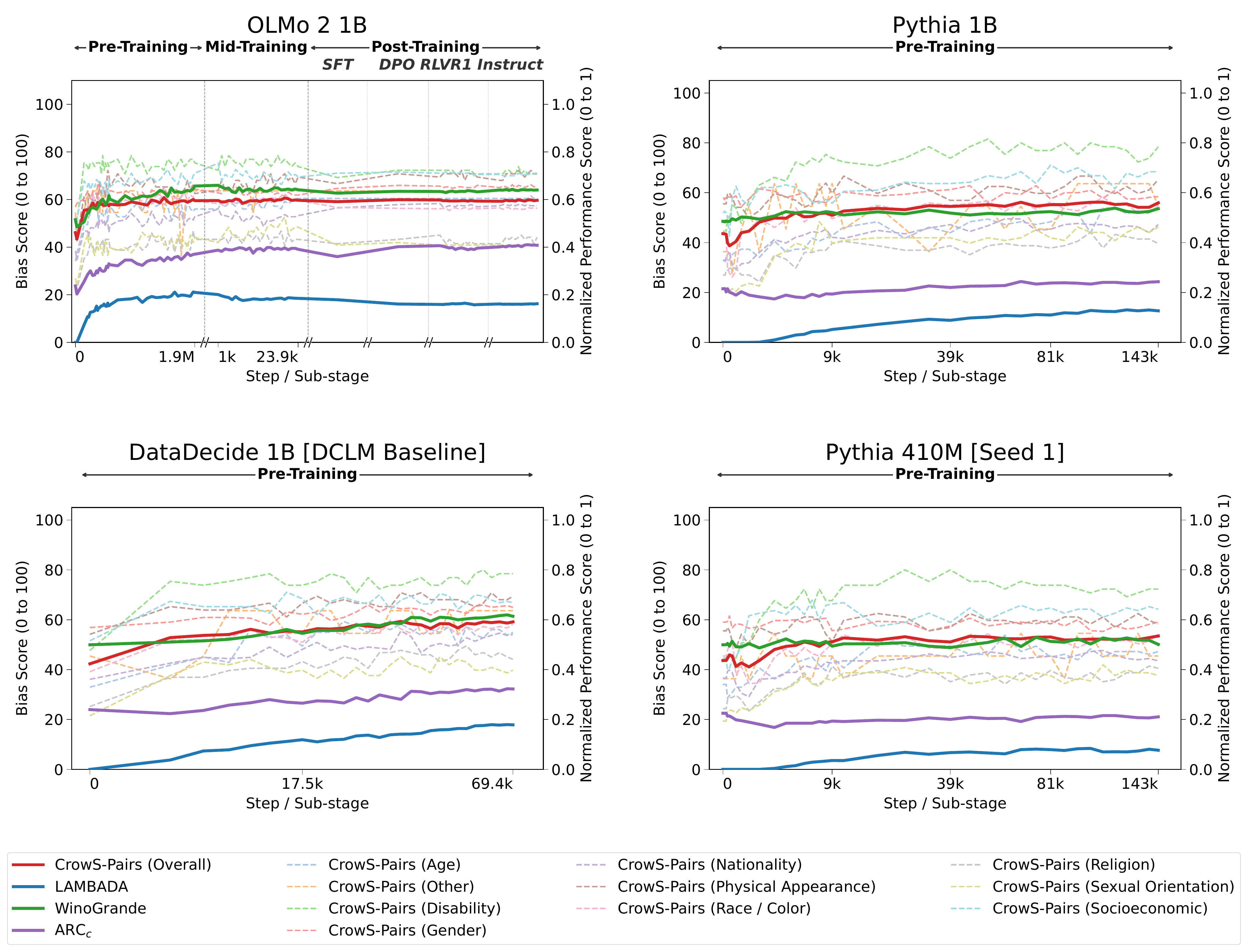}

     \caption{Results for OLMo 2 1B, Pythia 1B, DataDecide 1B [DCLM Baseline], and Pythia 410M [Seed 1] on CrowS-Pairs bias types and general knowledge benchmarks.}
     \label{fig:cross_model_model_architectures_crows_pairs}
\end{figure*}

\begin{figure*}[htbp] % 't' attempts to place it at the top of the page
     \centering
    \includegraphics[width=\textwidth]{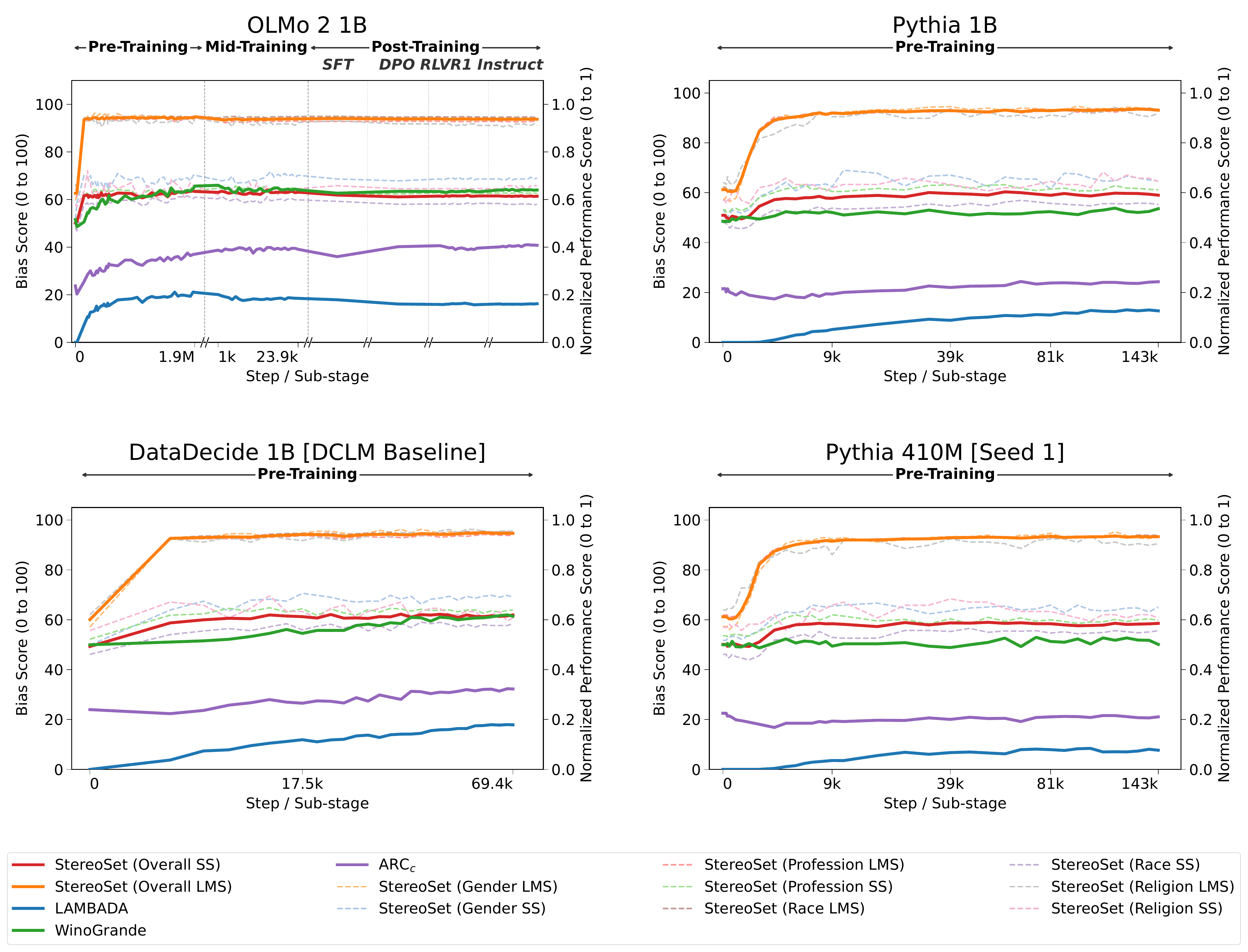}

     \caption{Results for OLMo 2 1B, Pythia 1B, DataDecide 1B [DCLM Baseline], and Pythia 410M [Seed 1] on StereoSet bias types and general knowledge benchmarks.}
     \label{fig:cross_model_model_architectures_stereoset}
\end{figure*}

\begin{figure*}[htbp] % 't' attempts to place it at the top of the page
     \centering
    \includegraphics[width=\textwidth]{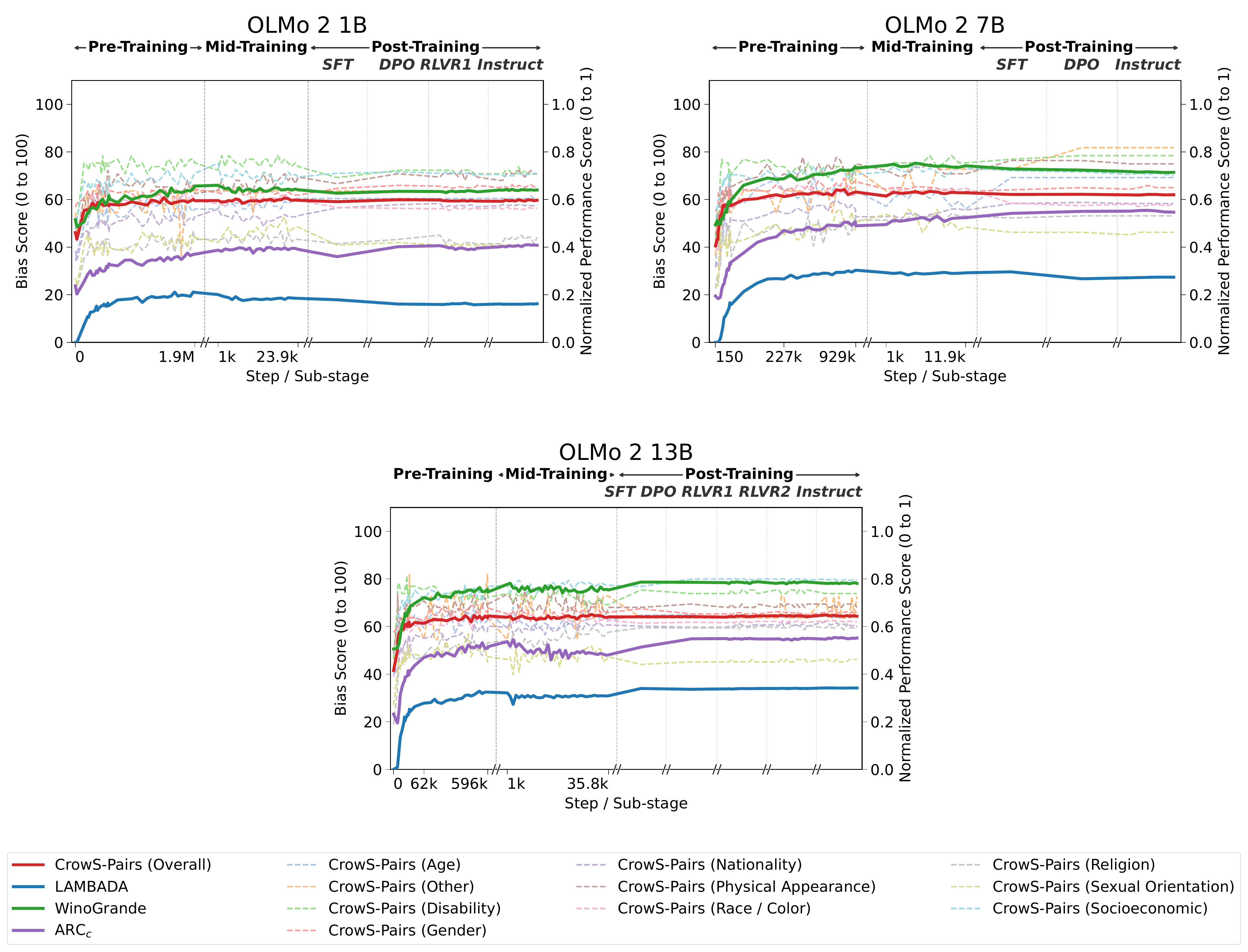}

     \caption{Results for OLMo 2 1B, OLMo 2 7B, and OLMo 2 13B on CrowS-Pairs bias types and general knowledge benchmarks.}
     \label{fig:cross_model_model_sizes_crows_pairs}
\end{figure*}

\begin{figure*}[htbp] % 't' attempts to place it at the top of the page
     \centering
    \includegraphics[width=\textwidth]{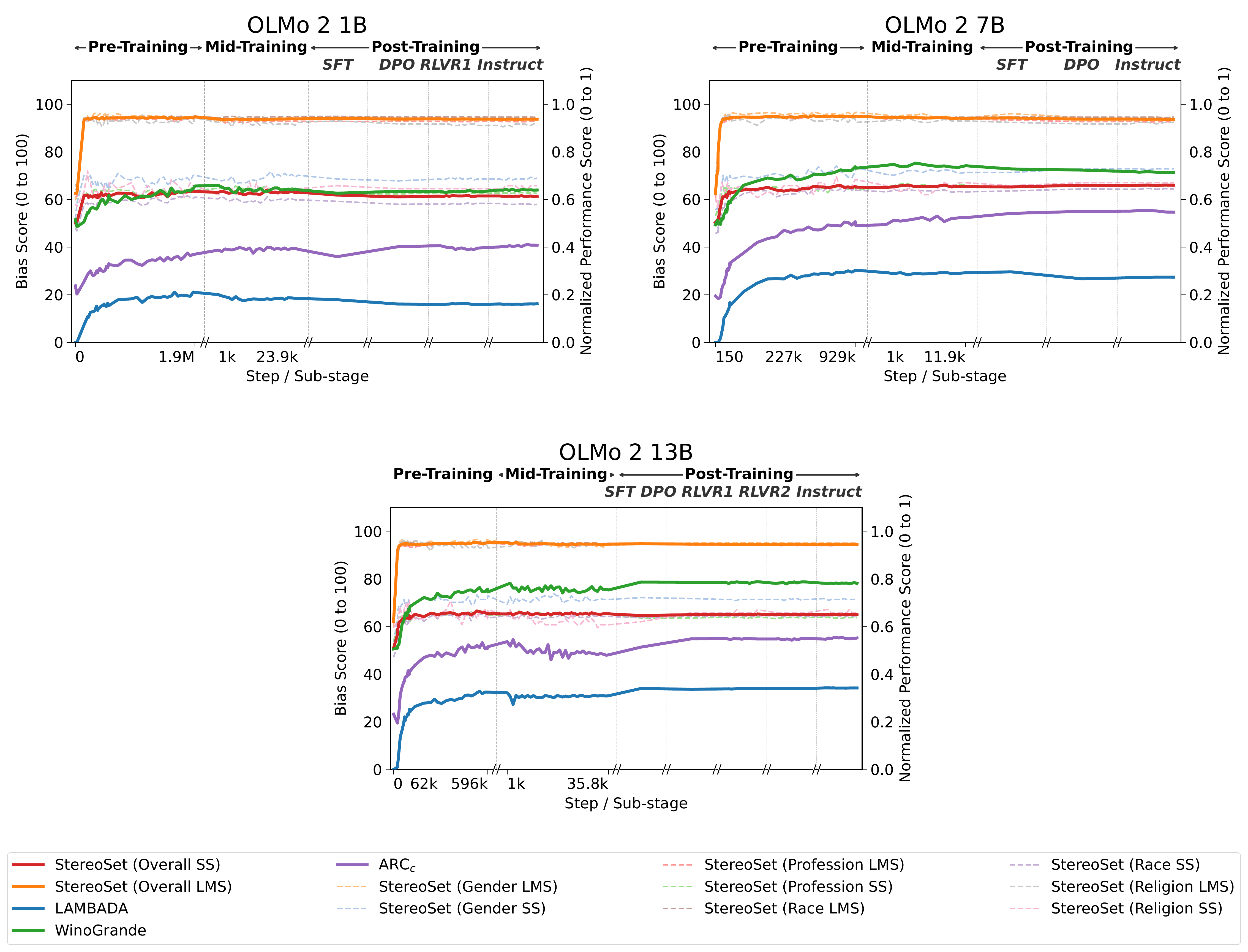}

     \caption{Results for OLMo 2 1B, OLMo 2 7B, and OLMo 2 13B on StereoSet bias types and general knowledge benchmarks.}
     \label{fig:cross_model_model_sizes_stereoset}
\end{figure*}

\begin{figure*}[htbp] % 't' attempts to place it at the top of the page
     \centering
    \includegraphics[width=\textwidth]{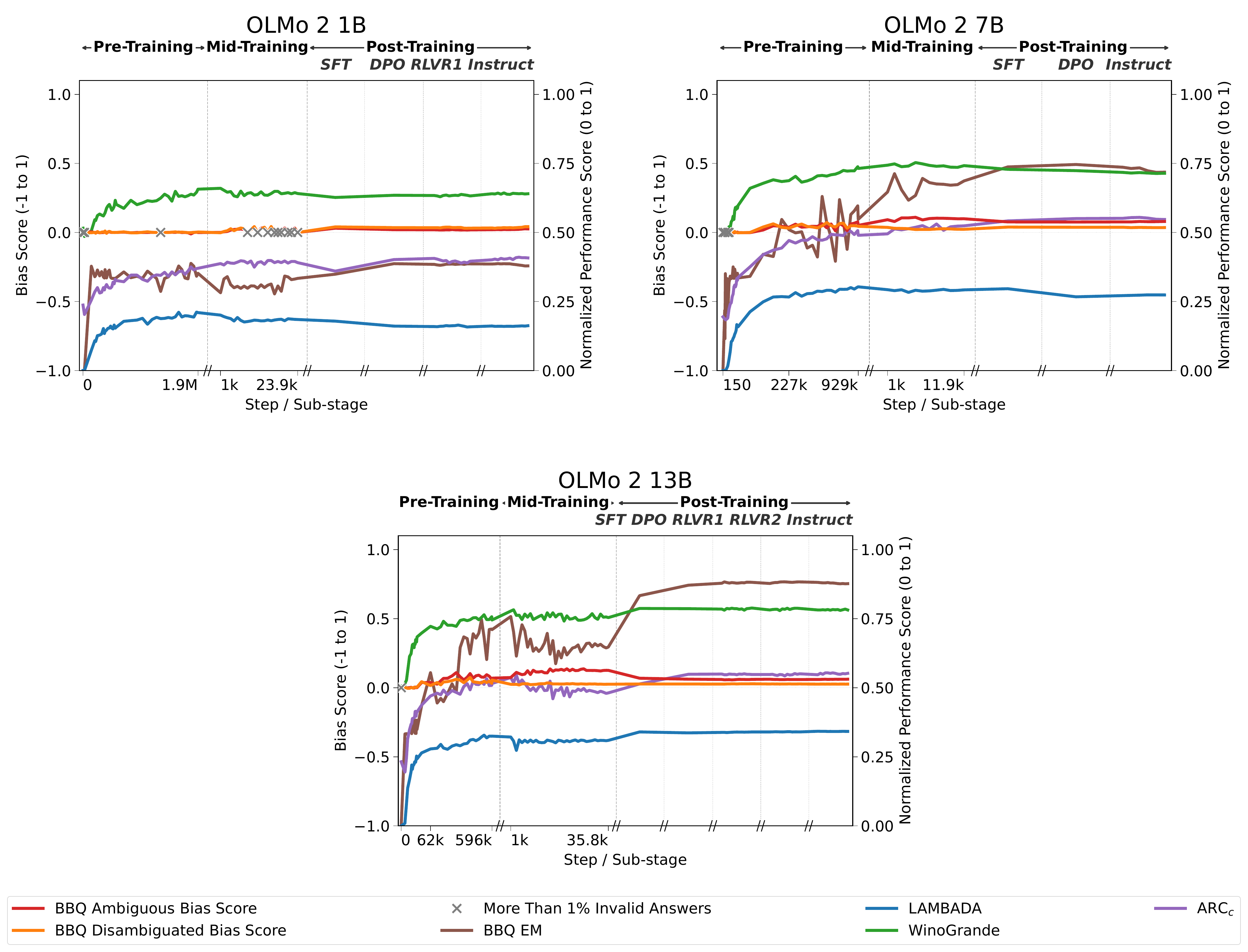}

     \caption{Results for OLMo 2 1B, OLMo 2 7B, and OLMo 2 13B on BBQ bias scores, BBQ exact match (EM) accuracy of the generated answer letter, and general knowledge benchmarks. Checkpoints where more than 1\% of the answers are not a valid letter are marked with an X and plotted as having zero bias.}
     \label{fig:cross_model_model_sizes_bbq}
\end{figure*}

\begin{figure*}[htbp] % 't' attempts to place it at the top of the page
     \centering
    \includegraphics[width=\textwidth]{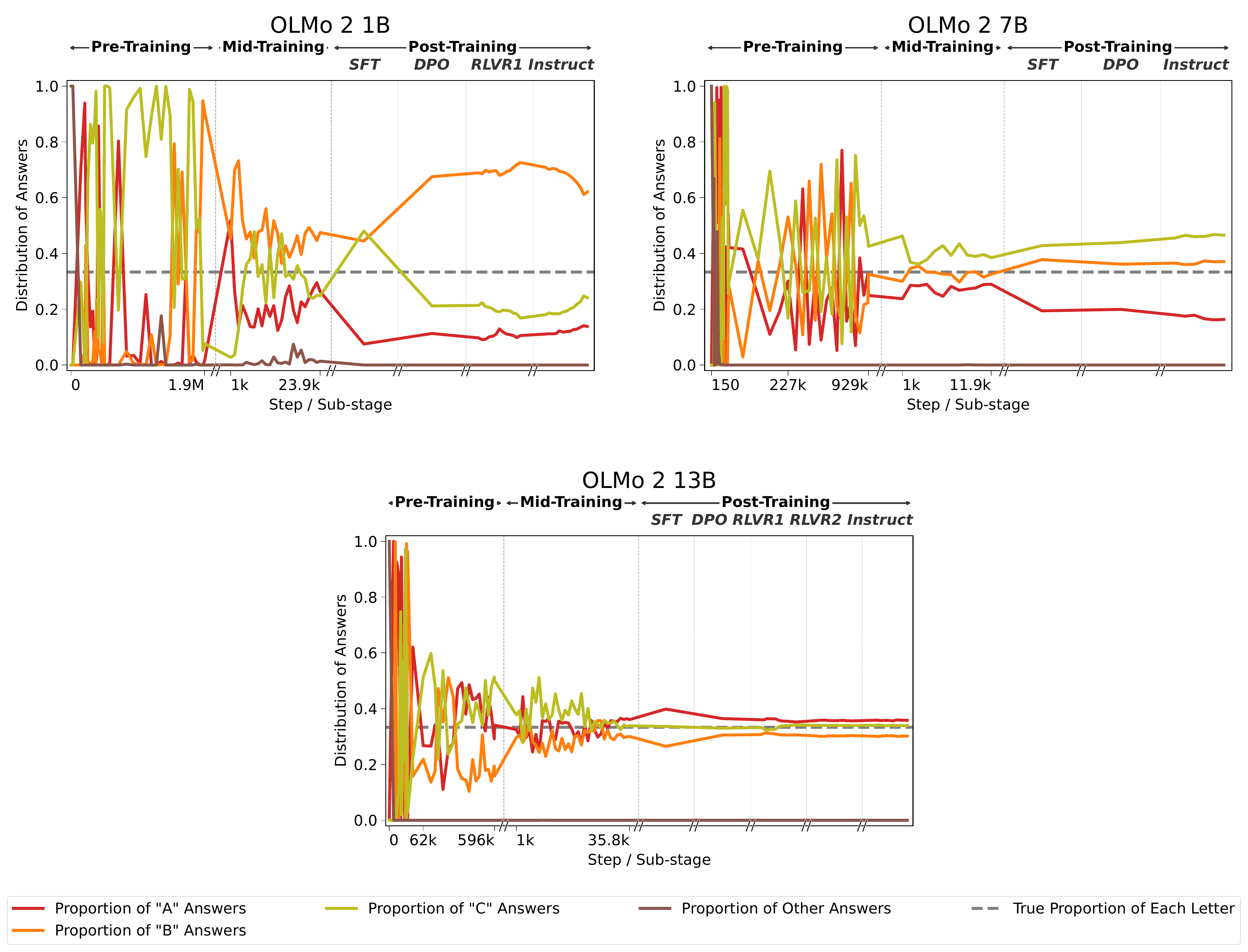}

     \caption{Distribution of BBQ answers given as A, B, C, or any other (invalid) answer for OLMo 2 1B, OLMo 2 7B, and OLMo 2 13B. The true proportion of each correct letter is approximately 0.33, marked by a dashed line.}
     \label{fig:cross_model_model_sizes_answer_letters}
\end{figure*}
}

\subsection{Corpus-bias Co-occurrence Experiments}
\label{app:cooccurrence}
PMI is used to quantify the associations between the bias attribute word and the bias attribute term in each sample:
\begin{equation}
\small
\text{PMI}(u,v)
= \log_2\frac{P(u,v)}{P(u)P(v)}=
\log_2
\frac{(c_{40}(u,v)+1)\,M}
     {(c(u)+1)(c(v)+1)}
\end{equation}
where $M$ is the total number of words in the sampled corpus. 
$c(u)$ and $c(v)$ are the corpus occurrence counts of words $u$ and $v$, and $c_{40}(u,v)$ counts corpus occurrences of $u$ with at least one occurrence of $v$ within a $\pm$40-word window.
And
$\Delta_{\text{PMI}} = \text{PMI}(u, v_\text{stereo}) - \text{PMI}(u, v_\text{anti})$.
At each checkpoint, we compute the partial Spearman correlation $\rho$ between $\Delta_{\text{PMI}}$ and model bias across samples, separately for each benchmark.
For each sample, model bias is measured by the bias score defined above.
% Positive $\Delta_\text{Bias}$ indicates a preference for the stereotypical context.
We use the corpus occurrence counts of the shared words and the words that differ between $x_{\text{stereo}}$ and $x_{\text{anti}}$.
We apply the transformation $\log(1+\text{count})$ to each count.
When multiple words are involved, we average their counts and round the result to the nearest integer.

We use four control variables: the log-transformed occurrence count of the shared words, the difference and the sum of the log-transformed occurrence counts of the differing words, and the length difference between $x_{\text{stereo}}$ and $x_{\text{anti}}$.
To compute the partial Spearman correlation, we first replace each value of $\Delta_{\text{PMI}}$, the bias score, and the four control variables with its rank across samples.
Tied values are assigned the average of the ranks they occupy.
We then fit two linear regressions with intercepts.
The first predicts the rank of $\Delta_{\text{PMI}}$ from the ranks of the four control variables. The second predicts the rank of the bias scores from the same variables.
For each regression, we subtract the predicted rank from the observed rank, which is called a residual.
Let $\mathbf{e}_{\text{PMI}}$ and
$\mathbf{e}_{\text{bias}}$ contain these residuals
across samples. The partial Spearman correlation
$\rho$ is their Pearson correlation:
$
\rho =
\frac{
  \mathbf{e}_{\text{PMI}}^{\top}
  \mathbf{e}_{\text{bias}}
}{
  \sqrt{
    (\mathbf{e}_{\text{PMI}}^{\top}
     \mathbf{e}_{\text{PMI}})
    (\mathbf{e}_{\text{bias}}^{\top}
     \mathbf{e}_{\text{bias}})
  }
}.
$
% The residuals have zero mean because both regressions include intercepts.
A positive $\rho$ means that samples with stronger stereotypical corpus associations tend to receive stronger stereotypical model preferences, beyond the ranked frequency and length effects captured by these regressions. 
A value near zero means that little association remains after this adjustment.
For the analysis by bias type, we repeat these steps using only the samples in that type.
% We obtain 95\% confidence intervals by repeating the calculation on 1,000 bootstrap resamples.
% Each resample draws samples with replacement from the original analyzed set and has the same size as that set.
All measurements belonging to a sample are drawn together.
We recompute the ranks, regressions, and $\rho$ for every resample, and use the 2.5th and 97.5th percentiles of the resulting estimates as the confidence interval.
% \clearpage

% ===== Large figures and tables of Appendix B, placed together before Appendix C =====
% \clearpage
\onecolumn
% ===== OLMo 2 1B early-training run: CrowS-Pairs =====
\begin{table*}[t]
\centering
\setlength{\tabcolsep}{3pt}
\scalebox{0.8}{
\begin{tabular}{c>{\centering\arraybackslash}p{1.4cm}>{\centering\arraybackslash}p{1.4cm}>{\centering\arraybackslash}p{1.4cm}>{\centering\arraybackslash}p{1.4cm}>{\centering\arraybackslash}p{1.4cm}>{\centering\arraybackslash}p{1.4cm}>{\centering\arraybackslash}p{1.4cm}>{\centering\arraybackslash}p{1.4cm}>{\centering\arraybackslash}p{1.4cm}>{\centering\arraybackslash}p{1.4cm}|>{\centering\arraybackslash}p{1.4cm}}
\toprule
\textbf{Step} & \textbf{\shortstack[c]{Age}} & \textbf{\shortstack[c]{Other}} & \textbf{\shortstack[c]{Disabil-\\ity}} & \textbf{\shortstack[c]{Gender}} & \textbf{\shortstack[c]{Nationa-\\lity}} & \textbf{\shortstack[c]{Physical\\Appeara-\\nce}} & \textbf{\shortstack[c]{Race /\\Color}} & \textbf{\shortstack[c]{Religion}} & \textbf{\shortstack[c]{Sexual\\Orienta-\\tion}} & \textbf{\shortstack[c]{Socioec-\\onomic}} & \textbf{\shortstack[c]{Overall}} \\
\midrule
0 & 0.3407 & 0.5455 & 0.5077 & 0.5687 & 0.3843 & 0.4306 & 0.5295 & 0.2072 & 0.2688 & 0.4737 & 0.4609 \\
1000 & 0.4615 & 0.4545 & 0.5385 & 0.5750 & 0.3009 & 0.5278 & 0.4528 & 0.2703 & 0.2258 & 0.4895 & 0.4431 \\
2000 & 0.3516 & 0.6364 & 0.5231 & 0.5875 & 0.3657 & 0.5278 & 0.5020 & 0.2793 & 0.2796 & 0.5789 & 0.4770 \\
3000 & 0.3956 & 0.4545 & 0.6615 & 0.6094 & 0.4028 & 0.6250 & 0.5079 & 0.2973 & 0.3441 & 0.5895 & 0.5045 \\
4000 & 0.4396 & 0.3636 & 0.6923 & 0.5906 & 0.4167 & 0.6111 & 0.5354 & 0.3423 & 0.3978 & 0.6053 & 0.5212 \\
5000 & 0.4505 & 0.3636 & 0.7538 & 0.6156 & 0.4028 & 0.6806 & 0.5354 & 0.4054 & 0.4086 & 0.6474 & 0.5397 \\
6000 & 0.4505 & 0.4545 & 0.7538 & 0.6000 & 0.4491 & 0.5972 & 0.5472 & 0.4144 & 0.3978 & 0.6895 & 0.5480 \\
7000 & 0.4945 & 0.3636 & 0.6923 & 0.6125 & 0.4676 & 0.6111 & 0.5453 & 0.3333 & 0.3871 & 0.6526 & 0.5420 \\
8000 & 0.4396 & 0.4545 & 0.7077 & 0.6219 & 0.4630 & 0.5972 & 0.5709 & 0.3784 & 0.4194 & 0.6474 & 0.5528 \\
9000 & 0.4615 & 0.4545 & 0.6923 & 0.6094 & 0.4815 & 0.5694 & 0.5295 & 0.4054 & 0.3656 & 0.6684 & 0.5408 \\
10000 & 0.5165 & 0.4545 & 0.7846 & 0.6000 & 0.4907 & 0.5972 & 0.5748 & 0.3514 & 0.4301 & 0.6421 & 0.5587 \\
11000 & 0.4286 & 0.5455 & 0.7385 & 0.6031 & 0.4769 & 0.5972 & 0.5551 & 0.3784 & 0.4086 & 0.6947 & 0.5522 \\
12000 & 0.4725 & 0.4545 & 0.7538 & 0.6219 & 0.4861 & 0.6250 & 0.5276 & 0.3694 & 0.3871 & 0.6947 & 0.5504 \\
13000 & 0.5165 & 0.4545 & 0.7692 & 0.6281 & 0.4815 & 0.6528 & 0.5807 & 0.3964 & 0.4086 & 0.6737 & 0.5719 \\
14000 & 0.5055 & 0.3636 & 0.7846 & 0.5969 & 0.4861 & 0.6806 & 0.5709 & 0.3604 & 0.3871 & 0.6632 & 0.5593 \\
15000 & 0.4286 & 0.4545 & 0.7692 & 0.6188 & 0.5046 & 0.6389 & 0.5807 & 0.4054 & 0.4301 & 0.6895 & 0.5713 \\
16000 & 0.4945 & 0.6364 & 0.7846 & 0.6000 & 0.5000 & 0.5694 & 0.5728 & 0.4234 & 0.3548 & 0.7000 & 0.5653 \\
17000 & 0.5275 & 0.5455 & 0.7538 & 0.6375 & 0.4861 & 0.6250 & 0.5492 & 0.3604 & 0.3763 & 0.6947 & 0.5623 \\
18000 & 0.5165 & 0.6364 & 0.7385 & 0.5844 & 0.5046 & 0.5833 & 0.5610 & 0.4505 & 0.3011 & 0.6789 & 0.5558 \\
19000 & 0.5275 & 0.4545 & 0.7538 & 0.6406 & 0.5185 & 0.6389 & 0.5787 & 0.3423 & 0.3871 & 0.6632 & 0.5719 \\
20000 & 0.5604 & 0.5455 & 0.7385 & 0.6344 & 0.5093 & 0.5972 & 0.5709 & 0.4144 & 0.3871 & 0.7105 & 0.5772 \\
21000 & 0.5714 & 0.5455 & 0.7538 & 0.6375 & 0.5093 & 0.6111 & 0.5689 & 0.3874 & 0.3656 & 0.7053 & 0.5754 \\
22000 & 0.4505 & 0.5455 & 0.7231 & 0.6094 & 0.4815 & 0.5694 & 0.5787 & 0.4054 & 0.3441 & 0.6474 & 0.5534 \\
23000 & 0.5934 & 0.6364 & 0.7692 & 0.6062 & 0.5231 & 0.6389 & 0.5728 & 0.4414 & 0.4194 & 0.7053 & 0.5826 \\
24000 & 0.5165 & 0.5455 & 0.7385 & 0.6281 & 0.5093 & 0.5972 & 0.5807 & 0.4234 & 0.3656 & 0.6684 & 0.5713 \\
25000 & 0.5055 & 0.6364 & 0.7538 & 0.6312 & 0.4907 & 0.6528 & 0.5906 & 0.4054 & 0.3656 & 0.6421 & 0.5713 \\
26000 & 0.5275 & 0.6364 & 0.7692 & 0.6156 & 0.5139 & 0.6528 & 0.6063 & 0.3964 & 0.3656 & 0.6789 & 0.5814 \\
27000 & 0.5275 & 0.6364 & 0.8000 & 0.6281 & 0.5046 & 0.6806 & 0.6122 & 0.4234 & 0.4194 & 0.6895 & 0.5927 \\
28000 & 0.5604 & 0.5455 & 0.7692 & 0.5906 & 0.4861 & 0.6667 & 0.5630 & 0.4144 & 0.4194 & 0.6632 & 0.5641 \\
29000 & 0.4835 & 0.6364 & 0.7692 & 0.6312 & 0.5000 & 0.6806 & 0.5768 & 0.4144 & 0.3763 & 0.6579 & 0.5719 \\
30000 & 0.5495 & 0.5455 & 0.7385 & 0.6438 & 0.4907 & 0.6389 & 0.5945 & 0.4414 & 0.3978 & 0.6421 & 0.5796 \\
31000 & 0.5275 & 0.5455 & 0.7692 & 0.6438 & 0.5000 & 0.6250 & 0.5965 & 0.3964 & 0.4086 & 0.7053 & 0.5856 \\
32000 & 0.4945 & 0.6364 & 0.7385 & 0.6094 & 0.4815 & 0.6528 & 0.5827 & 0.3874 & 0.3656 & 0.6632 & 0.5635 \\
33000 & 0.5604 & 0.5455 & 0.7692 & 0.6438 & 0.5046 & 0.6944 & 0.6024 & 0.4144 & 0.3656 & 0.7421 & 0.5957 \\
34000 & 0.5165 & 0.6364 & 0.7077 & 0.6312 & 0.4815 & 0.6806 & 0.5906 & 0.3874 & 0.3763 & 0.6789 & 0.5736 \\
35000 & 0.5604 & 0.6364 & 0.7846 & 0.6344 & 0.4722 & 0.6806 & 0.5768 & 0.3964 & 0.4086 & 0.7158 & 0.5808 \\
36000 & 0.5604 & 0.5455 & 0.7692 & 0.6469 & 0.4630 & 0.6667 & 0.5591 & 0.4054 & 0.3763 & 0.7053 & 0.5725 \\
\bottomrule
\end{tabular}
}
\caption{CrowS-Pairs bias scores of OLMo 2 1B checkpoints across bias types during the early-training stage.}
\label{tab:steps_olmo_2_1b_crows_pairs_early_training_bias}
\end{table*}
% ===== OLMo 2 1B early-training run: general knowledge =====
\begin{table*}[t]
\centering
\setlength{\tabcolsep}{3pt}
\scalebox{0.8}{
\begin{tabular}{cccc}
\toprule
\textbf{Step} & \textbf{ARC}$_{\mathbf{c}}$ & \textbf{LAMBADA} & \textbf{WinoGrande} \\
\midrule
0 & 0.2372 & 0.0000 & 0.5154 \\
1000 & 0.1749 & 0.0000 & 0.4949 \\
2000 & 0.1706 & 0.0014 & 0.5004 \\
3000 & 0.1843 & 0.0123 & 0.5114 \\
4000 & 0.2065 & 0.0225 & 0.5201 \\
5000 & 0.2125 & 0.0373 & 0.5170 \\
6000 & 0.2304 & 0.0456 & 0.5233 \\
7000 & 0.2210 & 0.0581 & 0.5028 \\
8000 & 0.2346 & 0.0606 & 0.5162 \\
9000 & 0.2449 & 0.0720 & 0.5359 \\
10000 & 0.2509 & 0.0721 & 0.5209 \\
11000 & 0.2389 & 0.0742 & 0.5201 \\
12000 & 0.2645 & 0.0766 & 0.5241 \\
13000 & 0.2679 & 0.0834 & 0.5320 \\
14000 & 0.2568 & 0.0899 & 0.5351 \\
15000 & 0.2628 & 0.0928 & 0.5114 \\
16000 & 0.2816 & 0.0927 & 0.5185 \\
17000 & 0.2688 & 0.0969 & 0.5343 \\
18000 & 0.2730 & 0.1001 & 0.5485 \\
19000 & 0.2713 & 0.1008 & 0.5620 \\
20000 & 0.2679 & 0.0961 & 0.5383 \\
21000 & 0.2756 & 0.1062 & 0.5399 \\
22000 & 0.2782 & 0.1001 & 0.5541 \\
23000 & 0.2918 & 0.1126 & 0.5549 \\
24000 & 0.2765 & 0.1075 & 0.5485 \\
25000 & 0.2867 & 0.1116 & 0.5612 \\
26000 & 0.2765 & 0.1071 & 0.5620 \\
27000 & 0.2969 & 0.1151 & 0.5478 \\
28000 & 0.2705 & 0.1108 & 0.5564 \\
29000 & 0.2858 & 0.1215 & 0.5643 \\
30000 & 0.2824 & 0.1204 & 0.5699 \\
31000 & 0.2875 & 0.1206 & 0.5635 \\
32000 & 0.2892 & 0.1240 & 0.5777 \\
33000 & 0.2867 & 0.1191 & 0.5762 \\
34000 & 0.2833 & 0.1196 & 0.5517 \\
35000 & 0.2884 & 0.1266 & 0.5509 \\
36000 & 0.2935 & 0.1208 & 0.5509 \\
\bottomrule
\end{tabular}
}
\caption{Performance of OLMo 2 1B checkpoints on general knowledge benchmarks during early-training.}
\label{tab:steps_olmo_2_1b_crows_pairs_early_training_general}
\end{table*}
% ===== OLMo 2 1B pre-training: CrowS-Pairs =====
\begin{table*}[t]
\centering
\setlength{\tabcolsep}{3pt}
\scalebox{0.8}{
\begin{tabular}{c>{\centering\arraybackslash}p{1.4cm}>{\centering\arraybackslash}p{1.4cm}>{\centering\arraybackslash}p{1.4cm}>{\centering\arraybackslash}p{1.4cm}>{\centering\arraybackslash}p{1.4cm}>{\centering\arraybackslash}p{1.4cm}>{\centering\arraybackslash}p{1.4cm}>{\centering\arraybackslash}p{1.4cm}>{\centering\arraybackslash}p{1.4cm}>{\centering\arraybackslash}p{1.4cm}|>{\centering\arraybackslash}p{1.4cm}}
\toprule
\textbf{Step} & \textbf{\shortstack[c]{Age}} & \textbf{\shortstack[c]{Other}} & \textbf{\shortstack[c]{Disabil-\\ity}} & \textbf{\shortstack[c]{Gender}} & \textbf{\shortstack[c]{Nationa-\\lity}} & \textbf{\shortstack[c]{Physical\\Appeara-\\nce}} & \textbf{\shortstack[c]{Race /\\Color}} & \textbf{\shortstack[c]{Religion}} & \textbf{\shortstack[c]{Sexual\\Orienta-\\tion}} & \textbf{\shortstack[c]{Socioec-\\onomic}} & \textbf{\shortstack[c]{Overall}} \\
\midrule
0 & 0.3407 & 0.5455 & 0.5231 & 0.5687 & 0.3843 & 0.4306 & 0.5256 & 0.2072 & 0.2796 & 0.4737 & 0.4609 \\
300 & 0.3736 & 0.4545 & 0.5077 & 0.5844 & 0.3472 & 0.4306 & 0.4291 & 0.2252 & 0.2366 & 0.5158 & 0.4341 \\
10000 & 0.4945 & 0.5455 & 0.7385 & 0.5969 & 0.4630 & 0.5972 & 0.5551 & 0.3423 & 0.4086 & 0.6789 & 0.5486 \\
20000 & 0.4835 & 0.6364 & 0.7538 & 0.6281 & 0.4630 & 0.5417 & 0.5610 & 0.3694 & 0.4516 & 0.6526 & 0.5558 \\
23100 & 0.5275 & 0.5455 & 0.7231 & 0.6406 & 0.4444 & 0.6250 & 0.5453 & 0.3874 & 0.4194 & 0.6737 & 0.5569 \\
30000 & 0.5385 & 0.5455 & 0.7538 & 0.6094 & 0.4815 & 0.6111 & 0.5866 & 0.4234 & 0.4409 & 0.6421 & 0.5695 \\
40000 & 0.6374 & 0.6364 & 0.7538 & 0.6219 & 0.4769 & 0.5833 & 0.6161 & 0.4144 & 0.3871 & 0.6632 & 0.5838 \\
50000 & 0.5604 & 0.6364 & 0.7385 & 0.6375 & 0.4815 & 0.5972 & 0.6063 & 0.3694 & 0.3763 & 0.6526 & 0.5754 \\
60000 & 0.5824 & 0.6364 & 0.7538 & 0.6188 & 0.5000 & 0.5972 & 0.5610 & 0.3964 & 0.4194 & 0.6895 & 0.5707 \\
66200 & 0.5824 & 0.5455 & 0.7538 & 0.6656 & 0.5046 & 0.6250 & 0.5748 & 0.4234 & 0.4301 & 0.6789 & 0.5862 \\
70000 & 0.5055 & 0.6364 & 0.7538 & 0.6656 & 0.4815 & 0.5972 & 0.6181 & 0.4144 & 0.3871 & 0.6684 & 0.5874 \\
80000 & 0.5385 & 0.5455 & 0.7538 & 0.6375 & 0.5139 & 0.6944 & 0.5709 & 0.4144 & 0.4086 & 0.6737 & 0.5790 \\
90000 & 0.5385 & 0.6364 & 0.7077 & 0.6312 & 0.5231 & 0.6250 & 0.6083 & 0.3964 & 0.3871 & 0.6526 & 0.5814 \\
100000 & 0.5385 & 0.5455 & 0.7846 & 0.6500 & 0.5000 & 0.6389 & 0.5925 & 0.4414 & 0.3656 & 0.6263 & 0.5790 \\
110000 & 0.5275 & 0.6364 & 0.7231 & 0.6312 & 0.5139 & 0.6389 & 0.5807 & 0.3874 & 0.4301 & 0.6632 & 0.5754 \\
120000 & 0.5385 & 0.6364 & 0.7538 & 0.6125 & 0.4861 & 0.6111 & 0.5945 & 0.4324 & 0.4194 & 0.6789 & 0.5772 \\
130000 & 0.5714 & 0.6364 & 0.7538 & 0.6344 & 0.5093 & 0.6667 & 0.5846 & 0.4324 & 0.4301 & 0.6316 & 0.5808 \\
140000 & 0.5055 & 0.6364 & 0.7538 & 0.6469 & 0.5046 & 0.7083 & 0.5925 & 0.4324 & 0.3656 & 0.6105 & 0.5772 \\
150000 & 0.5495 & 0.6364 & 0.7385 & 0.6188 & 0.4722 & 0.7083 & 0.5787 & 0.4234 & 0.3871 & 0.6842 & 0.5742 \\
160000 & 0.5824 & 0.5455 & 0.7538 & 0.6312 & 0.4861 & 0.6528 & 0.5925 & 0.3964 & 0.3871 & 0.6579 & 0.5772 \\
240000 & 0.5275 & 0.5455 & 0.7692 & 0.6312 & 0.5093 & 0.7083 & 0.5807 & 0.3874 & 0.3871 & 0.6737 & 0.5778 \\
330000 & 0.5275 & 0.6364 & 0.7077 & 0.6312 & 0.5139 & 0.6528 & 0.5984 & 0.3784 & 0.3978 & 0.6632 & 0.5784 \\
420000 & 0.5495 & 0.6364 & 0.7846 & 0.6312 & 0.5370 & 0.6528 & 0.5925 & 0.4054 & 0.3763 & 0.7105 & 0.5897 \\
510000 & 0.5714 & 0.6364 & 0.7385 & 0.6094 & 0.5185 & 0.6806 & 0.6083 & 0.4144 & 0.4086 & 0.6632 & 0.5856 \\
600000 & 0.5714 & 0.5455 & 0.7538 & 0.6344 & 0.4907 & 0.6528 & 0.5965 & 0.3874 & 0.4516 & 0.6632 & 0.5826 \\
690000 & 0.6044 & 0.5455 & 0.7077 & 0.6062 & 0.5093 & 0.6806 & 0.5787 & 0.4054 & 0.4409 & 0.6947 & 0.5796 \\
780000 & 0.6154 & 0.6364 & 0.7385 & 0.6438 & 0.5509 & 0.6944 & 0.5807 & 0.4595 & 0.4516 & 0.7000 & 0.6005 \\
870000 & 0.5714 & 0.6364 & 0.7385 & 0.6250 & 0.5093 & 0.6528 & 0.6260 & 0.4234 & 0.4301 & 0.6789 & 0.5951 \\
960000 & 0.5934 & 0.6364 & 0.7385 & 0.6188 & 0.4815 & 0.6528 & 0.6004 & 0.4234 & 0.4409 & 0.6842 & 0.5850 \\
1050000 & 0.5824 & 0.5455 & 0.7231 & 0.6531 & 0.5324 & 0.6528 & 0.6339 & 0.4685 & 0.4194 & 0.7053 & 0.6106 \\
1140000 & 0.5714 & 0.6364 & 0.7692 & 0.6250 & 0.5139 & 0.6667 & 0.6063 & 0.3874 & 0.4194 & 0.7211 & 0.5933 \\
1230000 & 0.5165 & 0.6364 & 0.7538 & 0.6219 & 0.5093 & 0.6250 & 0.5827 & 0.4414 & 0.4301 & 0.7000 & 0.5814 \\
1320000 & 0.5495 & 0.5455 & 0.7231 & 0.6500 & 0.5278 & 0.6111 & 0.6122 & 0.3784 & 0.3978 & 0.6789 & 0.5891 \\
1410000 & 0.5714 & 0.6364 & 0.7538 & 0.6375 & 0.5231 & 0.5972 & 0.6004 & 0.4414 & 0.4409 & 0.6632 & 0.5897 \\
1500000 & 0.5934 & 0.3636 & 0.7692 & 0.6250 & 0.5509 & 0.6250 & 0.6083 & 0.4054 & 0.4194 & 0.7053 & 0.5957 \\
1590000 & 0.5495 & 0.5455 & 0.7385 & 0.6344 & 0.4954 & 0.6389 & 0.6339 & 0.4234 & 0.4301 & 0.6842 & 0.5957 \\
1680000 & 0.5934 & 0.6364 & 0.7538 & 0.6375 & 0.5093 & 0.6806 & 0.5945 & 0.4144 & 0.3978 & 0.6737 & 0.5880 \\
1770000 & 0.5604 & 0.6364 & 0.7846 & 0.6312 & 0.5324 & 0.6389 & 0.6181 & 0.3784 & 0.4624 & 0.6737 & 0.5957 \\
1860000 & 0.5604 & 0.6364 & 0.7692 & 0.6469 & 0.5185 & 0.6667 & 0.6142 & 0.4685 & 0.4194 & 0.7000 & 0.6029 \\
1907359 & 0.5714 & 0.6364 & 0.7538 & 0.6469 & 0.5324 & 0.6250 & 0.5945 & 0.4324 & 0.4409 & 0.7105 & 0.5969 \\
\bottomrule
\end{tabular}
}
\caption{CrowS-Pairs bias scores of OLMo 2 1B checkpoints across bias types during pre-training.}
\label{tab:steps_olmo_2_1b_crows_pairs_pre_training_bias}
\end{table*}
% ===== OLMo 2 1B pre-training: general knowledge =====
\begin{table*}[t]
\centering
\setlength{\tabcolsep}{3pt}
\scalebox{0.8}{
\begin{tabular}{cccc}
\toprule
\textbf{Step} & \textbf{ARC}$_{\mathbf{c}}$ & \textbf{LAMBADA} & \textbf{WinoGrande} \\
\midrule
0 & 0.2372 & 0.0000 & 0.5154 \\
300 & 0.2031 & 0.0000 & 0.4854 \\
10000 & 0.2551 & 0.0733 & 0.5059 \\
20000 & 0.2850 & 0.1081 & 0.5462 \\
23100 & 0.2901 & 0.1053 & 0.5430 \\
30000 & 0.3003 & 0.1265 & 0.5612 \\
40000 & 0.2807 & 0.1285 & 0.5667 \\
50000 & 0.2833 & 0.1324 & 0.5627 \\
60000 & 0.3003 & 0.1523 & 0.5596 \\
66200 & 0.2918 & 0.1341 & 0.5667 \\
70000 & 0.3012 & 0.1416 & 0.5706 \\
80000 & 0.3114 & 0.1468 & 0.5880 \\
90000 & 0.3020 & 0.1534 & 0.5967 \\
100000 & 0.2986 & 0.1539 & 0.6014 \\
110000 & 0.2986 & 0.1611 & 0.5912 \\
120000 & 0.3038 & 0.1532 & 0.5801 \\
130000 & 0.3200 & 0.1508 & 0.5856 \\
140000 & 0.3140 & 0.1650 & 0.5943 \\
150000 & 0.3302 & 0.1535 & 0.6164 \\
160000 & 0.3259 & 0.1574 & 0.6014 \\
240000 & 0.3217 & 0.1784 & 0.5872 \\
330000 & 0.3456 & 0.1809 & 0.6164 \\
420000 & 0.3456 & 0.1829 & 0.6006 \\
510000 & 0.3302 & 0.1893 & 0.6069 \\
600000 & 0.3234 & 0.1677 & 0.6148 \\
690000 & 0.3404 & 0.1853 & 0.6125 \\
780000 & 0.3464 & 0.1928 & 0.6148 \\
870000 & 0.3447 & 0.1881 & 0.6243 \\
960000 & 0.3558 & 0.1891 & 0.6385 \\
1050000 & 0.3413 & 0.1921 & 0.6306 \\
1140000 & 0.3575 & 0.1896 & 0.6133 \\
1230000 & 0.3464 & 0.1945 & 0.6488 \\
1320000 & 0.3575 & 0.2111 & 0.6306 \\
1410000 & 0.3609 & 0.1940 & 0.6330 \\
1500000 & 0.3481 & 0.1986 & 0.6290 \\
1590000 & 0.3601 & 0.1979 & 0.6330 \\
1680000 & 0.3754 & 0.1956 & 0.6385 \\
1770000 & 0.3643 & 0.1936 & 0.6354 \\
1860000 & 0.3695 & 0.2119 & 0.6456 \\
1907359 & 0.3703 & 0.2099 & 0.6567 \\
\bottomrule
\end{tabular}
}
\caption{Performance of OLMo 2 1B checkpoints on general knowledge benchmarks during pre-training.}
\label{tab:steps_olmo_2_1b_crows_pairs_pre_training_general}
\end{table*}
% ===== OLMo 2 1B mid-training: CrowS-Pairs =====
\begin{table*}[t]
\centering
\setlength{\tabcolsep}{3pt}
\scalebox{0.8}{
\begin{tabular}{c>{\centering\arraybackslash}p{1.4cm}>{\centering\arraybackslash}p{1.4cm}>{\centering\arraybackslash}p{1.4cm}>{\centering\arraybackslash}p{1.4cm}>{\centering\arraybackslash}p{1.4cm}>{\centering\arraybackslash}p{1.4cm}>{\centering\arraybackslash}p{1.4cm}>{\centering\arraybackslash}p{1.4cm}>{\centering\arraybackslash}p{1.4cm}>{\centering\arraybackslash}p{1.4cm}|>{\centering\arraybackslash}p{1.4cm}}
\toprule
\textbf{Step} & \textbf{\shortstack[c]{Age}} & \textbf{\shortstack[c]{Other}} & \textbf{\shortstack[c]{Disabil-\\ity}} & \textbf{\shortstack[c]{Gender}} & \textbf{\shortstack[c]{Nationa-\\lity}} & \textbf{\shortstack[c]{Physical\\Appeara-\\nce}} & \textbf{\shortstack[c]{Race /\\Color}} & \textbf{\shortstack[c]{Religion}} & \textbf{\shortstack[c]{Sexual\\Orienta-\\tion}} & \textbf{\shortstack[c]{Socioec-\\onomic}} & \textbf{\shortstack[c]{Overall}} \\
\midrule
1000 & 0.5495 & 0.6364 & 0.7231 & 0.6250 & 0.5602 & 0.6250 & 0.5866 & 0.4324 & 0.4301 & 0.7474 & 0.5951 \\
2000 & 0.5934 & 0.6364 & 0.7846 & 0.6281 & 0.5324 & 0.6667 & 0.5807 & 0.4324 & 0.4409 & 0.6895 & 0.5909 \\
3000 & 0.5714 & 0.6364 & 0.7385 & 0.6469 & 0.5324 & 0.6389 & 0.5748 & 0.4414 & 0.4624 & 0.7105 & 0.5927 \\
4000 & 0.5824 & 0.6364 & 0.7692 & 0.6312 & 0.5278 & 0.6389 & 0.5925 & 0.4054 & 0.4731 & 0.6947 & 0.5927 \\
5000 & 0.5824 & 0.6364 & 0.7692 & 0.6375 & 0.5185 & 0.6250 & 0.5925 & 0.4414 & 0.4516 & 0.7000 & 0.5939 \\
6000 & 0.5604 & 0.6364 & 0.7538 & 0.6531 & 0.5463 & 0.6667 & 0.5846 & 0.4595 & 0.4409 & 0.7263 & 0.6017 \\
7000 & 0.6044 & 0.6364 & 0.7538 & 0.6500 & 0.5278 & 0.6806 & 0.5886 & 0.4324 & 0.4409 & 0.7053 & 0.5987 \\
8000 & 0.5604 & 0.6364 & 0.7385 & 0.6156 & 0.5093 & 0.6528 & 0.5846 & 0.4144 & 0.4301 & 0.7000 & 0.5820 \\
9000 & 0.5604 & 0.6364 & 0.7538 & 0.6188 & 0.5139 & 0.6667 & 0.5827 & 0.4505 & 0.4516 & 0.7211 & 0.5897 \\
10000 & 0.5824 & 0.6364 & 0.7692 & 0.6344 & 0.5185 & 0.6250 & 0.5984 & 0.4505 & 0.4624 & 0.6947 & 0.5957 \\
11000 & 0.5495 & 0.5455 & 0.7692 & 0.6312 & 0.5324 & 0.6389 & 0.6024 & 0.4414 & 0.4624 & 0.6842 & 0.5945 \\
12000 & 0.5934 & 0.6364 & 0.7538 & 0.6312 & 0.5324 & 0.6806 & 0.5768 & 0.4414 & 0.4946 & 0.6947 & 0.5939 \\
13000 & 0.5714 & 0.6364 & 0.7385 & 0.6250 & 0.5185 & 0.6806 & 0.6024 & 0.4324 & 0.4516 & 0.6789 & 0.5921 \\
14000 & 0.5604 & 0.6364 & 0.7692 & 0.6312 & 0.4954 & 0.7083 & 0.6043 & 0.4505 & 0.4516 & 0.7158 & 0.5981 \\
15000 & 0.6154 & 0.6364 & 0.7692 & 0.6375 & 0.5278 & 0.6667 & 0.5866 & 0.4505 & 0.5054 & 0.7000 & 0.6005 \\
16000 & 0.6044 & 0.6364 & 0.7538 & 0.6406 & 0.5046 & 0.7083 & 0.6063 & 0.4234 & 0.4946 & 0.7000 & 0.6023 \\
17000 & 0.5934 & 0.6364 & 0.7538 & 0.6219 & 0.5139 & 0.6528 & 0.6043 & 0.4414 & 0.4839 & 0.6842 & 0.5951 \\
18000 & 0.5824 & 0.6364 & 0.7846 & 0.6281 & 0.5185 & 0.6806 & 0.6063 & 0.4414 & 0.4731 & 0.6789 & 0.5981 \\
19000 & 0.5604 & 0.6364 & 0.7538 & 0.6125 & 0.5278 & 0.6944 & 0.6102 & 0.4595 & 0.5054 & 0.6632 & 0.5969 \\
20000 & 0.5824 & 0.6364 & 0.7692 & 0.6156 & 0.5370 & 0.7083 & 0.6201 & 0.4054 & 0.5161 & 0.6789 & 0.6029 \\
21000 & 0.5934 & 0.6364 & 0.7385 & 0.6344 & 0.5324 & 0.6806 & 0.6043 & 0.4505 & 0.4624 & 0.7105 & 0.6029 \\
22000 & 0.5934 & 0.6364 & 0.7385 & 0.6250 & 0.5324 & 0.6944 & 0.6004 & 0.4144 & 0.4839 & 0.6947 & 0.5975 \\
23000 & 0.6044 & 0.6364 & 0.7538 & 0.6156 & 0.5278 & 0.6806 & 0.6004 & 0.4324 & 0.4839 & 0.7053 & 0.5981 \\
23852 & 0.6044 & 0.6364 & 0.7231 & 0.6219 & 0.5324 & 0.6806 & 0.6004 & 0.4324 & 0.4839 & 0.6947 & 0.5975 \\
main & 0.5714 & 0.6364 & 0.7846 & 0.6125 & 0.5463 & 0.6111 & 0.5787 & 0.4685 & 0.4301 & 0.6947 & 0.5880 \\
\bottomrule
\end{tabular}
}
\caption{CrowS-Pairs bias scores of OLMo 2 1B checkpoints across bias types during mid-training.}
\label{tab:steps_olmo_2_1b_crows_pairs_mid_training_bias}
\end{table*}
% ===== OLMo 2 1B mid-training: general knowledge =====
\begin{table*}[t]
\centering
\setlength{\tabcolsep}{3pt}
\scalebox{0.8}{
\begin{tabular}{cccc}
\toprule
\textbf{Step} & \textbf{ARC}$_{\mathbf{c}}$ & \textbf{LAMBADA} & \textbf{WinoGrande} \\
\midrule
1000 & 0.3874 & 0.2008 & 0.6598 \\
2000 & 0.3823 & 0.1954 & 0.6527 \\
3000 & 0.3925 & 0.1913 & 0.6456 \\
4000 & 0.3891 & 0.1882 & 0.6480 \\
5000 & 0.3882 & 0.1794 & 0.6322 \\
6000 & 0.3925 & 0.1923 & 0.6290 \\
7000 & 0.3976 & 0.1816 & 0.6496 \\
8000 & 0.3959 & 0.1759 & 0.6330 \\
9000 & 0.3891 & 0.1775 & 0.6417 \\
10000 & 0.3968 & 0.1796 & 0.6440 \\
11000 & 0.3746 & 0.1826 & 0.6354 \\
12000 & 0.3908 & 0.1818 & 0.6369 \\
13000 & 0.3942 & 0.1799 & 0.6472 \\
14000 & 0.3848 & 0.1803 & 0.6354 \\
15000 & 0.3985 & 0.1832 & 0.6346 \\
16000 & 0.3985 & 0.1804 & 0.6409 \\
17000 & 0.3908 & 0.1831 & 0.6488 \\
18000 & 0.3951 & 0.1849 & 0.6488 \\
19000 & 0.3908 & 0.1859 & 0.6401 \\
20000 & 0.3968 & 0.1805 & 0.6417 \\
21000 & 0.3891 & 0.1876 & 0.6440 \\
22000 & 0.3951 & 0.1865 & 0.6401 \\
23000 & 0.3951 & 0.1854 & 0.6440 \\
23852 & 0.3908 & 0.1849 & 0.6409 \\
main & 0.3848 & 0.1927 & 0.6440 \\
\bottomrule
\end{tabular}
}
\caption{Performance of OLMo 2 1B checkpoints on general knowledge benchmarks during mid-training.}
\label{tab:steps_olmo_2_1b_crows_pairs_mid_training_general}
\end{table*}
% ===== OLMo 2 1B post-training: CrowS-Pairs =====
\begin{table*}[t]
\centering
\setlength{\tabcolsep}{3pt}
\scalebox{0.8}{
\begin{tabular}{cc>{\centering\arraybackslash}p{1.4cm}>{\centering\arraybackslash}p{1.4cm}>{\centering\arraybackslash}p{1.4cm}>{\centering\arraybackslash}p{1.4cm}>{\centering\arraybackslash}p{1.4cm}>{\centering\arraybackslash}p{1.4cm}>{\centering\arraybackslash}p{1.4cm}>{\centering\arraybackslash}p{1.4cm}>{\centering\arraybackslash}p{1.4cm}>{\centering\arraybackslash}p{1.4cm}|>{\centering\arraybackslash}p{1.4cm}}
\toprule
\textbf{Sub-stage} & \textbf{Step} & \textbf{\shortstack[c]{Age}} & \textbf{\shortstack[c]{Other}} & \textbf{\shortstack[c]{Disabil-\\ity}} & \textbf{\shortstack[c]{Gender}} & \textbf{\shortstack[c]{Nationa-\\lity}} & \textbf{\shortstack[c]{Physical\\Appeara-\\nce}} & \textbf{\shortstack[c]{Race /\\Color}} & \textbf{\shortstack[c]{Religion}} & \textbf{\shortstack[c]{Sexual\\Orienta-\\tion}} & \textbf{\shortstack[c]{Socioec-\\onomic}} & \textbf{\shortstack[c]{Overall}} \\
\midrule
SFT & - & 0.6044 & 0.6364 & 0.6923 & 0.6469 & 0.5648 & 0.6667 & 0.5669 & 0.4144 & 0.4086 & 0.7105 & 0.5909 \\
DPO & - & 0.6044 & 0.6364 & 0.7231 & 0.6594 & 0.5787 & 0.7083 & 0.5630 & 0.4324 & 0.4194 & 0.7158 & 0.5993 \\
RLVR1 & 200 & 0.6044 & 0.6364 & 0.7231 & 0.6562 & 0.5833 & 0.6944 & 0.5610 & 0.4505 & 0.4086 & 0.7105 & 0.5981 \\
RLVR1 & 400 & 0.6044 & 0.6364 & 0.7231 & 0.6500 & 0.5694 & 0.7083 & 0.5630 & 0.4324 & 0.4086 & 0.7105 & 0.5951 \\
RLVR1 & 600 & 0.6044 & 0.6364 & 0.7231 & 0.6531 & 0.5694 & 0.7083 & 0.5630 & 0.4144 & 0.4086 & 0.7158 & 0.5951 \\
RLVR1 & 800 & 0.6044 & 0.6364 & 0.7231 & 0.6531 & 0.5741 & 0.7083 & 0.5531 & 0.4324 & 0.4086 & 0.7158 & 0.5939 \\
RLVR1 & 1000 & 0.6044 & 0.6364 & 0.7231 & 0.6469 & 0.5741 & 0.7083 & 0.5630 & 0.4234 & 0.4086 & 0.7105 & 0.5939 \\
RLVR1 & 1200 & 0.6044 & 0.6364 & 0.7077 & 0.6500 & 0.5694 & 0.7083 & 0.5650 & 0.4144 & 0.4086 & 0.7105 & 0.5939 \\
RLVR1 & 1400 & 0.6044 & 0.6364 & 0.7077 & 0.6531 & 0.5787 & 0.6944 & 0.5610 & 0.4144 & 0.4086 & 0.7105 & 0.5939 \\
RLVR1 & 1600 & 0.6044 & 0.6364 & 0.7231 & 0.6531 & 0.5741 & 0.7222 & 0.5610 & 0.4054 & 0.4086 & 0.7158 & 0.5951 \\
RLVR1 & 1800 & 0.6044 & 0.6364 & 0.7231 & 0.6500 & 0.5741 & 0.6806 & 0.5669 & 0.4234 & 0.4086 & 0.7105 & 0.5951 \\
RLVR1 & 2000 & 0.6044 & 0.6364 & 0.7231 & 0.6500 & 0.5741 & 0.7083 & 0.5591 & 0.4144 & 0.4086 & 0.7105 & 0.5933 \\
RLVR1 & 2200 & 0.6044 & 0.6364 & 0.7231 & 0.6500 & 0.5787 & 0.7083 & 0.5610 & 0.4234 & 0.4086 & 0.7158 & 0.5957 \\
RLVR1 & 2400 & 0.6044 & 0.6364 & 0.7077 & 0.6500 & 0.5741 & 0.7083 & 0.5669 & 0.4144 & 0.4086 & 0.7000 & 0.5939 \\
RLVR1 & 2600 & 0.6044 & 0.6364 & 0.7077 & 0.6594 & 0.5694 & 0.6944 & 0.5610 & 0.4144 & 0.4086 & 0.7053 & 0.5933 \\
RLVR1 & main & 0.6044 & 0.6364 & 0.7231 & 0.6531 & 0.5741 & 0.7083 & 0.5630 & 0.4324 & 0.4086 & 0.7105 & 0.5963 \\
Instruct & 200 & 0.6044 & 0.6364 & 0.7077 & 0.6500 & 0.5741 & 0.6806 & 0.5650 & 0.4144 & 0.4086 & 0.7053 & 0.5927 \\
Instruct & 400 & 0.6044 & 0.6364 & 0.7077 & 0.6562 & 0.5787 & 0.6806 & 0.5591 & 0.4234 & 0.4086 & 0.7000 & 0.5927 \\
Instruct & 600 & 0.6044 & 0.6364 & 0.7231 & 0.6531 & 0.5787 & 0.7083 & 0.5669 & 0.4414 & 0.4086 & 0.7000 & 0.5975 \\
Instruct & 800 & 0.6044 & 0.6364 & 0.7231 & 0.6469 & 0.5741 & 0.7083 & 0.5650 & 0.4324 & 0.4086 & 0.7000 & 0.5945 \\
Instruct & 1000 & 0.6044 & 0.6364 & 0.7077 & 0.6625 & 0.5787 & 0.7083 & 0.5571 & 0.4414 & 0.4086 & 0.7000 & 0.5957 \\
Instruct & 1200 & 0.6044 & 0.6364 & 0.7077 & 0.6500 & 0.5741 & 0.7083 & 0.5630 & 0.4324 & 0.4086 & 0.7000 & 0.5939 \\
Instruct & 1400 & 0.6044 & 0.6364 & 0.7385 & 0.6594 & 0.5787 & 0.7222 & 0.5591 & 0.4414 & 0.4086 & 0.7105 & 0.5987 \\
Instruct & 1600 & 0.6044 & 0.6364 & 0.7077 & 0.6562 & 0.5787 & 0.7083 & 0.5591 & 0.4505 & 0.4086 & 0.7053 & 0.5963 \\
Instruct & 1800 & 0.6044 & 0.6364 & 0.7231 & 0.6500 & 0.5741 & 0.6944 & 0.5591 & 0.4324 & 0.4086 & 0.7053 & 0.5933 \\
Instruct & 2000 & 0.6044 & 0.6364 & 0.7077 & 0.6562 & 0.5787 & 0.7083 & 0.5669 & 0.4414 & 0.4086 & 0.7053 & 0.5981 \\
Instruct & 2200 & 0.6154 & 0.6364 & 0.7077 & 0.6625 & 0.5741 & 0.7222 & 0.5630 & 0.4414 & 0.4086 & 0.7053 & 0.5987 \\
Instruct & 2400 & 0.6044 & 0.6364 & 0.7077 & 0.6531 & 0.5741 & 0.7083 & 0.5650 & 0.4144 & 0.4086 & 0.7105 & 0.5951 \\
Instruct & 2600 & 0.6044 & 0.6364 & 0.7077 & 0.6562 & 0.5741 & 0.7083 & 0.5630 & 0.4414 & 0.4086 & 0.7105 & 0.5969 \\
Instruct & main & 0.6044 & 0.6364 & 0.7077 & 0.6562 & 0.5787 & 0.7083 & 0.5630 & 0.4414 & 0.4086 & 0.7053 & 0.5969 \\
\bottomrule
\end{tabular}
}
\caption{CrowS-Pairs bias scores of OLMo 2 1B checkpoints across bias types during post-training.}
\label{tab:steps_olmo_2_1b_crows_pairs_post_training_bias}
\end{table*}
% ===== OLMo 2 1B post-training: general knowledge =====
\begin{table*}[t]
\centering
\setlength{\tabcolsep}{3pt}
\scalebox{0.8}{
\begin{tabular}{ccccc}
\toprule
\textbf{Sub-stage} & \textbf{Step} & \textbf{ARC}$_{\mathbf{c}}$ & \textbf{LAMBADA} & \textbf{WinoGrande} \\
\midrule
SFT & - & 0.3601 & 0.1790 & 0.6267 \\
DPO & - & 0.4019 & 0.1610 & 0.6346 \\
RLVR1 & 200 & 0.4061 & 0.1593 & 0.6338 \\
RLVR1 & 400 & 0.4036 & 0.1589 & 0.6322 \\
RLVR1 & 600 & 0.4002 & 0.1604 & 0.6290 \\
RLVR1 & 800 & 0.3968 & 0.1608 & 0.6338 \\
RLVR1 & 1000 & 0.3985 & 0.1628 & 0.6354 \\
RLVR1 & 1200 & 0.3899 & 0.1618 & 0.6322 \\
RLVR1 & 1400 & 0.3951 & 0.1624 & 0.6338 \\
RLVR1 & 1600 & 0.3968 & 0.1628 & 0.6346 \\
RLVR1 & 1800 & 0.3933 & 0.1648 & 0.6338 \\
RLVR1 & 2000 & 0.3891 & 0.1626 & 0.6377 \\
RLVR1 & 2200 & 0.3925 & 0.1605 & 0.6361 \\
RLVR1 & 2400 & 0.3899 & 0.1578 & 0.6346 \\
RLVR1 & 2600 & 0.3959 & 0.1585 & 0.6322 \\
RLVR1 & main & 0.3959 & 0.1586 & 0.6354 \\
Instruct & 200 & 0.4010 & 0.1614 & 0.6401 \\
Instruct & 400 & 0.4036 & 0.1602 & 0.6393 \\
Instruct & 600 & 0.4010 & 0.1603 & 0.6425 \\
Instruct & 800 & 0.4061 & 0.1605 & 0.6385 \\
Instruct & 1000 & 0.4036 & 0.1599 & 0.6433 \\
Instruct & 1200 & 0.4044 & 0.1604 & 0.6385 \\
Instruct & 1400 & 0.4070 & 0.1605 & 0.6377 \\
Instruct & 1600 & 0.4019 & 0.1601 & 0.6417 \\
Instruct & 1800 & 0.4087 & 0.1608 & 0.6409 \\
Instruct & 2000 & 0.4104 & 0.1611 & 0.6401 \\
Instruct & 2200 & 0.4087 & 0.1619 & 0.6409 \\
Instruct & 2400 & 0.4087 & 0.1612 & 0.6393 \\
Instruct & 2600 & 0.4078 & 0.1624 & 0.6401 \\
Instruct & main & 0.4104 & 0.1611 & 0.6401 \\
\bottomrule
\end{tabular}
}
\caption{Performance of OLMo 2 1B checkpoints on general knowledge benchmarks during post-training.}
\label{tab:steps_olmo_2_1b_crows_pairs_post_training_general}
\end{table*}
% ===== Pythia 1B pre-training: CrowS-Pairs =====
\begin{table*}[t]
\centering
\setlength{\tabcolsep}{3pt}
\scalebox{0.8}{
\begin{tabular}{c>{\centering\arraybackslash}p{1.4cm}>{\centering\arraybackslash}p{1.4cm}>{\centering\arraybackslash}p{1.4cm}>{\centering\arraybackslash}p{1.4cm}>{\centering\arraybackslash}p{1.4cm}>{\centering\arraybackslash}p{1.4cm}>{\centering\arraybackslash}p{1.4cm}>{\centering\arraybackslash}p{1.4cm}>{\centering\arraybackslash}p{1.4cm}>{\centering\arraybackslash}p{1.4cm}|>{\centering\arraybackslash}p{1.4cm}}
\toprule
\textbf{Step} & \textbf{\shortstack[c]{Age}} & \textbf{\shortstack[c]{Other}} & \textbf{\shortstack[c]{Disabil-\\ity}} & \textbf{\shortstack[c]{Gender}} & \textbf{\shortstack[c]{Nationa-\\lity}} & \textbf{\shortstack[c]{Physical\\Appeara-\\nce}} & \textbf{\shortstack[c]{Race /\\Color}} & \textbf{\shortstack[c]{Religion}} & \textbf{\shortstack[c]{Sexual\\Orienta-\\tion}} & \textbf{\shortstack[c]{Socioec-\\onomic}} & \textbf{\shortstack[c]{Overall}} \\
\midrule
0 & 0.3297 & 0.3636 & 0.4462 & 0.5781 & 0.3287 & 0.5556 & 0.4390 & 0.2703 & 0.2151 & 0.5211 & 0.4359 \\
1 & 0.3297 & 0.3636 & 0.4462 & 0.5781 & 0.3287 & 0.5556 & 0.4390 & 0.2703 & 0.2151 & 0.5211 & 0.4359 \\
2 & 0.3297 & 0.3636 & 0.4462 & 0.5719 & 0.3287 & 0.5556 & 0.4390 & 0.2703 & 0.2151 & 0.5263 & 0.4353 \\
4 & 0.3297 & 0.3636 & 0.4462 & 0.5781 & 0.3241 & 0.5556 & 0.4370 & 0.2703 & 0.2151 & 0.5211 & 0.4347 \\
8 & 0.3407 & 0.3636 & 0.4462 & 0.5875 & 0.3333 & 0.5556 & 0.4173 & 0.2883 & 0.2151 & 0.5211 & 0.4335 \\
16 & 0.3297 & 0.3636 & 0.5077 & 0.5813 & 0.3519 & 0.5972 & 0.2992 & 0.2883 & 0.2043 & 0.5000 & 0.3995 \\
32 & 0.3297 & 0.3636 & 0.4923 & 0.5844 & 0.3657 & 0.5972 & 0.2776 & 0.2973 & 0.2043 & 0.4263 & 0.3870 \\
64 & 0.3516 & 0.2727 & 0.5538 & 0.5781 & 0.3148 & 0.5833 & 0.2618 & 0.2973 & 0.2151 & 0.5737 & 0.3942 \\
128 & 0.3626 & 0.3636 & 0.5385 & 0.5875 & 0.2917 & 0.5417 & 0.2854 & 0.2973 & 0.2043 & 0.6263 & 0.4043 \\
256 & 0.3736 & 0.4545 & 0.4769 & 0.5437 & 0.3148 & 0.5417 & 0.4449 & 0.2703 & 0.2258 & 0.5737 & 0.4395 \\
512 & 0.3736 & 0.5455 & 0.5692 & 0.5750 & 0.3194 & 0.5000 & 0.4449 & 0.2703 & 0.2366 & 0.5474 & 0.4460 \\
1000 & 0.3407 & 0.3636 & 0.6308 & 0.6156 & 0.3611 & 0.5278 & 0.4980 & 0.2432 & 0.2258 & 0.6211 & 0.4818 \\
2000 & 0.4066 & 0.6364 & 0.6615 & 0.6062 & 0.4120 & 0.5972 & 0.4606 & 0.3514 & 0.3441 & 0.6211 & 0.4985 \\
3000 & 0.3626 & 0.4545 & 0.6462 & 0.5813 & 0.3981 & 0.5833 & 0.4921 & 0.3333 & 0.3656 & 0.6316 & 0.4979 \\
4000 & 0.3736 & 0.6364 & 0.7231 & 0.5906 & 0.4213 & 0.5694 & 0.5394 & 0.2973 & 0.3978 & 0.6211 & 0.5194 \\
5000 & 0.3736 & 0.5455 & 0.7077 & 0.6000 & 0.4306 & 0.5417 & 0.4823 & 0.3604 & 0.3871 & 0.5947 & 0.5033 \\
6000 & 0.4176 & 0.5455 & 0.7385 & 0.5625 & 0.4444 & 0.5556 & 0.5079 & 0.3604 & 0.3656 & 0.5632 & 0.5051 \\
7000 & 0.4066 & 0.6364 & 0.7231 & 0.5656 & 0.4398 & 0.6111 & 0.5177 & 0.3694 & 0.4839 & 0.5579 & 0.5164 \\
8000 & 0.4176 & 0.5455 & 0.7538 & 0.5594 & 0.4537 & 0.6389 & 0.5157 & 0.3694 & 0.4516 & 0.5737 & 0.5188 \\
9000 & 0.4286 & 0.6364 & 0.7385 & 0.5844 & 0.4213 & 0.5972 & 0.4961 & 0.3964 & 0.3978 & 0.5579 & 0.5092 \\
11000 & 0.4615 & 0.4545 & 0.7231 & 0.6031 & 0.4167 & 0.6667 & 0.5118 & 0.4054 & 0.4086 & 0.6053 & 0.5265 \\
18000 & 0.4835 & 0.5455 & 0.7077 & 0.5875 & 0.4491 & 0.5972 & 0.5492 & 0.3874 & 0.4194 & 0.6105 & 0.5373 \\
25000 & 0.4176 & 0.4545 & 0.7385 & 0.5938 & 0.4352 & 0.6389 & 0.5295 & 0.3694 & 0.4194 & 0.6421 & 0.5319 \\
32000 & 0.4615 & 0.3636 & 0.7846 & 0.6062 & 0.4583 & 0.6111 & 0.5591 & 0.3694 & 0.4194 & 0.6368 & 0.5480 \\
39000 & 0.4396 & 0.4545 & 0.7385 & 0.6125 & 0.4722 & 0.5694 & 0.5492 & 0.3874 & 0.4194 & 0.6368 & 0.5450 \\
46000 & 0.4835 & 0.5455 & 0.7846 & 0.6281 & 0.4444 & 0.5694 & 0.5571 & 0.3514 & 0.3978 & 0.6421 & 0.5480 \\
53000 & 0.4945 & 0.5455 & 0.8154 & 0.5594 & 0.4676 & 0.6250 & 0.5669 & 0.3874 & 0.4194 & 0.6684 & 0.5522 \\
60000 & 0.4725 & 0.4545 & 0.7538 & 0.6000 & 0.4954 & 0.6250 & 0.5256 & 0.3964 & 0.4086 & 0.6579 & 0.5456 \\
67000 & 0.4835 & 0.5455 & 0.8000 & 0.5938 & 0.4954 & 0.6111 & 0.5650 & 0.3874 & 0.4194 & 0.6842 & 0.5617 \\
74000 & 0.4725 & 0.4545 & 0.7692 & 0.6281 & 0.4491 & 0.5972 & 0.5335 & 0.3874 & 0.4516 & 0.6368 & 0.5462 \\
81000 & 0.4835 & 0.4545 & 0.7692 & 0.5750 & 0.4537 & 0.5972 & 0.5630 & 0.3784 & 0.4194 & 0.7105 & 0.5522 \\
88000 & 0.5165 & 0.5455 & 0.8000 & 0.6000 & 0.4722 & 0.6250 & 0.5354 & 0.3874 & 0.4086 & 0.6789 & 0.5522 \\
95000 & 0.5165 & 0.6364 & 0.7538 & 0.6000 & 0.4769 & 0.6667 & 0.5276 & 0.4054 & 0.4624 & 0.7000 & 0.5575 \\
102000 & 0.4396 & 0.6364 & 0.8000 & 0.6094 & 0.4722 & 0.6667 & 0.5551 & 0.4414 & 0.4409 & 0.6684 & 0.5617 \\
109000 & 0.5385 & 0.6364 & 0.7846 & 0.5781 & 0.4722 & 0.6111 & 0.5807 & 0.3784 & 0.4409 & 0.6684 & 0.5623 \\
116000 & 0.5385 & 0.6364 & 0.7846 & 0.6094 & 0.4583 & 0.5972 & 0.5512 & 0.4324 & 0.4086 & 0.6158 & 0.5528 \\
123000 & 0.5165 & 0.6364 & 0.7692 & 0.5750 & 0.4676 & 0.6250 & 0.5512 & 0.4234 & 0.4409 & 0.6789 & 0.5552 \\
130000 & 0.5275 & 0.6364 & 0.7231 & 0.5750 & 0.4630 & 0.5972 & 0.5197 & 0.4144 & 0.4624 & 0.6632 & 0.5414 \\
137000 & 0.5165 & 0.6364 & 0.7385 & 0.5813 & 0.4398 & 0.6111 & 0.5197 & 0.4144 & 0.4409 & 0.6842 & 0.5414 \\
143000 & 0.5495 & 0.5455 & 0.7846 & 0.5844 & 0.4722 & 0.6528 & 0.5472 & 0.3964 & 0.4624 & 0.6842 & 0.5593 \\
\bottomrule
\end{tabular}
}
\caption{CrowS-Pairs bias scores of Pythia 1B checkpoints across bias types during pre-training.}
\label{tab:steps_pythia_1b_crows_pairs_pre_training_bias}
\end{table*}

% ===== Pythia 1B pre-training: general knowledge =====
\begin{table*}[t]
\centering
\setlength{\tabcolsep}{3pt}
\scalebox{0.8}{
\begin{tabular}{cccc}
\toprule
\textbf{Step} & \textbf{ARC}$_{\mathbf{c}}$ & \textbf{LAMBADA} & \textbf{WinoGrande} \\
\midrule
0 & 0.2150 & 0.0000 & 0.4854 \\
1 & 0.2150 & 0.0000 & 0.4854 \\
2 & 0.2159 & 0.0000 & 0.4830 \\
4 & 0.2150 & 0.0000 & 0.4838 \\
8 & 0.2022 & 0.0000 & 0.4870 \\
16 & 0.2150 & 0.0000 & 0.4846 \\
32 & 0.2014 & 0.0000 & 0.4846 \\
64 & 0.1971 & 0.0000 & 0.4957 \\
128 & 0.1894 & 0.0000 & 0.4901 \\
256 & 0.2031 & 0.0000 & 0.5028 \\
512 & 0.1894 & 0.0000 & 0.5004 \\
1000 & 0.1826 & 0.0007 & 0.4941 \\
2000 & 0.1741 & 0.0098 & 0.5067 \\
3000 & 0.1894 & 0.0202 & 0.5241 \\
4000 & 0.1817 & 0.0294 & 0.5185 \\
5000 & 0.1792 & 0.0325 & 0.5233 \\
6000 & 0.1928 & 0.0431 & 0.5209 \\
7000 & 0.1843 & 0.0448 & 0.5185 \\
8000 & 0.1945 & 0.0464 & 0.5241 \\
9000 & 0.1937 & 0.0517 & 0.5209 \\
11000 & 0.2005 & 0.0566 & 0.5107 \\
18000 & 0.2065 & 0.0722 & 0.5233 \\
25000 & 0.2090 & 0.0833 & 0.5154 \\
32000 & 0.2261 & 0.0927 & 0.5304 \\
39000 & 0.2201 & 0.0884 & 0.5185 \\
46000 & 0.2253 & 0.0978 & 0.5107 \\
53000 & 0.2261 & 0.1012 & 0.5170 \\
60000 & 0.2278 & 0.1078 & 0.5146 \\
67000 & 0.2440 & 0.1058 & 0.5154 \\
74000 & 0.2338 & 0.1114 & 0.5217 \\
81000 & 0.2381 & 0.1096 & 0.5241 \\
88000 & 0.2389 & 0.1185 & 0.5178 \\
95000 & 0.2372 & 0.1165 & 0.5122 \\
102000 & 0.2338 & 0.1279 & 0.5249 \\
109000 & 0.2406 & 0.1246 & 0.5304 \\
116000 & 0.2406 & 0.1234 & 0.5383 \\
123000 & 0.2372 & 0.1304 & 0.5249 \\
130000 & 0.2363 & 0.1265 & 0.5209 \\
137000 & 0.2415 & 0.1299 & 0.5249 \\
143000 & 0.2432 & 0.1264 & 0.5359 \\
\bottomrule
\end{tabular}
}
\caption{Performance of Pythia 1B checkpoints on general knowledge benchmarks during pre-training.}
\label{tab:steps_pythia_1b_crows_pairs_pre_training_general}
\end{table*}

\the\appBfloats
\clearpage
\twocolumn

\section{Experimental Details for RQ2}
\label{app:rq2}
\subsection{Pretraining Configuration Details}
\label{app:rq2_pretraining_config_details}

We provide more details on the experiments for different pre-training configurations in Section~\ref{sec:pretraining_config_setup}. 
% To study the effect of random seeds, we use nine 410M PolyPythias models with seeds 1--9 and EleutherAI
% /pythia-410m with seed 1234. 
% For the data-composition setting, we use all 25 1B DataDecide models with different pretraining data recipes. 
% Model repository links are provided below for reproducibility.

\paragraph{Pythia seed models.}
The experiments of PolyPythias models \citep{polypythias} are designed to study variation across random seeds while holding model architecture, corpus, and training hyperparameters fixed. 
The seed variation changes the pre-training initial conditions, including parameter initialization and training data ordering. 
We use the 410M PolyPythias\footnote{\href{https://huggingface.co/collections/EleutherAI/polypythias}{PolyPythias model collection}.} models with seeds 1--9, together with the standard pythia-410m\footnote{\url{https://huggingface.co/EleutherAI/pythia-410m}} trained with seed 1234.
Table~\ref{tab:pythia_410m_config} summarizes the shared pre-training configurations of the Pythia 410M seed models.

\begin{table}[htbp]
\centering
\small
\begin{tabular}{ll}
\toprule
\textbf{Configuration Key} & \textbf{Value} \\
\midrule
Model suite & PolyPythias / Pythia \\
Model size & 410M \\
Non-embedding parameters & 302,311,424 \\
Number of layers & 24 \\
Hidden dimension & 1,024 \\
Number of attention heads & 16 \\
Pretraining data & The Pile \\
Training tokens & 300B \\
Batch size & 2M tokens \\
Learning rate & 3.0 $\times$ 10$^{-4}$ \\
Sequence length & 2,048 \\
Number of checkpoints & 154 \\
Evaluated checkpoints & 40 \\
Seeds used & 1--9, 1234 (original) \\
\bottomrule
\end{tabular}
\caption{Shared pre-training configurations of the Pythia seed models.}
\label{tab:pythia_410m_config}
\end{table}

\paragraph{DataDecide models.}
% \label{app:datadecide_models}

The DataDecide experiments are designed to study variation across pre-training data recipes while holding model architecture and training budget fixed. 
We use 25 DataDecide 1B models, corresponding to the data recipes analyzed in Section~\ref{sec:pretraining_config_setup}.
Table~\ref{tab:datadecide_1b_config} reports the shared training configurations of the DataDecide 1B models. 
% This shared training configuration indicates that the DataDecide comparison focuses on data-recipe variation under a fixed model size and training budget.
Table~\ref{tab:datadecide_data_mixture_models} lists the evaluated models and their pretraining data recipes.
Detailed descriptions of the source corpora and data recipes are available in the DataDecide model cards linked in the table.

\begin{table*}[t]
\centering
\small
\begin{tabular}{ll}
\toprule
\textbf{Data Recipe Groups} & \textbf{Data Recipes and Models} \\
\midrule
\textbf{Cross-Corpus Baselines}
& \href{https://huggingface.co/allenai/DataDecide-dolma1_7-1B}{100\% Dolma1.7} \\
& \href{https://huggingface.co/allenai/DataDecide-c4-1B}{C4} \\
& \href{https://huggingface.co/allenai/DataDecide-fineweb-pro-1B}{FineWeb-Pro} \\
& \href{https://huggingface.co/allenai/DataDecide-fineweb-edu-1B}{FineWeb-Edu} \\
& \href{https://huggingface.co/allenai/DataDecide-falcon-1B}{Falcon} \\
& \href{https://huggingface.co/allenai/DataDecide-dclm-baseline-1B}{100\% DCLM-Baseline} \\
\midrule
\textbf{Dolma Source Ablations and Version Reference}
& \href{https://huggingface.co/allenai/DataDecide-dolma1_7-no-code-1B}{Dolma1.7 (no code)} \\
& \href{https://huggingface.co/allenai/DataDecide-dolma1_7-no-math-code-1B}{Dolma1.7 (no math, code)} \\
& \href{https://huggingface.co/allenai/DataDecide-dolma1_7-no-reddit-1B}{Dolma1.7 (no Reddit)} \\
& \href{https://huggingface.co/allenai/DataDecide-dolma1_7-no-flan-1B}{Dolma1.7 (no Flan)} \\
& \href{https://huggingface.co/allenai/DataDecide-dolma1_6plus-1B}{Dolma1.6++} \\
\midrule
\textbf{DCLM-Dolma Mixture}
& \href{https://huggingface.co/allenai/DataDecide-dclm-baseline-25p-dolma1.7-75p-1B}{25\% DCLM-Baseline + 75\% Dolma1.7} \\
& \href{https://huggingface.co/allenai/DataDecide-dclm-baseline-50p-dolma1.7-50p-1B}{50\% DCLM-Baseline + 50\% Dolma1.7} \\
& \href{https://huggingface.co/allenai/DataDecide-dclm-baseline-75p-dolma1.7-25p-1B}{75\% DCLM-Baseline + 25\% Dolma1.7} \\
\midrule
\textbf{Falcon and Falcon+CC Filtering}
& \href{https://huggingface.co/allenai/DataDecide-falcon-and-cc-1B}{Falcon+CC} \\
& \href{https://huggingface.co/allenai/DataDecide-falcon-and-cc-qc-10p-1B}{Falcon+CC QC 10\%} \\
& \href{https://huggingface.co/allenai/DataDecide-falcon-and-cc-qc-20p-1B}{Falcon+CC QC 20\%} \\
& \href{https://huggingface.co/allenai/DataDecide-falcon-and-cc-qc-orig-10p-1B}{Falcon+CC QC Orig 10\%} \\
& \href{https://huggingface.co/allenai/DataDecide-falcon-and-cc-qc-tulu-10p-1B}{Falcon+CC QC Tulu 10\%} \\
\midrule
\textbf{DCLM Classifier Quality Filtering}
& \href{https://huggingface.co/allenai/DataDecide-dclm-baseline-qc-10p-1B}{DCLM-Baseline QC 10\%} \\
& \href{https://huggingface.co/allenai/DataDecide-dclm-baseline-qc-20p-1B}{DCLM-Baseline QC 20\%} \\
\midrule
\textbf{DCLM FineWeb-Edu Classifier Quality Filtering}
& \href{https://huggingface.co/allenai/DataDecide-dclm-baseline-qc-7p-fw2-1B}{DCLM-Baseline QC 7\% FW2} \\
& \href{https://huggingface.co/allenai/DataDecide-dclm-baseline-qc-7p-fw3-1B}{DCLM-Baseline QC 7\% FW3} \\
& \href{https://huggingface.co/allenai/DataDecide-dclm-baseline-qc-fw-3p-1B}{DCLM-Baseline QC FW 3\%} \\
& \href{https://huggingface.co/allenai/DataDecide-dclm-baseline-qc-fw-10p-1B}{DCLM-Baseline QC FW 10\%} \\
\bottomrule
\end{tabular}
\caption{DataDecide 1B models grouped by pre-training data recipes. Model names are linked to their corresponding Hugging Face repositories. Models shared across data recipe groups are listed only once.}
\label{tab:datadecide_data_mixture_models}
\end{table*}

%\paragraph{Evaluation.}
% \label{app:pretraining_config_figure_construction}

\paragraph{Checkpoints used in the analysis.}
For PolyPythias, we evaluate each of the ten Pythia 410M models at 40 checkpoints from \texttt{step0} to \texttt{step143000}. This subset includes the dense early checkpoints \texttt{step0}, \texttt{step1}, \texttt{step2}, \texttt{step4}, \texttt{step8}, \texttt{step16}, \texttt{step32}, \texttt{step64}, \texttt{step128}, \texttt{step256}, \texttt{step512}, and \texttt{step1000}; every 1,000 steps from \texttt{step2000} to \texttt{step9000}; and then every 7,000 steps from \texttt{step11000} to \texttt{step137000}, plus the final checkpoint \texttt{step143000}. For the DataDecide data-recipe comparisons, we use the final checkpoint, \texttt{step69369-seed-default}, of each of the 25 1B models. Thus, the Pythia analyses reflect variation across ten random seeds, whereas the DataDecide analyses compare bias across 25 data recipes on the default seed branch.

\begin{table}[H]
\centering
\small
\begin{tabular}{ll}
\toprule
\textbf{Configuration Key} & \textbf{Value} \\
\midrule
Model suite & DataDecide \\
Architecture & OLMo-style model \\
Model size & 1B \\
Non-embedding parameters & 1,176.8M \\
Number of layers & 16 \\
Hidden dimension & 2,048 \\
Number of attention heads & 16 \\
Sequence length & 2,048 \\
Vocabulary size & 50,280 \\
Embedding size & 50,304 \\
Batch size & 704 sequences \\
Learning rate & $2.1 \times 10^{-3}$ \\
Training steps & 69,369 \\
Training tokens & 100B \\
\bottomrule
\end{tabular}
\caption{Shared pre-training configurations of the DataDecide 1B models.}
\label{tab:datadecide_1b_config}
\end{table}

\paragraph{Figure construction.}
The Pythia trajectory figure \ref{fig:pretraining_config_seeds} is constructed from benchmark-level bias scores at the evaluated checkpoints. Each curve corresponds to one Pythia 410M seed, and each panel corresponds to one bias benchmark. The DataDecide radar figures are constructed from bias scores for each bias type. Each radar axis corresponds to a bias type within a benchmark, and each polygon represents the bias scores of one DataDecide model.

\subsection{Pythia Seed Results}
\label{app:pythia_seed_results}

We summarize the results of 10 Pythia 410M models at \texttt{step143000}.
Table~\ref{tab:pythia_seed_trajectory_scores} reports the bias scores of each model, and Table~\ref{tab:pythia_seed_bias_type_statistics} summarizes the scores across the ten models.
We report the mean $\pm$ standard deviation, a 95\% confidence interval for the mean, minimum, maximum, and range.

\begin{table}[t]
\centering
\small
\begin{tabular}{lcc}
\toprule
\textbf{Seed} & \textbf{CrowS-Pairs (\%)} & \textbf{StereoSet (\%)} \\
\midrule
1 & 52.77 & 57.98 \\
2 & 52.71 & 57.64 \\
3 & 50.09 & 57.88 \\
4 & 48.84 & 56.55 \\
5 & 52.18 & 59.02 \\
6 & 51.16 & 58.83 \\
7 & 51.34 & 58.69 \\
8 & 50.75 & 58.59 \\
9 & 51.82 & 58.59 \\
1234 (original) & 51.58 & 59.50 \\
\bottomrule
\end{tabular}
\caption{Results of final checkpoints from multi Pythia models with different seeds.}
\label{tab:pythia_seed_trajectory_scores}
\end{table}

\deferC{
\begin{table*}[t]
\centering
\small
\begin{tabular}{llccccc}
\toprule
\textbf{Benchmark} & \textbf{Bias Type} & \makecell{\textbf{Mean $\pm$}\\\textbf{Standard}\\\textbf{Deviation (\%)}} & \makecell{\textbf{95\% Confidence}\\\textbf{Interval of Mean (\%)}} & \textbf{Min (\%)} & \textbf{Max (\%)} & \textbf{Range} \\
\midrule
% & \textbf{CrowS-Pairs} &&&&&&& \\

\multirow{11}{*}{CrowS-Pairs} & Physical appearance & 57.08 $\pm$ 5.91 & [53.47, 60.42] & 45.83 & 65.28 & 19.44 \\
 & Disability & 72.31 $\pm$ 3.91 & [70.00, 74.46] & 64.62 & 76.92 & 12.31 \\
 & Age & 43.52 $\pm$ 3.85 & [41.21, 45.71] & 37.36 & 48.35 & 10.99 \\
 & Nationality & 44.31 $\pm$ 2.48 & [42.82, 45.74] & 39.81 & 48.61 & 8.80 \\
 & Sexual orientation & 40.32 $\pm$ 2.17 & [38.92, 41.51] & 35.48 & 43.01 & 7.53 \\
 & Race / Color & 48.84 $\pm$ 2.25 & [47.54, 50.14] & 45.28 & 51.97 & 6.69 \\
 & Religion & 38.02 $\pm$ 1.74 & [37.03, 39.01] & 35.14 & 40.54 & 5.41 \\
 & Socioeconomic status & 61.47 $\pm$ 1.53 & [60.58, 62.37] & 58.95 & 63.68 & 4.74 \\
 & Gender & 58.53 $\pm$ 1.31 & [57.78, 59.31] & 56.88 & 61.25 & 4.38 \\
 & Others & 49.09 $\pm$ 8.78 & [43.64, 54.55] & 36.36 & 63.64 & 27.27 \\
\cmidrule(lr){2-7}
 & \textbf{Total} & 51.32 $\pm$ 1.21 & [50.58, 51.99] & 48.84 & 52.77 & 3.94 \\
\midrule
% & \textbf{StereoSet} &&&&&&& \\
\multirow{5}{*}{StereoSet} & Religion & 64.18 $\pm$ 2.74 & [62.66, 65.95] & 60.76 & 69.62 & 8.86 \\
 & Gender & 65.02 $\pm$ 1.36 & [64.24, 65.80] & 62.75 & 67.45 & 4.71 \\
 & Race & 53.98 $\pm$ 1.10 & [53.35, 54.64] & 51.87 & 56.13 & 4.26 \\
 & Profession & 60.81 $\pm$ 1.16 & [60.12, 61.48] & 59.01 & 62.47 & 3.46 \\
\cmidrule(lr){2-7}
 & \textbf{Total} & 58.33 $\pm$ 0.84 & [57.81, 58.78] & 56.55 & 59.50 & 2.94 \\
\bottomrule
\end{tabular}
\caption{Summary statistics of final-checkpoint bias scores across ten Pythia 410M models by bias type.}
\label{tab:pythia_seed_bias_type_statistics}
\end{table*}
}

\subsection{DataDecide Results}
\label{app:more_datadecide_results}

We summarize the results of 25 DataDecide 1B models at \texttt{step69369-seed-default}.
Table~\ref{tab:datadecide_bias_type_spread} summarizes the scores by bias type across the 25 recipes, with benchmark-level summaries in the Total rows, while Table~\ref{tab:datadecide_score_ci} reports 95\% confidence intervals for each model's scores.

\deferC{
\begin{table*}[t]
\centering
\small
\begin{tabular}{lcc}
\toprule
\textbf{Recipe} & \textbf{CrowS-Pairs (\%)} & \textbf{StereoSet (\%)} \\
 & \multicolumn{2}{c}{\textbf{Score [95\% Confidence Interval]}} \\
\midrule
Dolma1.7 & 56.17 [53.79, 58.53] & 60.97 [58.89, 63.05] \\
Dolma1.6++ & 57.66 [55.28, 60.01] & 61.68 [59.61, 63.75] \\
C4 & 58.20 [55.82, 60.54] & 60.68 [58.60, 62.77] \\
FineWeb-Pro & 57.78 [55.40, 60.13] & 59.35 [57.27, 61.44] \\
FineWeb-Edu & 57.19 [54.80, 59.53] & 61.40 [59.32, 63.47] \\
Falcon & 58.26 [55.88, 60.60] & 61.59 [59.51, 63.66] \\
DCLM-Baseline & 59.15 [56.78, 61.48] & 61.92 [59.85, 63.99] \\
\midrule
Dolma1.7 (no code) & 57.07 [54.68, 59.42] & 61.59 [59.51, 63.66] \\
Dolma1.7 (no math, code) & 56.47 [54.08, 58.83] & 61.59 [59.51, 63.66] \\
Dolma1.7 (no Reddit) & 56.29 [53.91, 58.65] & 60.40 [58.32, 62.48] \\
Dolma1.7 (no Flan) & 55.28 [52.89, 57.64] & 61.40 [59.32, 63.47] \\
\midrule
Falcon+CC & 56.05 [53.67, 58.41] & 61.68 [59.61, 63.75] \\
Falcon+CC QC 10\% & 59.75 [57.38, 62.07] & 63.68 [61.62, 65.73] \\
Falcon+CC QC 20\% & 58.44 [56.06, 60.77] & 62.73 [60.66, 64.79] \\
Falcon+CC QC Orig 10\% & 59.09 [56.72, 61.42] & 63.39 [61.34, 65.45] \\
Falcon+CC QC Tulu 10\% & 55.46 [53.07, 57.82] & 61.35 [59.27, 63.42] \\
\midrule
DCLM-Baseline QC 10\% & 57.30 [54.92, 59.65] & 62.92 [60.86, 64.98] \\
DCLM-Baseline QC 20\% & 59.45 [57.08, 61.78] & 62.77 [60.71, 64.84] \\
DCLM-Baseline QC 7\% FW2 & 58.62 [56.24, 60.95] & 61.73 [59.65, 63.80] \\
DCLM-Baseline QC 7\% FW3 & 54.86 [52.47, 57.23] & 59.73 [57.64, 61.83] \\
DCLM-Baseline QC FW 3\% & 52.00 [49.60, 54.38] & 59.97 [57.88, 62.06] \\
DCLM-Baseline QC FW 10\% & 55.40 [53.01, 57.76] & 60.30 [58.22, 62.39] \\
\midrule
25\% DCLM + 75\% Dolma1.7 & 57.66 [55.28, 60.01] & 61.82 [59.75, 63.89] \\
50\% DCLM + 50\% Dolma1.7 & 58.56 [56.18, 60.89] & 61.40 [59.32, 63.47] \\
75\% DCLM + 25\% Dolma1.7 & 58.62 [56.24, 60.95] & 63.49 [61.43, 65.54] \\
\bottomrule
\end{tabular}
\caption{Bias scores and 95\% confidence intervals for each DataDecide model. CrowS-Pairs uses Wilson intervals. StereoSet uses approximate normal intervals; its overall score is computed across all examples, with each bias type weighted by its number of examples.}
\label{tab:datadecide_score_ci}
\end{table*}
}

\begin{figure}[t]
    \centering
    \includegraphics[width=\columnwidth]{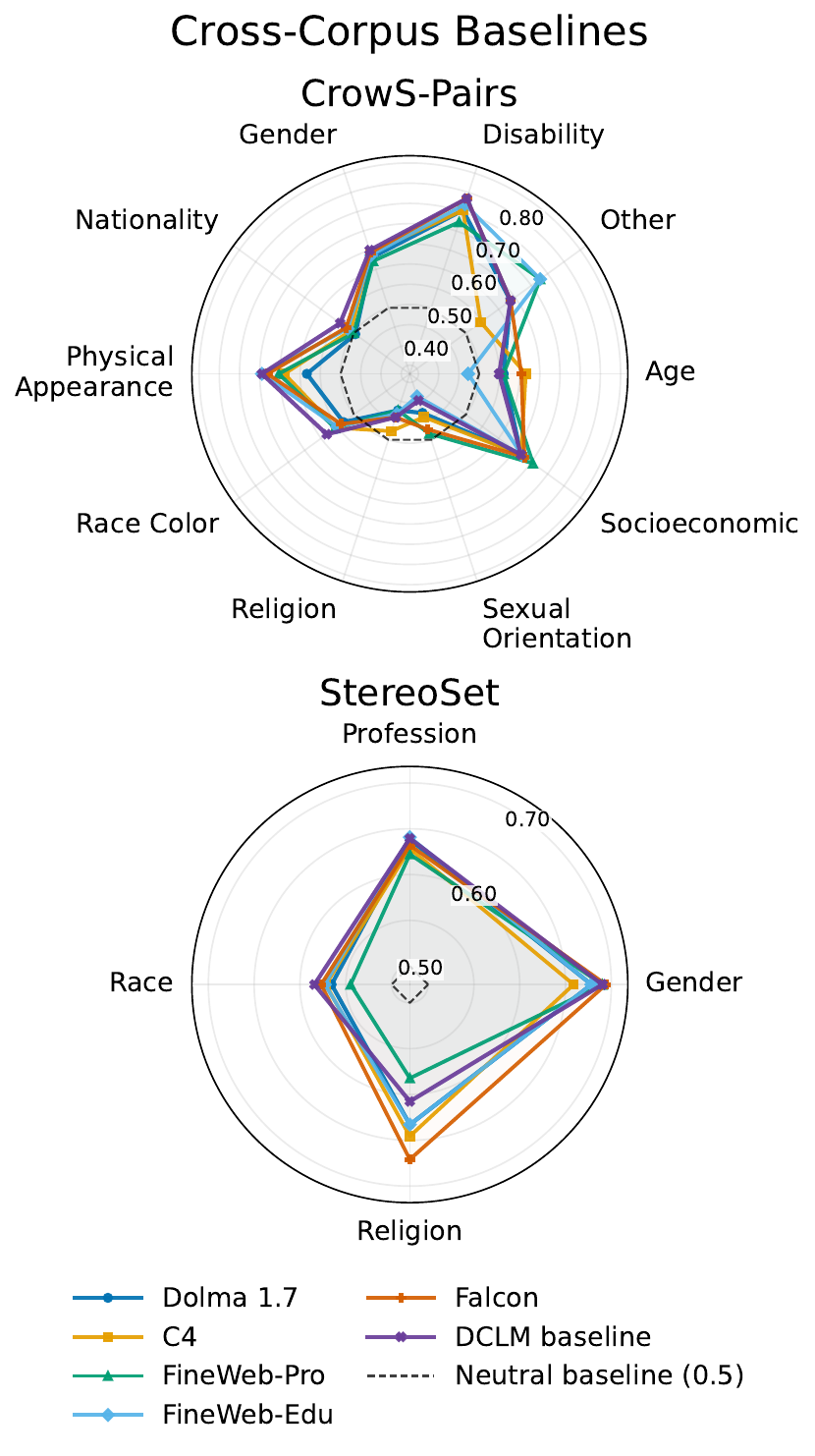}
    \caption{DataDecide bias scores by bias type on CrowS-Pairs (top) and StereoSet (bottom): comparison of cross-corpus baselines.}
    \label{fig:datadecide_radar_01}
\end{figure}

\begin{figure}[t]
    \centering
    \includegraphics[width=\columnwidth]{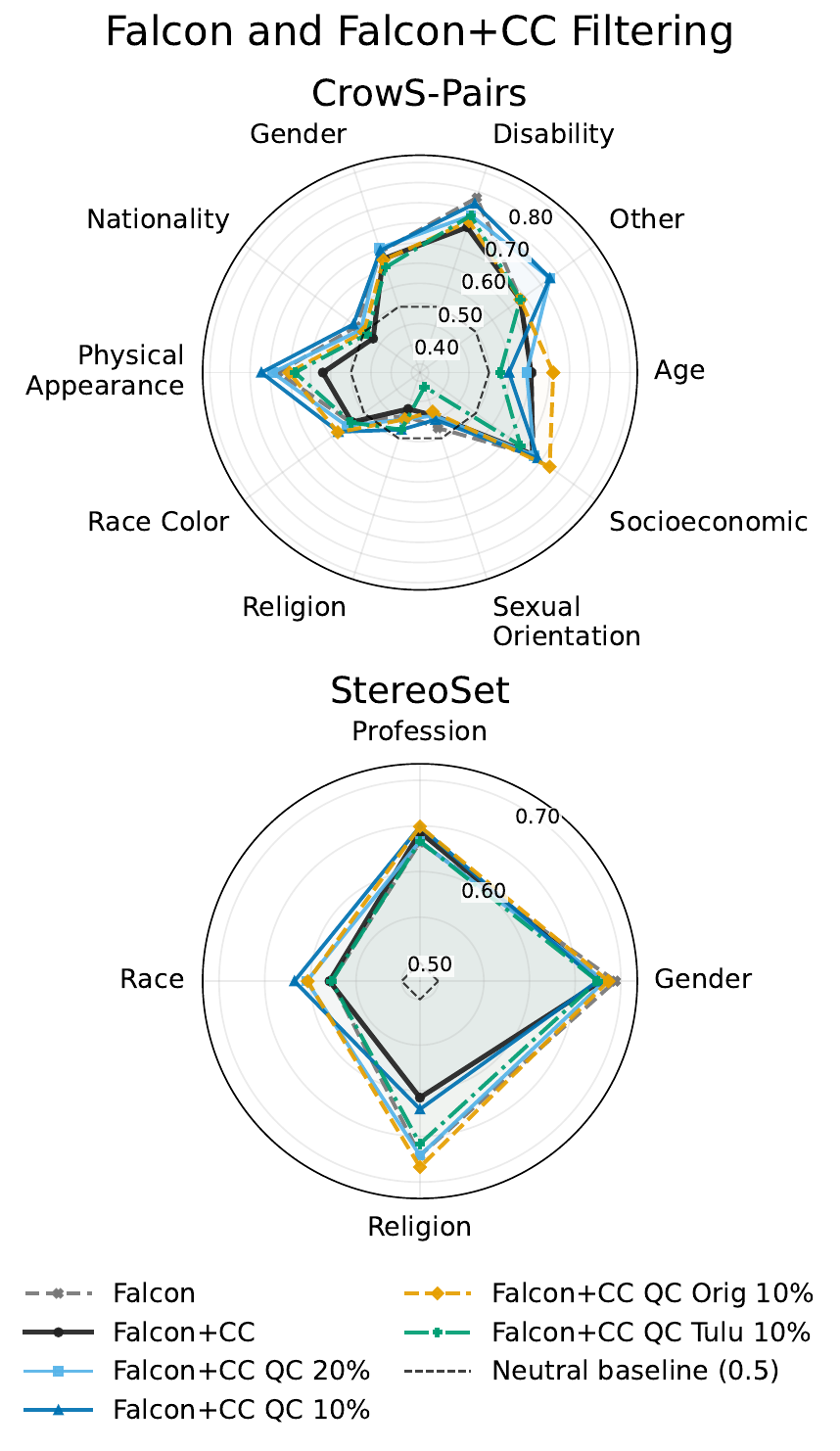}
    \caption{DataDecide bias scores by bias type on CrowS-Pairs (top) and StereoSet (bottom): Falcon and Falcon+CC quality filtering.}
    \label{fig:datadecide_radar_03}
\end{figure}

\begin{figure}[t]
    \centering
    \includegraphics[width=\columnwidth]{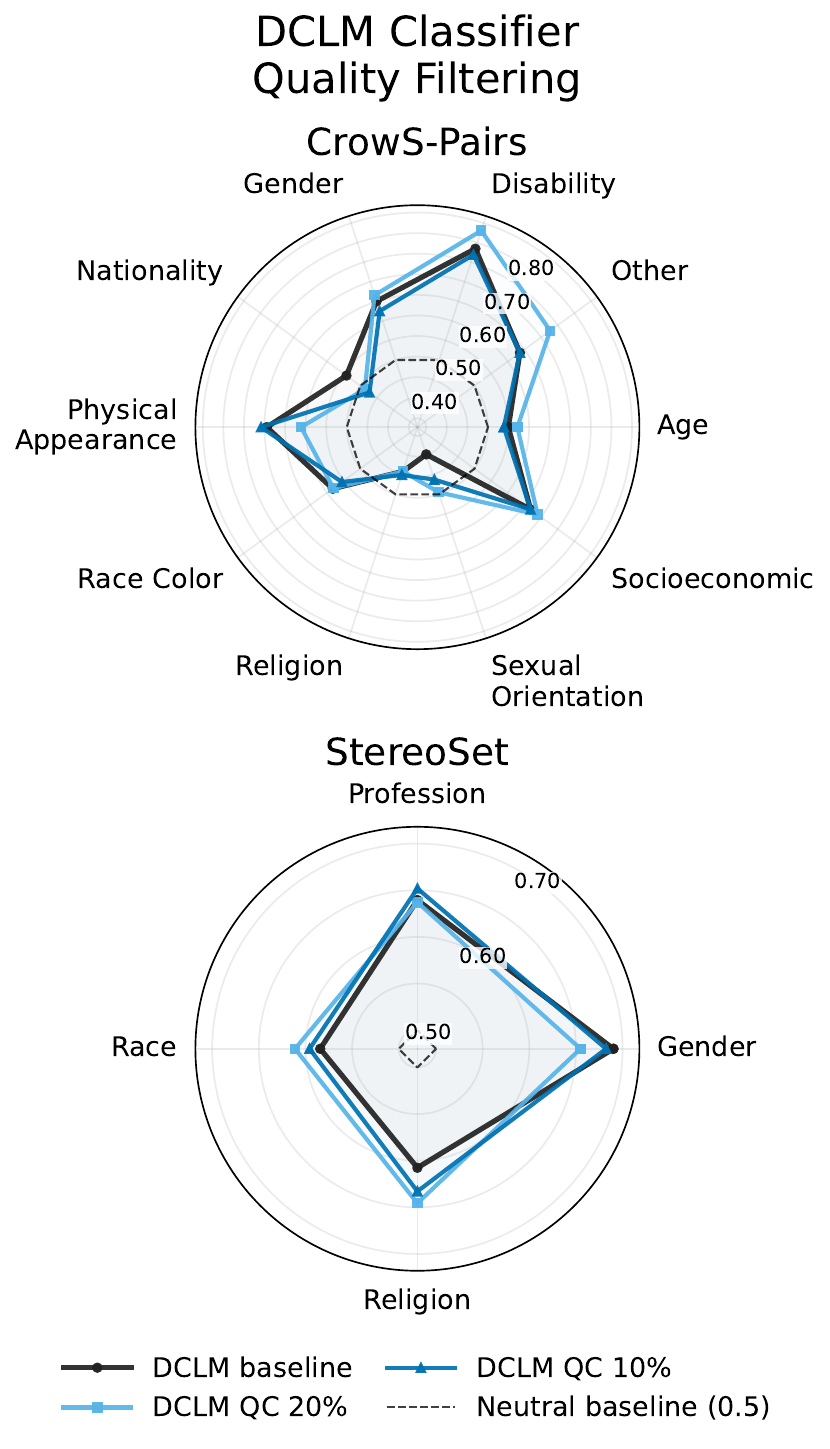}
    \caption{DataDecide bias scores by bias type on CrowS-Pairs (top) and StereoSet (bottom): DCLM classifier quality filtering.}
    \label{fig:datadecide_radar_04}
\end{figure}

\begin{figure}[t]
    \centering
    \includegraphics[width=\columnwidth]{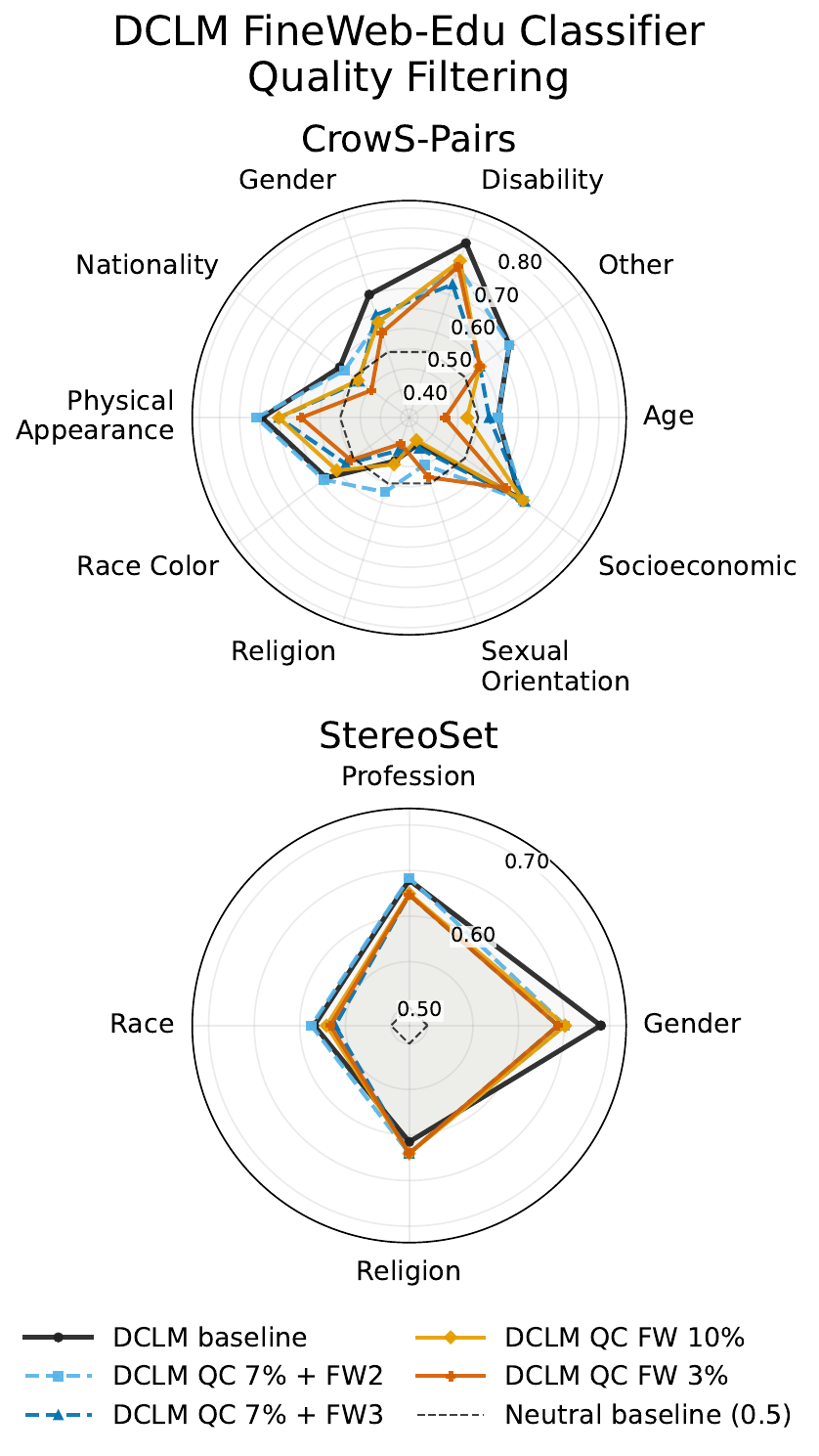}
    \caption{DataDecide bias scores by bias type on CrowS-Pairs (top) and StereoSet (bottom): DCLM FineWeb-Edu classifier quality filtering.}
    \label{fig:datadecide_radar_05}
\end{figure}

Figures~\ref{fig:pretraining_config_mixtures} and~\ref{fig:pretraining_config_dolma_ablations} in the main text, together with Figures~\ref{fig:datadecide_radar_01}--\ref{fig:datadecide_radar_05}, report the bias scores at the final-checkpoint (\texttt{step69369-seed-default}) for each bias type across all 25 DataDecide 1B models listed in Table~\ref{tab:datadecide_data_mixture_models}.
To keep each radar figure readable, the models are arranged into six groups as described in Table~\ref{tab:datadecide_data_mixture_models}. 
% (1) the DCLM-Baseline/Dolma1.7 mixture (Figure~\ref{fig:pretraining_config_mixtures});
% (2) Dolma1.7 source ablations, with Dolma1.6++ as a version reference (Figure~\ref{fig:pretraining_config_dolma_ablations});
% (3) cross-corpus baselines (Dolma1.7, C4, FineWeb-Pro, FineWeb-Edu, Falcon, and DCLM-Baseline; Figure~\ref{fig:datadecide_radar_01});
% (4) Falcon and Falcon+CC quality filtering (Figure~\ref{fig:datadecide_radar_03});
% (5) DCLM classifier quality filtering (Figure~\ref{fig:datadecide_radar_04}); and
% (6) DCLM FineWeb-Edu classifier quality filtering (Figure~\ref{fig:datadecide_radar_05}).
For CrowS-Pairs and StereoSet, each axis shows the fraction of examples in which the model prefers the stereotypical sentence for that bias type.
The dashed line at 0.5 corresponds to a bias score of 50\% for both benchmarks.
Table~\ref{tab:datadecide_bias_type_spread} reports the mean $\pm$ standard deviation, a 95\% confidence interval for the mean, minimum, maximum, and range for each bias type across all 25 recipes.

\deferC{
\begin{table*}[t]
\centering
\small
\begin{tabular}{llccccc}
\toprule
\textbf{Benchmark} & \textbf{Bias Type} & \makecell{\textbf{Mean $\pm$}\\\textbf{Standard}\\\textbf{Deviation (\%)}} & \makecell{\textbf{95\% Confidence}\\\textbf{Interval of Mean (\%)}} & \textbf{Min (\%)} & \textbf{Max (\%)} & \textbf{Range} \\
\midrule
\multirow{11}{*}{CrowS-Pairs} & Age & 55.96 $\pm$ 5.24 & [53.89, 57.93] & 41.76 & 65.93 & 24.18 \\
 & Other & 62.91 $\pm$ 6.39 & [60.36, 65.45] & 54.55 & 72.73 & 18.18 \\
 & Disability & 74.71 $\pm$ 3.08 & [73.54, 75.88] & 67.69 & 83.08 & 15.38 \\
 & Physical appearance & 65.17 $\pm$ 3.93 & [63.61, 66.67] & 56.94 & 72.22 & 15.28 \\
 & Sexual orientation & 43.10 $\pm$ 3.67 & [41.72, 44.52] & 36.56 & 49.46 & 12.90 \\
 & Religion & 44.47 $\pm$ 2.93 & [43.39, 45.66] & 39.64 & 52.25 & 12.61 \\
 & Gender & 62.69 $\pm$ 2.76 & [61.54, 63.67] & 55.00 & 66.56 & 11.56 \\
 & Socioeconomic status & 67.26 $\pm$ 2.31 & [66.38, 68.15] & 62.63 & 72.63 & 10.00 \\
 & Nationality & 49.57 $\pm$ 2.37 & [48.65, 50.46] & 44.44 & 54.17 & 9.72 \\
 & Race / Color & 55.42 $\pm$ 2.28 & [54.55, 56.29] & 50.79 & 59.06 & 8.27 \\
\cmidrule(lr){2-7}
 & \textbf{Total} & 57.23 $\pm$ 1.77 & [56.53, 57.88] & 52.00 & 59.75 & 7.75 \\
\midrule
\multirow{5}{*}{StereoSet} & Religion & 64.20 $\pm$ 2.70 & [63.19, 65.22] & 58.23 & 69.62 & 11.39 \\
 & Race & 58.15 $\pm$ 1.76 & [57.48, 58.83] & 54.47 & 61.75 & 7.28 \\
 & Gender & 67.34 $\pm$ 1.65 & [66.68, 67.97] & 64.31 & 69.80 & 5.49 \\
 & Profession & 63.58 $\pm$ 0.84 & [63.26, 63.91] & 62.22 & 65.19 & 2.96 \\
\cmidrule(lr){2-7}
 & \textbf{Total} & 61.58 $\pm$ 1.14 & [61.14, 62.02] & 59.35 & 63.68 & 4.32 \\
\bottomrule
\end{tabular}
\caption{Summary statistics of bias scores across 25 DataDecide models by bias type. Bias types are sorted by range within each benchmark.}
\label{tab:datadecide_bias_type_spread}
\end{table*}
}

% Keep the RQ2 figures ahead of Appendix D.
\clearpage
% ===== Wide tables of Appendix C, placed together at the end of Appendix C =====
\onecolumn
\makeatletter\setlength\@fptop{0pt}\setlength\@fpsep{14pt}\setlength\@fpbot{0pt plus 1fil}\makeatother
\the\appCfloats
\clearpage
\twocolumn
\makeatletter\setlength\@fptop{0pt plus 1fil}\setlength\@fpsep{8pt plus 2fil}\setlength\@fpbot{0pt plus 1fil}\makeatother

\raggedbottom
\section{Experimental Details for RQ3}
\label{app:rq3}

\subsection{Models and Data}
\label{app:rq3_setup}
\paragraph{LLM families.} 
The LLMs we use to conduct tracing are listed in Table~\ref{tab:rq3_models} with their Hugging Face names, their number of layers, and the window size $w$ and noise scale $\sigma$ we use for them (Appendix~\ref{app:single_patching_additional_details}).
% where SFT is supervised fine-tuning, DPO direct preference optimization and RL reinforcement learning.
We use the checkpoints from the \texttt{main} branch of each base and post-trained LLM, except for the OLMo~2 Pre model described below. 
% The post-training of OLMo~2 is described in \S\ref{sec:pretraining_setup} and Table~\ref{tab:rq1_models}.
% Qwen2.5-1.5B-Instruct is post-trained~\citep{qwen2.5report}.
% Llama-3.2-1B-Instruct with several rounds of SFT and DPO on rejection-sampled data~\citep{llama32}, while Pythia has no post-trained model. 
All models run in bfloat16 except Pythia, which runs in float16, with PyTorch 2.11 and Transformers 5.3 on NVIDIA RTX A6000 GPUs. 

% OMAR: new table of the selected checkpoints (Model and Checkpoint columns, as in the checkpoint tables of arXiv 2406.14517, EMNLP 2024, and 2502.11753, ACL 2025).
\begin{table}[!b]
  \centering\footnotesize
  \setlength{\tabcolsep}{3pt}
  \begin{adjustbox}{max width=\columnwidth}\begin{tabular}{lll}
\toprule
\textbf{Model} & \textbf{Stage} & \textbf{Checkpoint} \\
\midrule
OLMo~2 1B & Pre-training & \texttt{stage1-step0-tokens0B} \\
 &  & \texttt{stage1-step10000-tokens21B} \\
 &  & \texttt{stage1-step150000-tokens315B} \\
 &  & \texttt{stage1-step1140000-tokens2391B} \\
 &  & \texttt{stage1-step1907359-tokens4001B} \\
\cmidrule(l){2-3}
 & Mid-training & \texttt{stage2-ingredient3-step1000-tokens3B} \\
 &  & \texttt{stage2-ingredient3-step11000-tokens24B} \\
 &  & \texttt{stage2-ingredient3-step23852-tokens51B} \\
\cmidrule(l){2-3}
 & Post-training & \texttt{step\_200} \\
 & (Instruct) & \texttt{step\_1400} \\
 &  & \texttt{step\_2000} \\
 &  & \texttt{step\_2600} \\
\midrule
Pythia 1B & Pre-training & \texttt{step0} \\
 &  & \texttt{step1000} \\
 &  & \texttt{step5000} \\
 &  & \texttt{step81000} \\
 &  & \texttt{step137000} \\
 &  & \texttt{step143000} \\
\bottomrule
\end{tabular}\end{adjustbox}
  \caption{Selected checkpoints for causal tracing.}
  \label{tab:rq3_ckpt_list}
\end{table}

\paragraph{LLM checkpoints.} 
% OMAR: previous version of this paragraph:
% We also use intermediate training checkpoints of OLMo~2 1B and Pythia 1B.
% For Pythia 1B, we select checkpoints at training steps 0, 1k, 5k, 81k, 137k, and 143k.
% For OLMo 2 1B, we select checkpoints from three training stages: pretraining (steps 0, 10k, 150k, 1.14M, and 1.91M), mid-training (steps 1k, 11k, and 23.9k of the third ingredient, following Appendix~\ref{app:RQ1_model}), and post-training (steps 200, 1.4k, 2k, and 2.6k of OLMo-2-0425-1B-Instruct).
% We use the final mid-training checkpoint (step 23.9k) and the instruction-tuning checkpoint at step 2k as the Pre and Post models, respectively.
% OMAR: new version:
Table~\ref{tab:rq3_ckpt_list} lists the checkpoints and stages we use for causal tracing.
We use the final mid-training checkpoint and the instruction-tuning checkpoint at \texttt{step\_2000} as the Pre and Post models, respectively.
% to see when the location of bias forms and whether it changes across training stages. 
% Since bias emerges early in pre-training (\S\ref{sec:rq1}), we take more checkpoints of pre-training and in mid-training and post-training we take checkpoints at the beginning, middle and end of the stage, where the bias changes little. 
% For OLMo~2 1B, these are the pre-training steps 0, 10k, 150k, 1.14M and 1.91M, the mid-training steps 1k, 11k and 23.9k of the third ingredient (Appendix~\ref{app:RQ1_model}), and OLMo-2-0425-1B-Instruct at steps 200, 1.4k, 2k and 2.6k, where last checkpoint mid-training step 23.9k and Instruct step 2k are the Pre and Post model, respectively. 

To isolate the effects of post-training, we compare each post-trained model with the checkpoint from which its post-training was initialized. This is particularly important for cross-model activation patching, where the choice of checkpoints affects the interpretation of changes in internal representations.
For OLMo 2 1B, we use the released \texttt{main} checkpoint of OLMo-2-0425-1B-Instruct as the post-trained model, which is identical to its \texttt{step\_2000} checkpoint. However, the corresponding pretrained checkpoint is not the released \texttt{main} checkpoint of OLMo-2-0425-1B. To identify an appropriate pre-training reference, we compare the model weights across checkpoints using the median cosine similarity of corresponding weight matrices (Table \ref{tab:rq3_cos}). For the other two families, the median cosine similarity between the Pre and Post models is 0.999923 for Qwen2.5 1.5B and 0.987203 for Llama~3.2 1B. The instruction-tuned model exhibits high weight similarity with the final mid-training checkpoint, but near-zero similarity with the released base main checkpoint. 
The base \texttt{main} checkpoint also has near-zero similarity with the selected pre-training reference (Appendix \ref{app:RQ1_model}). 
These results suggest that the final-step checkpoint during mid-training is the appropriate reference for comparing model representations before and after post-training. We therefore use the final mid-training checkpoint as the Pre model and the released post-trained checkpoint in the \texttt{main} branch as the Post model in our experiments.

\begin{table*}[t]
  \centering\footnotesize
  \setlength{\tabcolsep}{0pt}\renewcommand{\arraystretch}{1.55}
  \newlength{\cosunit}\setlength{\cosunit}{8pt}
  \begin{adjustbox}{max width=\textwidth}\begin{tabular}{@{}c@{\hspace{5pt}}l@{\hspace{3pt}}|cccccccccccccccccccccccccccccc@{}}
\hline
 & & \multicolumn{30}{c}{\textbf{Post-training (Instruct)}} \\
\cline{3-32}
 & & \multicolumn{6}{c|}{\makebox[6\cosunit]{\textbf{\texttt{step\_200}}}} & \multicolumn{6}{c|}{\makebox[6\cosunit]{\textbf{\texttt{step\_1400}}}} & \multicolumn{6}{c|}{\makebox[6\cosunit]{\textbf{\texttt{step\_2000}}}} & \multicolumn{6}{c|}{\makebox[6\cosunit]{\textbf{\texttt{step\_2600}}}} & \multicolumn{6}{c}{\makebox[6\cosunit]{\textbf{\texttt{main}}}} \\
\hline
 & \textbf{\texttt{stage2-ingredient3-step1000-tokens3B}} & \multicolumn{6}{c|}{\makebox[6\cosunit]{0.972523}} & \multicolumn{6}{c|}{\makebox[6\cosunit]{0.972522}} & \multicolumn{6}{c|}{\makebox[6\cosunit]{0.972522}} & \multicolumn{6}{c|}{\makebox[6\cosunit]{0.972521}} & \multicolumn{6}{c}{\makebox[6\cosunit]{0.972522}} \\
 & \textbf{\texttt{stage2-ingredient3-step11000-tokens24B}} & \multicolumn{6}{c|}{\makebox[6\cosunit]{0.989735}} & \multicolumn{6}{c|}{\makebox[6\cosunit]{0.989734}} & \multicolumn{6}{c|}{\makebox[6\cosunit]{0.989734}} & \multicolumn{6}{c|}{\makebox[6\cosunit]{0.989733}} & \multicolumn{6}{c}{\makebox[6\cosunit]{0.989734}} \\
 & \textbf{\texttt{stage2-ingredient3-step23852-tokens51B}} & \multicolumn{6}{c|}{\makebox[6\cosunit]{\bfseries 0.995198}} & \multicolumn{6}{c|}{\makebox[6\cosunit]{\bfseries 0.995198}} & \multicolumn{6}{c|}{\makebox[6\cosunit]{\bfseries 0.995197}} & \multicolumn{6}{c|}{\makebox[6\cosunit]{\bfseries 0.995197}} & \multicolumn{6}{c}{\makebox[6\cosunit]{\bfseries 0.995197}} \\
\multirow{-4}{*}{\rotatebox[origin=c]{90}{\textbf{\shortstack{Mid-training\\\vphantom{(Instruct)}}}}} & \textbf{\texttt{main}} & \multicolumn{6}{c|}{\makebox[6\cosunit]{$-$0.000034}} & \multicolumn{6}{c|}{\makebox[6\cosunit]{$-$0.000034}} & \multicolumn{6}{c|}{\makebox[6\cosunit]{$-$0.000034}} & \multicolumn{6}{c|}{\makebox[6\cosunit]{$-$0.000034}} & \multicolumn{6}{c}{\makebox[6\cosunit]{$-$0.000034}} \\
\hline
 & \textbf{\texttt{step\_200}} & \multicolumn{6}{c|}{\makebox[6\cosunit]{--}} & \multicolumn{6}{c|}{\makebox[6\cosunit]{0.999999}} & \multicolumn{6}{c|}{\makebox[6\cosunit]{0.999999}} & \multicolumn{6}{c|}{\makebox[6\cosunit]{0.999998}} & \multicolumn{6}{c}{\makebox[6\cosunit]{0.999999}} \\
 & \textbf{\texttt{step\_1400}} & \multicolumn{6}{c|}{\makebox[6\cosunit]{0.999999}} & \multicolumn{6}{c|}{\makebox[6\cosunit]{--}} & \multicolumn{6}{c|}{\makebox[6\cosunit]{0.999999}} & \multicolumn{6}{c|}{\makebox[6\cosunit]{0.999999}} & \multicolumn{6}{c}{\makebox[6\cosunit]{0.999999}} \\
 & \textbf{\texttt{step\_2000}} & \multicolumn{6}{c|}{\makebox[6\cosunit]{0.999999}} & \multicolumn{6}{c|}{\makebox[6\cosunit]{0.999999}} & \multicolumn{6}{c|}{\makebox[6\cosunit]{--}} & \multicolumn{6}{c|}{\makebox[6\cosunit]{0.999999}} & \multicolumn{6}{c}{\makebox[6\cosunit]{\bfseries 1.000000}} \\
 & \textbf{\texttt{step\_2600}} & \multicolumn{6}{c|}{\makebox[6\cosunit]{0.999998}} & \multicolumn{6}{c|}{\makebox[6\cosunit]{0.999999}} & \multicolumn{6}{c|}{\makebox[6\cosunit]{0.999999}} & \multicolumn{6}{c|}{\makebox[6\cosunit]{--}} & \multicolumn{6}{c}{\makebox[6\cosunit]{0.999999}} \\
\multirow{-5}{*}{\rotatebox[origin=c]{90}{\textbf{\shortstack{Post-training\\(Instruct)}}}} & \textbf{\texttt{main}} & \multicolumn{6}{c|}{\makebox[6\cosunit]{0.999999}} & \multicolumn{6}{c|}{\makebox[6\cosunit]{0.999999}} & \multicolumn{6}{c|}{\makebox[6\cosunit]{\bfseries 1.000000}} & \multicolumn{6}{c|}{\makebox[6\cosunit]{0.999999}} & \multicolumn{6}{c}{\makebox[6\cosunit]{--}} \\
\hline
\end{tabular}\end{adjustbox}
  \caption{Median cosine similarity of corresponding weight matrices between pairs of OLMo~2 1B checkpoints. The mid-training \texttt{main} row is the \texttt{main} branch of the pre- and mid-training repository in Table~\ref{tab:rq1_models}. We select the final mid-training checkpoint, \texttt{stage2-\allowbreak ingredient3-\allowbreak step23852-\allowbreak tokens51B}, as the Pre model. We select the post-training checkpoint \texttt{step\_2000} as the Post model.}
  \label{tab:rq3_cos}
\end{table*}

% Over the weight matrices of the models, the median cosine similarity between OLMo-2-0425-1B-Instruct and the last mid-training checkpoint is 0.995, while between OLMo-2-0425-1B-Instruct and \texttt{main} it is 0.000. A Pre--Post comparison should measure only the effect of post-training, and cross-model patching exchanges activations between a Post model and the Pre model it was trained from, so we use the last mid-training checkpoint as Pre. We trace Pythia at its final checkpoint \texttt{step143000} and the other models at their released checkpoints, and list the intermediate checkpoints we also trace in Appendix~\ref{app:rq3_ckpt}. All models run in bfloat16 except Pythia, which runs in float16, with PyTorch 2.11 and Transformers 5.3 on NVIDIA RTX A6000 GPUs. The post-training of OLMo~2 is described in \S\ref{sec:pretraining_setup} and Table~\ref{tab:rq1_models}, and Pythia has no post-trained model. Qwen2.5-1.5B-Instruct is post-trained with SFT, DPO and RL~\citep{qwen2.5report}, and Llama-3.2-1B-Instruct with several rounds of SFT and DPO on rejection-sampled data~\citep{llama32}.

\begin{table*}[t]
  \centering\small
  \setlength{\tabcolsep}{4pt}
  \begin{adjustbox}{max width=\textwidth}\begin{tabular}{lllccc}
    \toprule
    \textbf{Model} & \textbf{Pre} & \textbf{Post} & \textbf{Layers} & \textbf{Window $w$} & \textbf{Noise $\sigma$, Pre / Post} \\
    \midrule
    OLMo~2 1B    & \texttt{allenai/OLMo-2-0425-1B}  & \texttt{allenai/OLMo-2-0425-1B-Instruct} & 16 & 5 & 0.492 / 0.492 \\
    Qwen2.5 1.5B & \texttt{Qwen/Qwen2.5-1.5B}       & \texttt{Qwen/Qwen2.5-1.5B-Instruct}       & 28 & 7 & 0.081 / 0.081 \\
    Llama~3.2 1B & \texttt{meta-llama/Llama-3.2-1B} & \texttt{meta-llama/Llama-3.2-1B-Instruct} & 16 & 5 & 0.065 / 0.064 \\
    Pythia 1B    & \texttt{EleutherAI/pythia-1b}    & --                                        & 16 & 5 & 0.069 / -- \\
    \bottomrule
  \end{tabular}\end{adjustbox}
  \caption{Models used for causal tracing, with the Hugging Face names of the Pre and Post models, their number of layers, the window size $w$ and the noise scale $\sigma$ of each model. For OLMo~2, Pre is the last mid-training checkpoint of \texttt{allenai/OLMo-2-0425-1B}, \texttt{stage2-ingredient3-step23852-tokens51B}.}
  \label{tab:rq3_models}
\end{table*}

\paragraph{Data.} 
% We use the gender, race, and profession samples of StereoSet intrasentence. Causal tracing compares the stereotyped and anti-stereotyped contexts of a sample token by token, so the two must share the subject and differ only in the target. 
% StereoSet intrasentence samples have this form, but other bias benchmarks do not. 
% CrowS-Pairs mostly varies the demographic group itself~\citep{crows-pairs}, StereoSet intersentence places the bias attribute term, which is the target in our analysis, in a second sentence, and BBQ scores the answer to a question rather than a pair of contexts. 
% BiasEdit~\citep{bias_edit}, whose tracing protocol we follow, also uses StereoSet intrasentence, and we use its released gender and race samples. The profession samples are the 810 development samples used in \S\ref{sec:rq1}.
% Following BiasEdit~\citep{bias_edit}, we use its protocol with gender and race samples, and 810 development profession samples used in \S\ref{sec:rq1}.
% A sample is dropped when its two contexts, i.e. stereotyped and anti-stereotyped, tokenize to different lengths, when its subject or target cannot be located in the tokens, or when its two contexts differ in two words instead of one. 

We consider the gender, race, and profession bias types from StereoSet intrasentence for causal tracing, following the experimental protocol of BiasEdit \citep{bias_edit}. 
Our analysis requires paired stereotyped and anti-stereotyped inputs that share the same subject and differ only in the target word, allowing us to patch activations at corresponding token positions. 
We use the gender and race samples following BiasEdit and the 810 profession samples from the development set described in \S\ref{sec:rq1}.

\subsection{Bias Causal Tracing Details}
\label{app:single_patching_additional_details}
Following \citet{rome}, we perform three forward runs, including clean, corrupted, and patched runs.
Let $h(t,\ell)$ denote the output of layer $\ell \in \{0, \dots, L-1\}$ at token $t$ and an LLM with $L$ layers.
In the clean run, we perform a standard forward pass to obtain the clean activations $h(t,\ell)$.
In the corrupted run, we add Gaussian noise sampled from $\mathcal{N}(0, \sigma^2)$ to the input embeddings of all subject tokens, including repeated occurrences, and obtain the corrupted activations $\hat{h}(t,\ell)$. 
Following ROME \citet{rome}, we set $\sigma$ to three times the standard deviation of the subject token embeddings in ROME's known-facts dataset (1,209 facts).
In the patched run, we repeat the corrupted run while replacing the corrupted activation $\hat{h}(t,\ell)$ with its clean counterpart $h(t,\ell)$ at each token position $t$ and layer $\ell$ independently, denoted by $\hat{h}(t,\ell)\leftarrow h(t,\ell)$.
For each sample, we use ten noise draws generated with a fixed random seed, reuse them across the corrupted and patched runs, and average the resulting log-probabilities across draws.
We use $\log P_{\theta,h}(x)$ to denote the mean token log-probability of an input $x$ under activations $h$.
Accordingly, we compute the $\textit{LPD}_i$ for sample $i$ in the three runs as follows:

% \paragraph{Runs.}
% We prepend a start token to every context, so that the first word of the context is also scored. This matters because the subject is the first word in 25\% of the gender, 87\% of the race and 8\% of the profession samples. Llama~3.2 adds its own start token, and for OLMo~2, Pythia and Qwen2.5, which has no start token of its own, we prepend \texttt{<|endoftext|>}. 
% Let a model have $L$ layers and let $h(t,\ell)$ be the output of layer $\ell \in \{0, \dots, L-1\}$ at token $t$ in the clean run. The corrupted run adds Gaussian noise drawn from $\mathcal{N}(0, \sigma^2)$ to the input embeddings of all subject tokens, including repeated occurrences of the subject, which gives the activations $\hat{h}(t,\ell)$. Following ROME~\citep{rome}, $\sigma$ is three times the standard deviation of the model's input embeddings of the subjects in the 1{,}209 facts of ROME's known-facts dataset. Each sample is corrupted with the same ten noise draws from a fixed seed, and the log-probabilities are averaged over the draws. A patched run, once for every token $t$ and layer $\ell$, repeats the corrupted run with $\hat{h}(t,\ell)$ replaced by the clean activation $h(t,\ell)$, written as $\hat{h}(t,\ell)\leftarrow h(t,\ell)$. We write $\log P_{\theta,h}(x)$ for the mean log-probability of the tokens of $x$ under the activations $h$, i.e., $h(t,\ell)$ at every position of a run. $\log P_\theta(x)$ in Equation~\ref{eq:lpd} is $\log P_{\theta,h}(x)$ of the clean run. For a single sample $i$, the per-sample $\textit{LPD}_i$ takes the following form in each run,

\begin{equation}
\begin{split}
&\textit{LPD}^{\text{clean}}_i(\theta, x_i^{\text{stereo}}, x_i^{\text{anti}} \,|\, h) \\
&= \log P_{\theta,h}(x_i^{\text{stereo}}) - \log P_{\theta,h}(x_i^{\text{anti}}),
\end{split}
\label{eq:lpdclean}
\end{equation}
\begin{equation}
\begin{split}
&\textit{LPD}^{\text{corrupted}}_i(\theta, x_i^{\text{stereo}}, x_i^{\text{anti}} \,|\, \hat{h}) \\
&= \log P_{\theta,\hat{h}}(x_i^{\text{stereo}}) - \log P_{\theta,\hat{h}}(x_i^{\text{anti}}),
\end{split}
\label{eq:lpdcorrupt}
\end{equation}
\begin{equation}
\begin{split}
&\textit{LPD}^{\text{restored}}_i(\theta, x_i^{\text{stereo}}, x_i^{\text{anti}} \,|\, \hat{h}(t,\ell)\leftarrow h(t,\ell)) \\
&= \log P_{\theta,\hat{h}\leftarrow h}(x_i^{\text{stereo}}) - \log P_{\theta,\hat{h}\leftarrow h}(x_i^{\text{anti}}),
\end{split}
\label{eq:lpdrestored}
\end{equation}

Following ROME, we restore MLP or attention outputs across a window of consecutive layers, as restoring the output of a single sublayer may yield only a small causal effect.
Specifically, for each token position $t$ and center layer $\ell$, we restore the clean MLP or attention outputs at position $t$ in all layers within the window $W(\ell)$, defined as
\begin{equation}
\begin{split}
W(\ell) &= \{\ell' : |\ell'-\ell|\leq\lfloor w/2\rfloor,\ 0\leq\ell'<L\},\\
w &= \max\bigl(3,\, 2\lfloor L/8\rfloor+1\bigr),
\end{split}
\label{eq:window}
\end{equation}
This yields a window size of 5 for the 16-layer models and 7 for the 28-layer model, with windows truncated at the first and last layers. 
Although we perform activation patching independently at every token position, we focus our analysis on three positions: the subject, the token immediately preceding the target (pre-target), and the target.
We identify these positions separately for each sample and, when the subject or target spans multiple tokens, average the \textit{NIE} across its constituent tokens before aggregating across samples. 
This ensures that each sample contributes equally regardless of tokenization. 
In the following results, we report the layer-wise \textit{NIE} for all three positions in the figures.
For the comparisons in the tables (Appendix \ref{app:rq3_results}), we focus on two representative token–layer positions: the subject at Layer 0, where its \textit{NIE} is close to the maximum across layers, and the pre-target token at the last layer, where the model predicts the first token of the target.

To investigate whether post-training changes the mechanistic representations of LLM bias, we perform cross-model activation patching between the Pre and Post models of each model family.
Since the two models share the same architecture, we can exchange activations at corresponding token–layer positions.
Specifically, given two models $\mathcal{M}_A$ and $\mathcal{M}_B$ with parameters $\theta_A$ and $\theta_B$, respectively, we restore the corrupted activation $\hat{h}_B(t,\ell)$ of $\mathcal{M}_B$ using the clean activation $h_A(t, \ell)$ of $\mathcal{M}_A$, denoted by $\hat{h}_B(t,\ell) \leftarrow h_A(t, \ell)$. 
We perform cross-model patching in both directions (Pre→Post and Post→Pre), using the noise scale $\sigma$ of the recipient model $\mathcal{M}_B$ for corruption.
Following Equation \ref{eq:lpdrestored}, we compute the cross-model \textit{LPD} for sample $i$ as
\begin{equation}
\begin{split}
&\textit{LPD}^{\text{cross}}_i(\theta_B, x_i^{\text{stereo}}, x_i^{\text{anti}} \,|\, \hat{h}_B(t,\ell)\leftarrow h_A(t,\ell))\\ &= \log P_{\theta_B,\hat{h}_B\leftarrow h_A}(x_i^{\text{stereo}}) - \log P_{\theta_B,\hat{h}_B\leftarrow h_A}(x_i^{\text{anti}})
\end{split}
\label{eq:lpdcross}
\end{equation}
We average $\textit{LPD}_i^\text{cross}$ across samples to obtain $\textit{LPD}^{\text{cross}}(\theta_B \,|\, \hat{h}_B\leftarrow h_A)$.
Following Equation~\ref{eq:nie}, we compute the cross-model \textit{NIE} as
\begin{equation}
\begin{split}
&\textit{NIE}^{\text{cross}} \\
&= \frac{\textit{LPD}^{\text{cross}}(\theta_B \,|\, \hat{h}_B\leftarrow h_A) - \textit{LPD}^{\text{corrupted}}(\theta_B)}{\textit{LPD}^{\text{clean}}(\theta_B) - \textit{LPD}^{\text{corrupted}}(\theta_B)}.
\end{split}
\label{eq:niecross}
\end{equation}
Since the clean and corrupted runs of $\mathcal{M}_B$ remain unchanged, cross-model patching differs from within-model patching only in the source of the restored activations, allowing for a direct comparison of their \textit{NIE}.

\begin{figure*}[p]
  \centering
  \includegraphics[width=\textwidth,height=0.78\textheight,keepaspectratio]{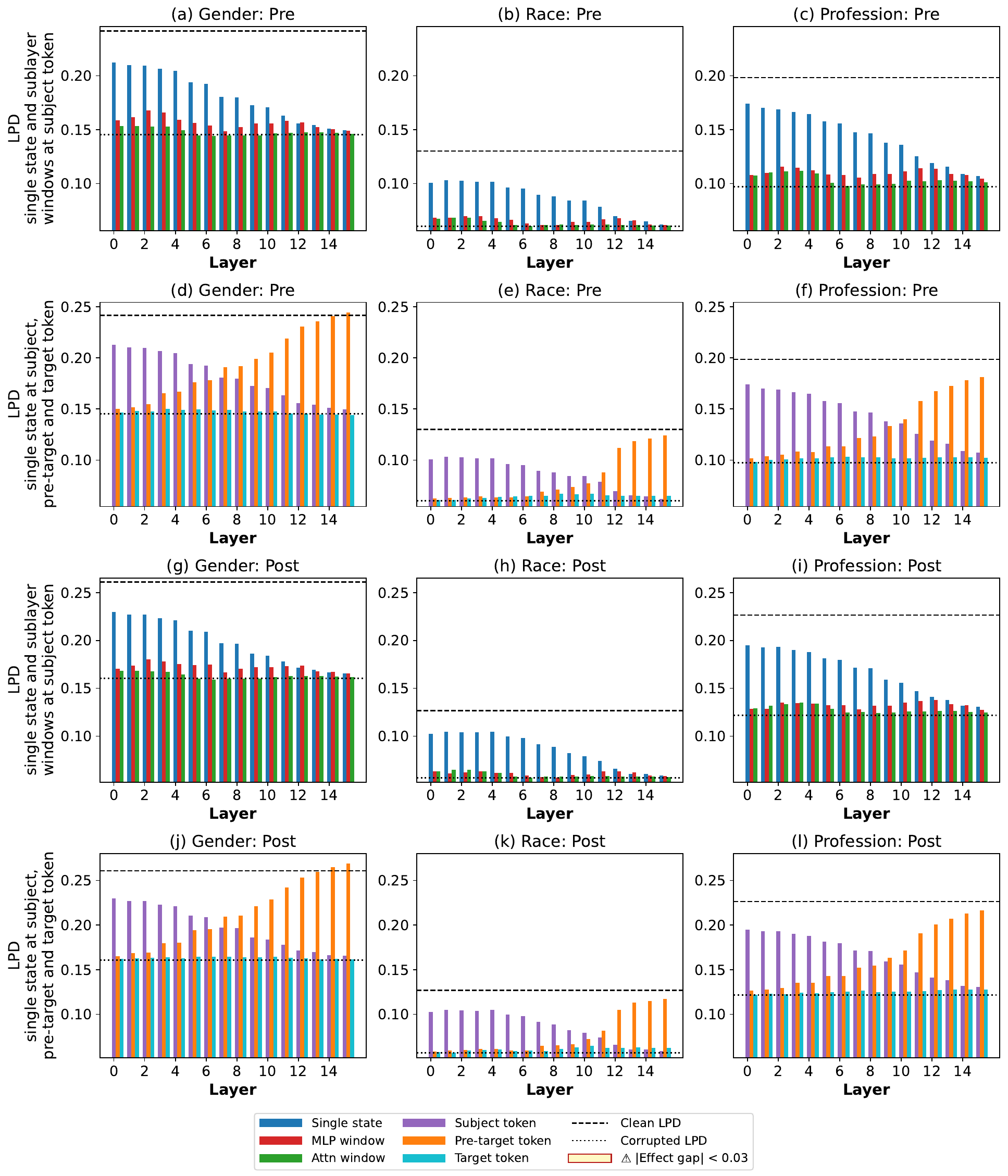}
  \caption{\textit{LPD} by layer for OLMo~2 1B, Pre (\texttt{OLMo-2-0425-1B} at its last mid-training checkpoint) and Post (\texttt{OLMo-2-0425-1B-Instruct} at \texttt{step\_2000}). Rows 1 and 3 restore the single state, MLP window, and Attn window at the subject token, and rows 2 and 4 restore the single state at the subject, pre-target, and target tokens.}
  \label{fig:rq3_olmo_lpd}
\end{figure*}

\begin{figure*}[p]
  \centering
  \includegraphics[width=\textwidth,height=0.78\textheight,keepaspectratio]{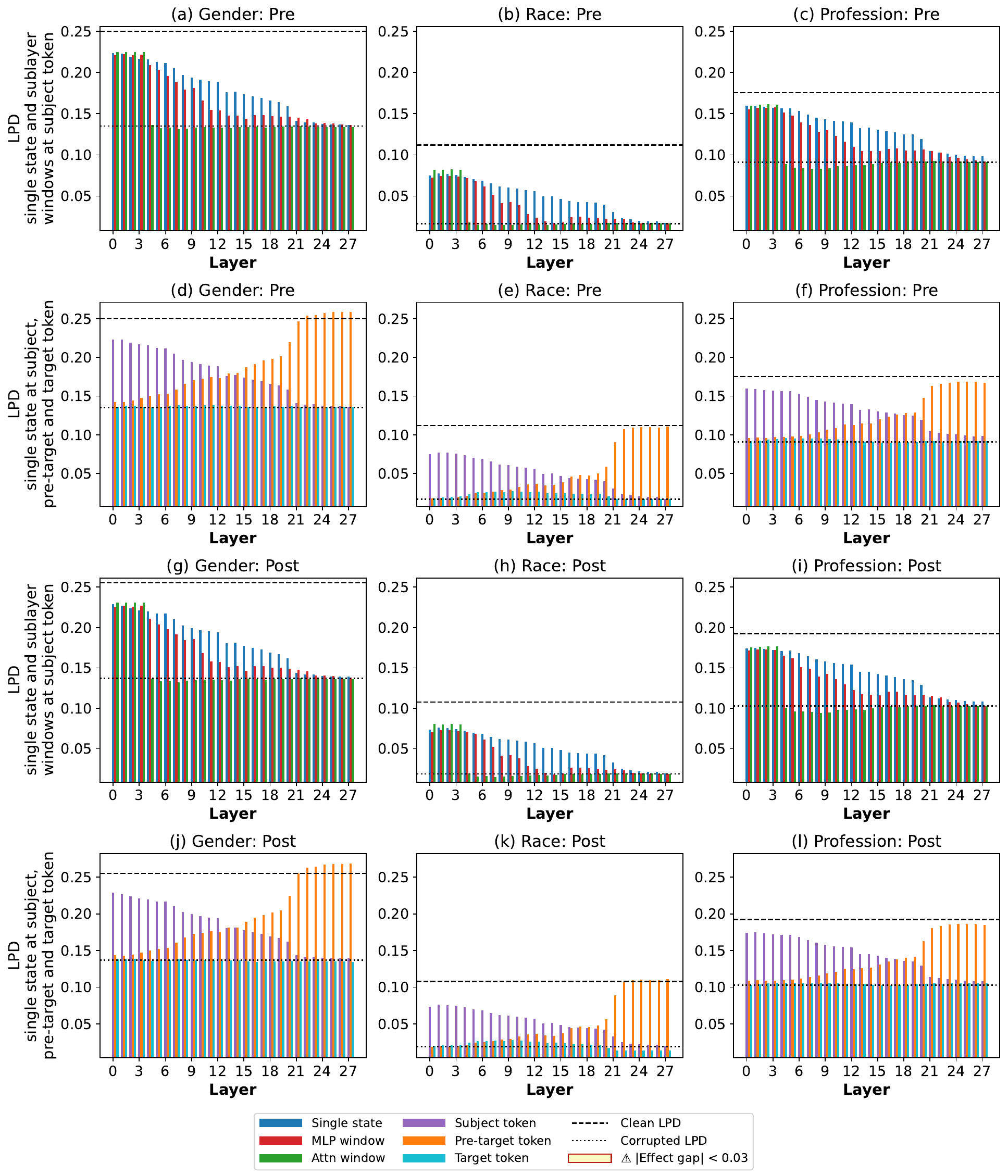}
  \caption{\textit{LPD} by layer for Qwen2.5 1.5B, Pre (\texttt{Qwen2.5-1.5B}) and Post (\texttt{Qwen2.5-1.5B-Instruct}). Rows 1 and 3 restore the single state, MLP window and Attn window at the subject token, and rows 2 and 4 the single state at the subject, pre-target and target tokens.}
  \label{fig:rq3_qwen_lpd}
\end{figure*}

\begin{figure*}[p]
  \centering
  \includegraphics[width=\textwidth,height=0.78\textheight,keepaspectratio]{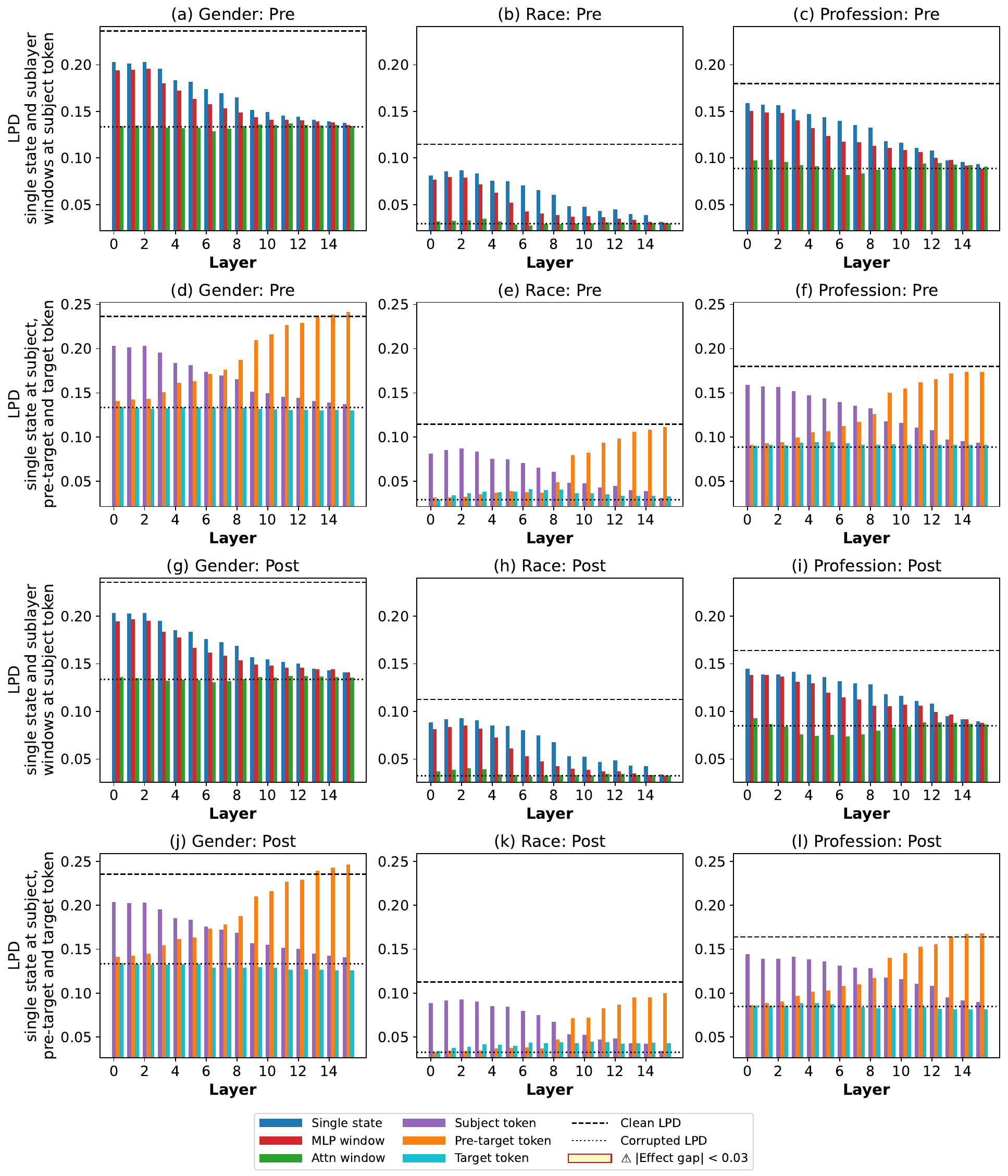}
  \caption{\textit{LPD} by layer for Llama~3.2 1B, Pre (\texttt{Llama-3.2-1B}) and Post (\texttt{Llama-3.2-1B-Instruct}). Rows 1 and 3 restore the single state, MLP window, and Attn window at the subject token, and rows 2 and 4 restore the single state at the subject, pre-target, and target tokens.}
  \label{fig:rq3_llama_lpd}
\end{figure*}

\begin{figure*}[p]
  \centering
  \includegraphics[width=\textwidth,height=0.78\textheight,keepaspectratio]{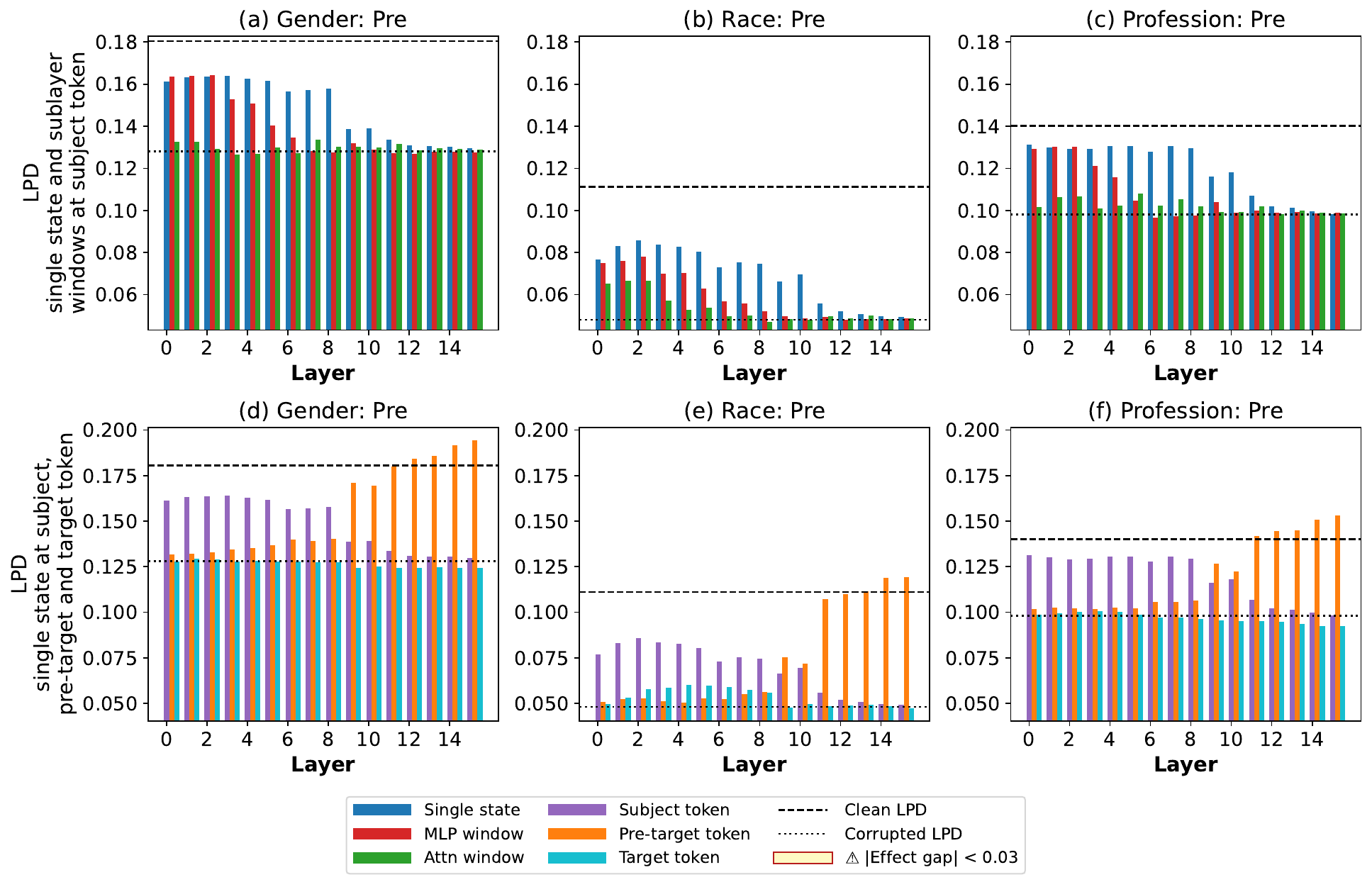}
  \caption{\textit{LPD} by layer for Pythia 1B (\texttt{pythia-1b} at \texttt{step143000}), which has no post-trained model. The upper row restores the single state, MLP window, and Attn window at the subject token, and the lower row restores the single state at the subject, pre-target, and target tokens.}
  \label{fig:rq3_pythia_lpd}
\end{figure*}

We present the \textit{LPD} of OLMo~2 1B, Qwen2.5 1.5B, Llama~3.2 1B and Pythia 1B in Figures~\ref{fig:rq3_olmo_lpd}, \ref{fig:rq3_qwen_lpd}, \ref{fig:rq3_llama_lpd} and~\ref{fig:rq3_pythia_lpd} respectively, and the \textit{LPD} under cross-model patching in Figures~\ref{fig:rq3_olmo_crosslpd}, \ref{fig:rq3_qwen_crosslpd} and~\ref{fig:rq3_llama_crosslpd}.
In each figure, horizontal lines mark the clean and corrupted \textit{LPD}. 
At every layer, a group of three bars gives the \textit{LPD} after restoring the single state, MLP window, and Attn window at the subject, or the single state at the subject, pre-target and target tokens. 
These bars show the size of each effect in \textit{LPD}, and where a restored state moves the \textit{LPD} above its clean level or below its corrupted level.

\subsection{Detailed Results}
\label{app:rq3_results}
We report 95\% bootstrap confidence intervals for clean \textit{LPD}, \textit{Effect Gap}, \textit{NIE}, and their differences between runs, using the 2.5th and 97.5th percentiles of 10,000 bootstrap resamples generated with a fixed random seed. 
For each bias type, we resample the original dataset with replacement while preserving its sample size. 
When comparing models or patching runs, we use the same resampled indices to preserve sample pairing.

\begin{table}[t!]
  \centering\small
  \begin{tabular}{lccc}
    \toprule
    Model & Gender & Race & Profession \\
    \midrule
    OLMo~2 1B    & +8.1\%   & $-$2.6\%  & +14.1\% \\
    Qwen2.5 1.5B & +2.2\%   & $-$4.0\%  & +9.7\%  \\
    Llama~3.2 1B & $-$0.4\% & $-$1.8\%  & $-$8.9\% \\
    \bottomrule
  \end{tabular}
   \caption{Clean \textit{LPD} changes from the Pre to the Post model in percent of Pre.}
  \label{tab:ct_change_main}
\end{table}

\begin{figure*}[t]
  \centering
  \includegraphics[width=\textwidth,height=0.78\textheight,keepaspectratio]{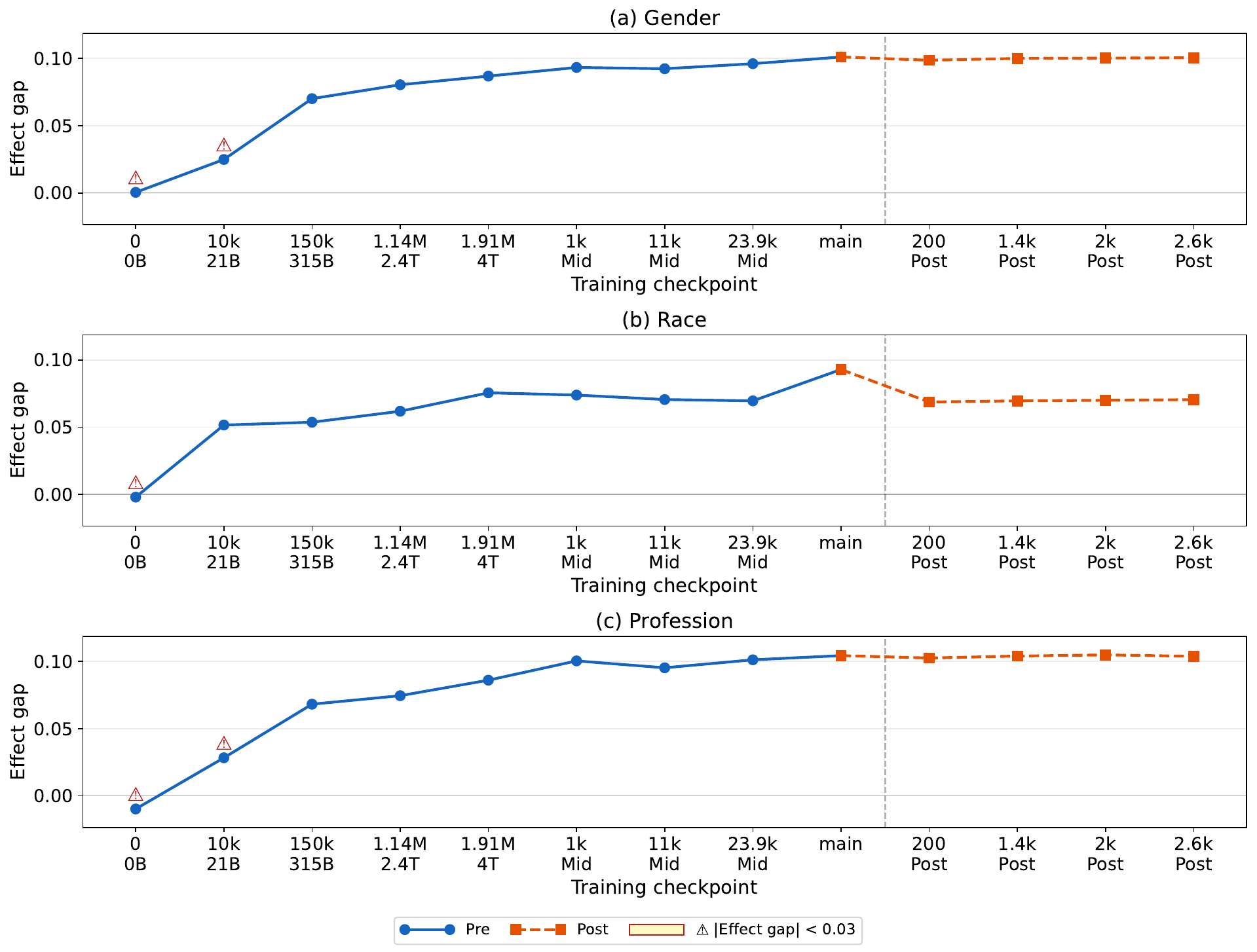}
  \caption{\textit{Effect Gap} of OLMo~2 1B across pre-training, mid-training on the third ingredient, the released \texttt{main} checkpoint, and post-training of OLMo-2-0425-1B-Instruct, for (a) gender, (b) race, and (c) profession. Ticks give the step and, below it, the tokens seen or the training stage, and the dashed line separates pre-training from post-training.}
  \label{fig:rq3_olmo_ckpt}
\end{figure*}

\begin{figure*}[t]
  \centering
  \includegraphics[width=\textwidth,height=0.78\textheight,keepaspectratio]{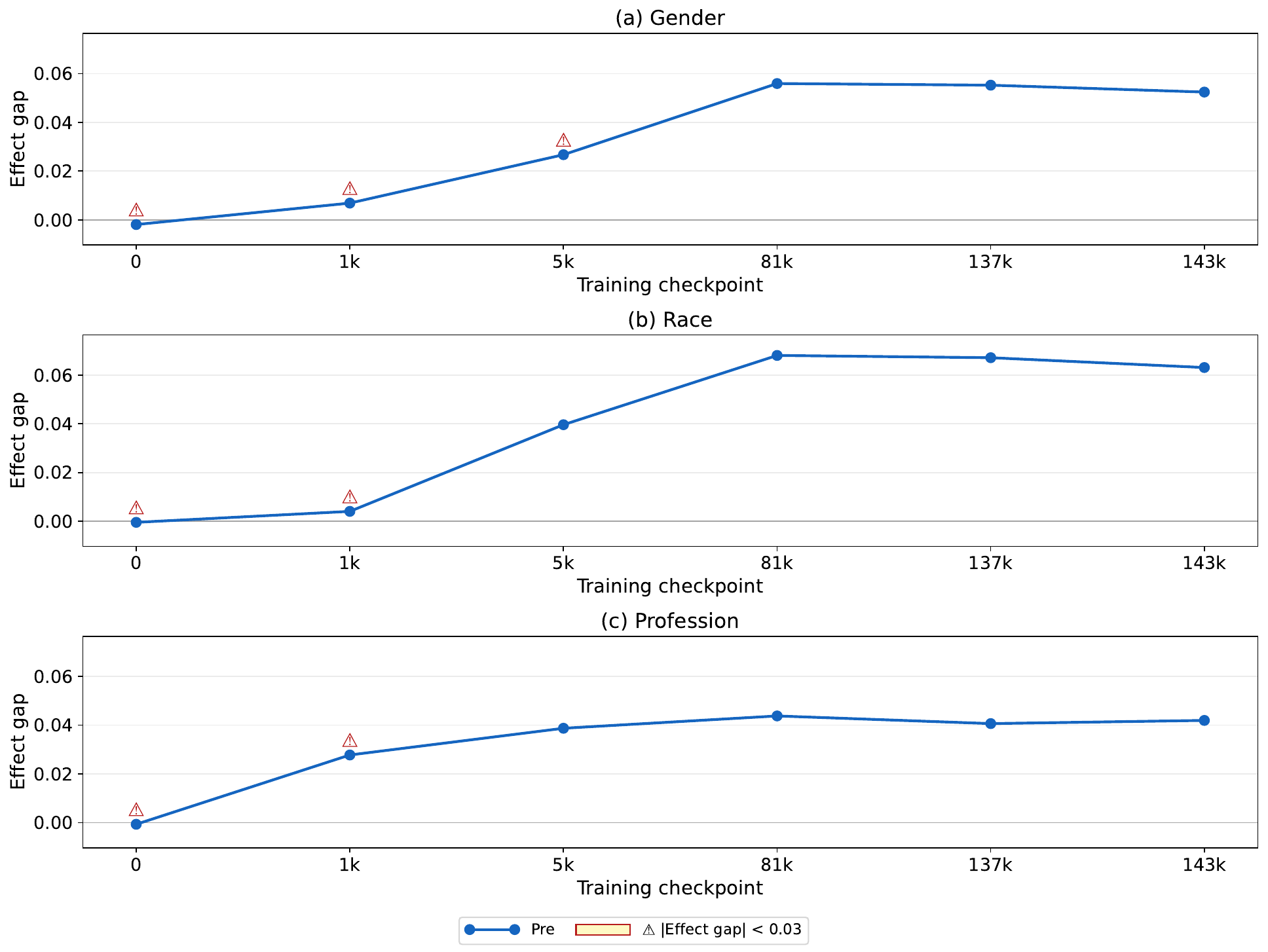}
  \caption{\textit{Effect Gap} of Pythia 1B across pre-training, at \texttt{step0}, \texttt{step1000}, \texttt{step5000}, \texttt{step81000}, \texttt{step137000} and \texttt{step143000} for (a) gender, (b) race, (c) profession. A warning sign marks an absolute effect gap below 0.03.}
  \label{fig:rq3_pythia_gap}
\end{figure*}

\begin{figure*}[p]
  \centering
  \includegraphics[width=\textwidth,height=0.78\textheight,keepaspectratio]{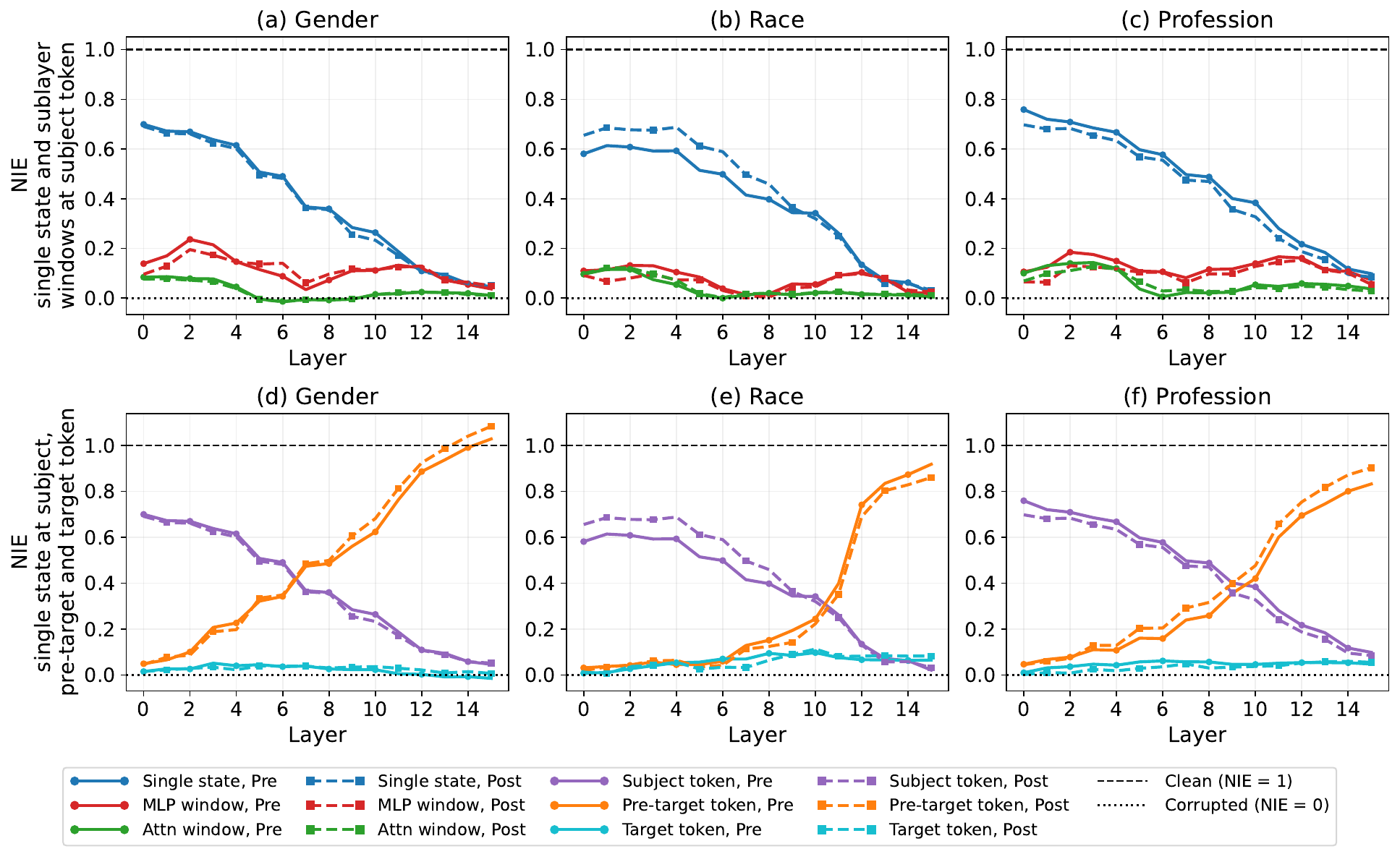}
  \caption{\textit{NIE} by layer for OLMo~2 1B, Pre (\texttt{OLMo-2-0425-1B} at its last mid-training checkpoint) and Post (\texttt{OLMo-2-0425-1B-Instruct} at \texttt{step\_2000}), Pre (solid) vs.\ Post (dashed). The upper row restores the single state, MLP window and Attn window at the subject token, and the lower row restores the single state at the subject, pre-target and target tokens.}
  \label{fig:rq3_olmo}
\end{figure*}

\begin{figure*}[p]
  \centering
  \includegraphics[width=\textwidth,height=0.78\textheight,keepaspectratio]{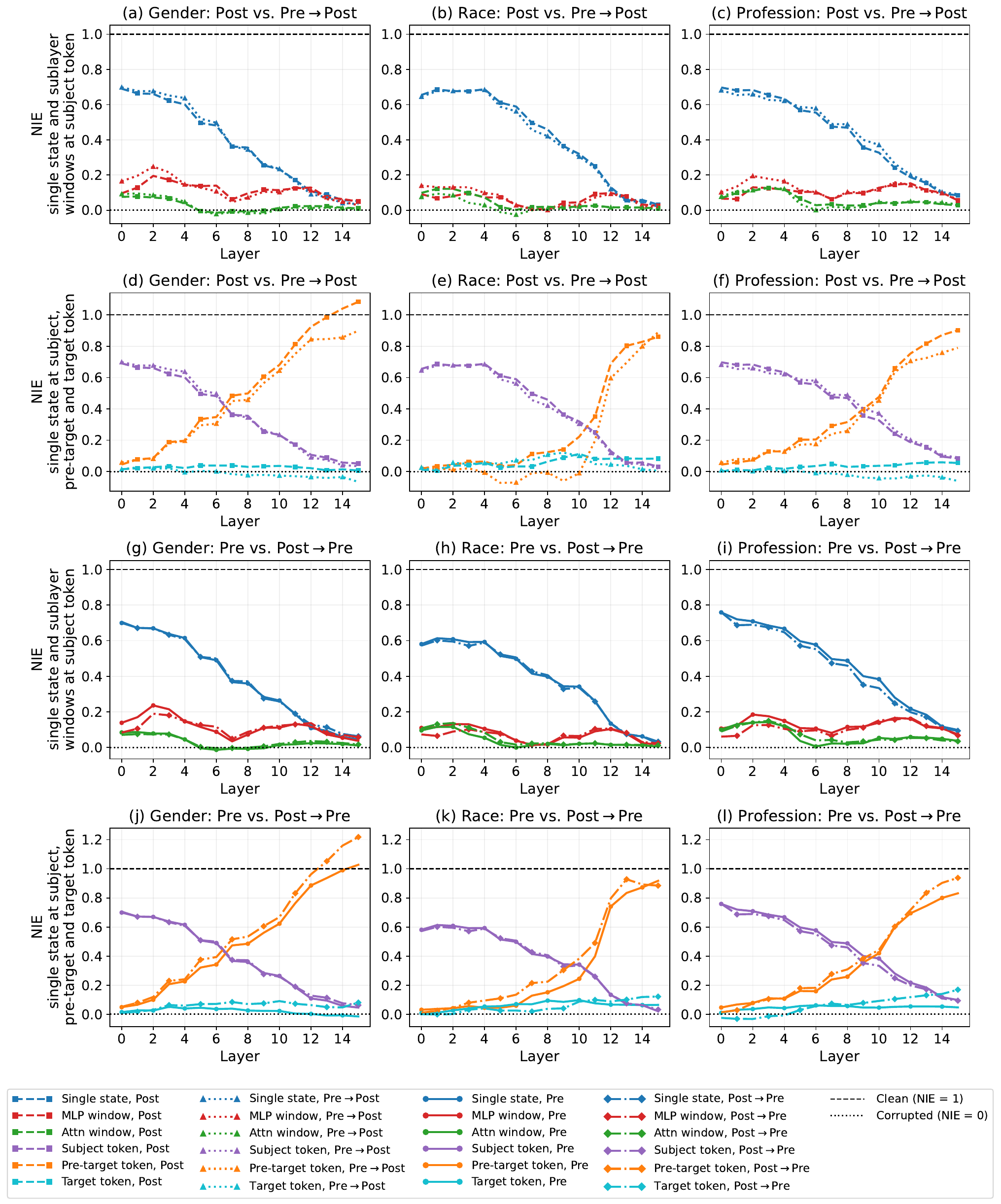}
  \caption{\textit{NIE} by layer for OLMo~2 1B under within-model and cross-model patching. Rows 1--2 patch the Post model and rows 3--4 the Pre model, each first restoring the single state, MLP window, and Attn window at the subject token, then the single state at the subject, pre-target, and target tokens.}
  \label{fig:rq3_olmo_cross}
\end{figure*}

\begin{figure*}[p]
  \centering
  \includegraphics[width=\textwidth,height=0.78\textheight,keepaspectratio]{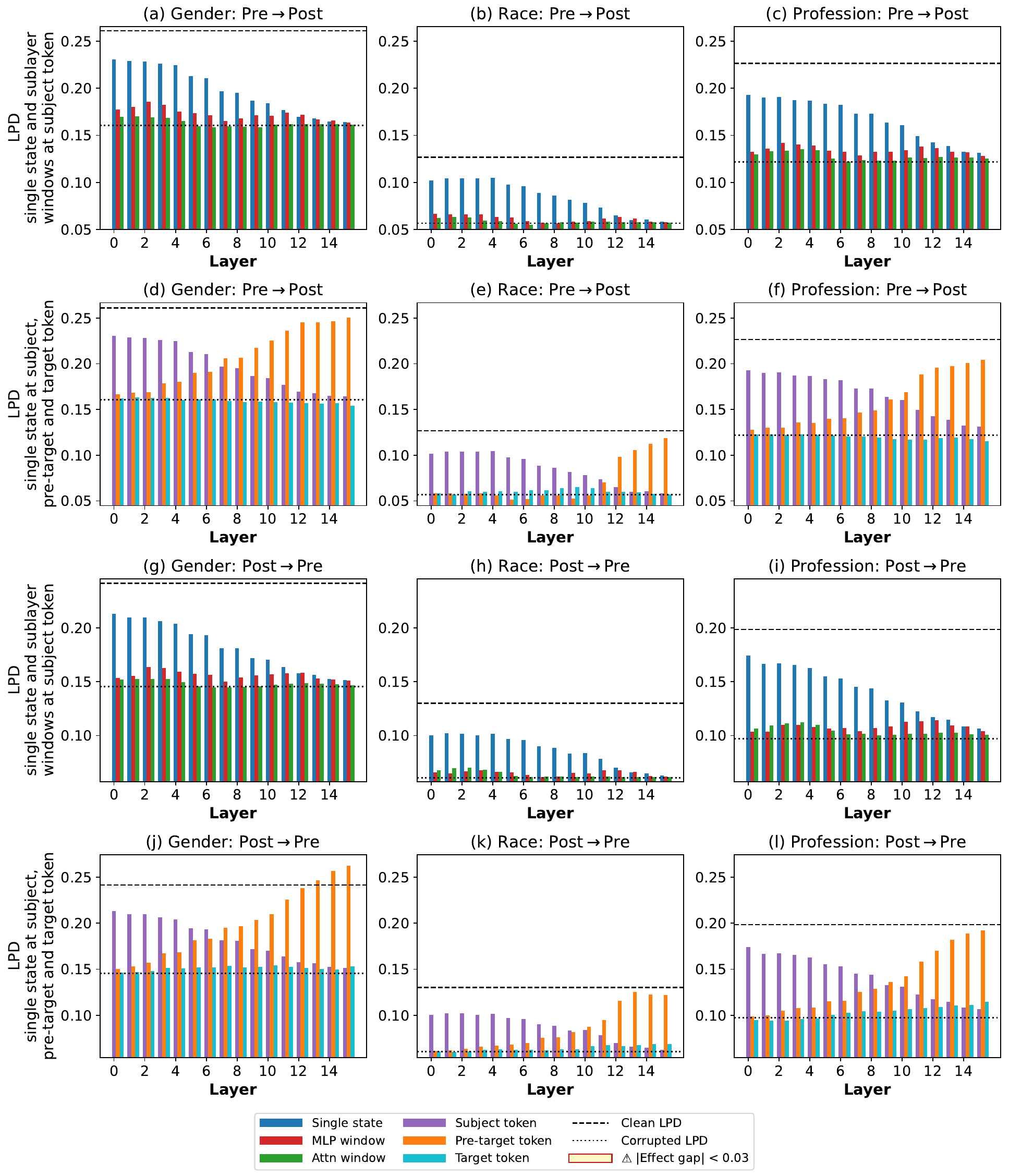}
  \caption{\textit{LPD} by layer for OLMo~2 1B under cross-model patching. Rows 1--2 show Pre$\to$Post and rows 3--4 Post$\to$Pre, each first restoring the single state, MLP window, and Attn window at the subject token, then the single state at the subject, pre-target, and target tokens.}
  \label{fig:rq3_olmo_crosslpd}
\end{figure*}

\begin{figure*}[p]
  \centering
  \includegraphics[width=\textwidth,height=0.78\textheight,keepaspectratio]{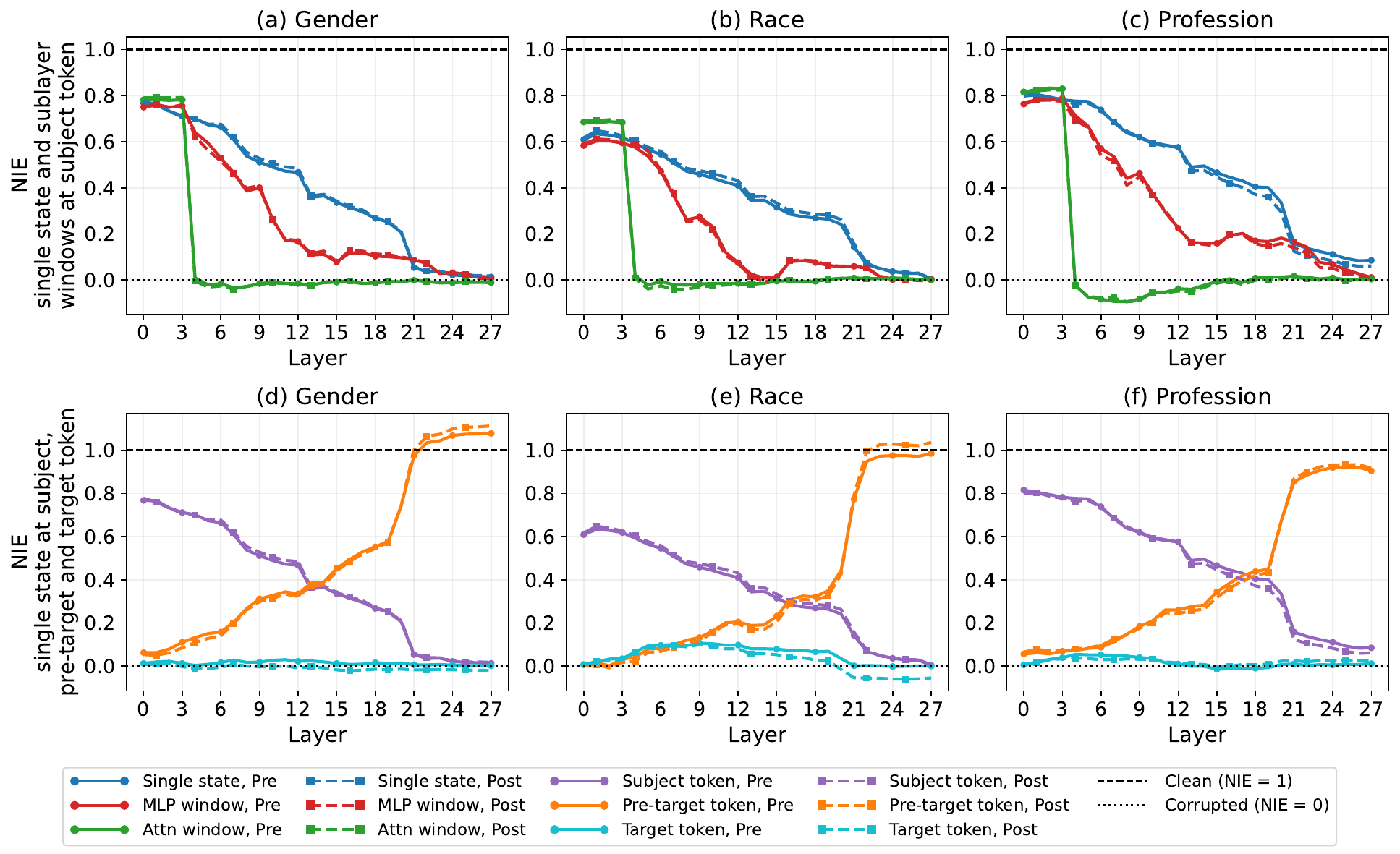}
  \caption{\textit{NIE} by layer for Qwen2.5 1.5B, Pre (\texttt{Qwen2.5-1.5B}) and Post (\texttt{Qwen2.5-1.5B-Instruct}), Pre (solid) vs.\ Post (dashed). The upper row restores the single state, MLP window, and Attn window at the subject token, and the lower row restores the single state at the subject, pre-target, and target tokens.}
  \label{fig:rq3_qwen}
\end{figure*}

\begin{figure*}[p]
  \centering
  \includegraphics[width=\textwidth,height=0.78\textheight,keepaspectratio]{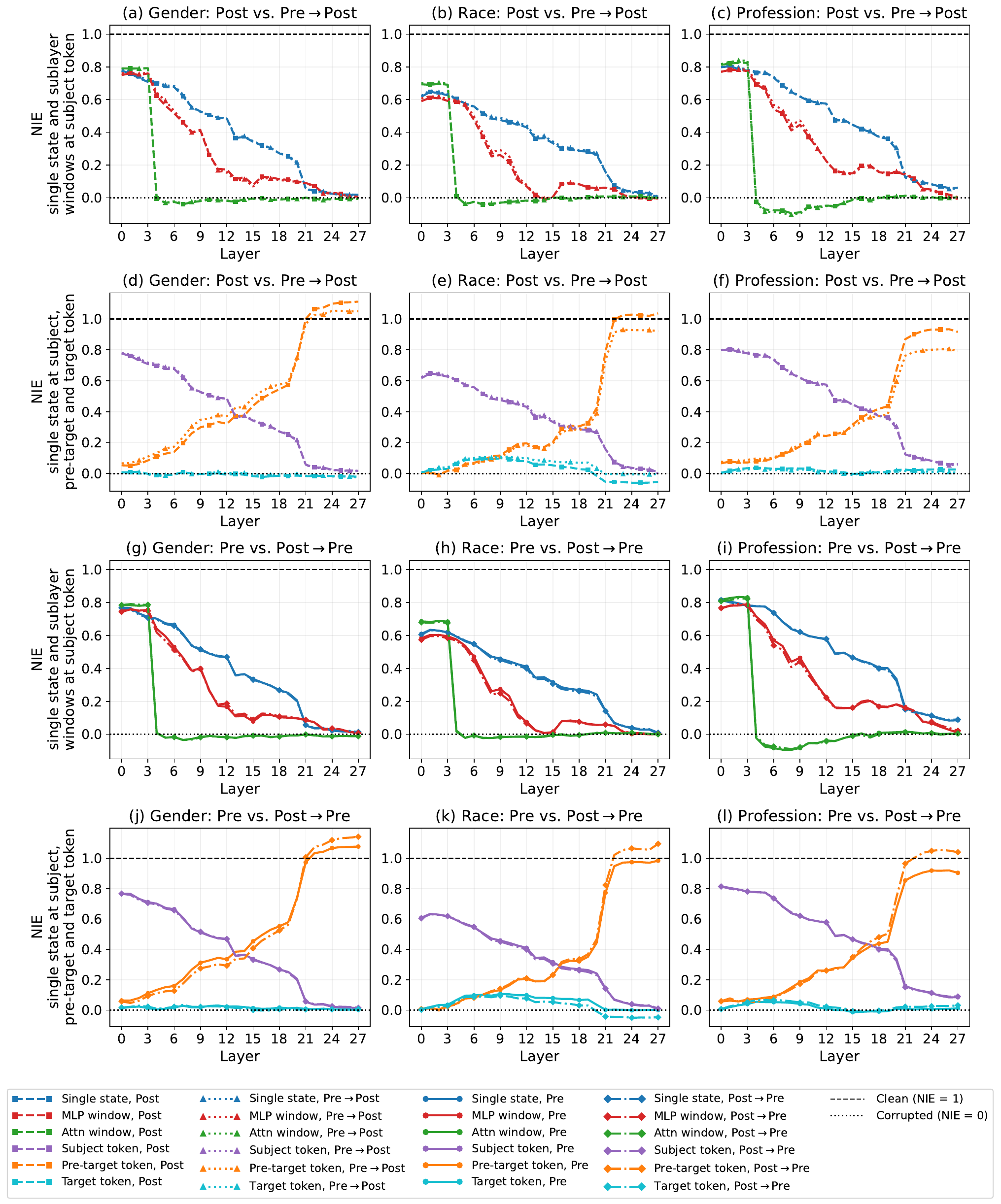}
  \caption{\textit{NIE} by layer for Qwen2.5 1.5B under within-model and cross-model patching. Rows 1--2 patch the Post model and rows 3--4 the Pre model, each first restoring the single state, MLP window and Attn window at the subject token, then the single state at the subject, pre-target and target tokens.}
  \label{fig:rq3_qwen_cross}
\end{figure*}

\begin{figure*}[p]
  \centering
  \includegraphics[width=\textwidth,height=0.78\textheight,keepaspectratio]{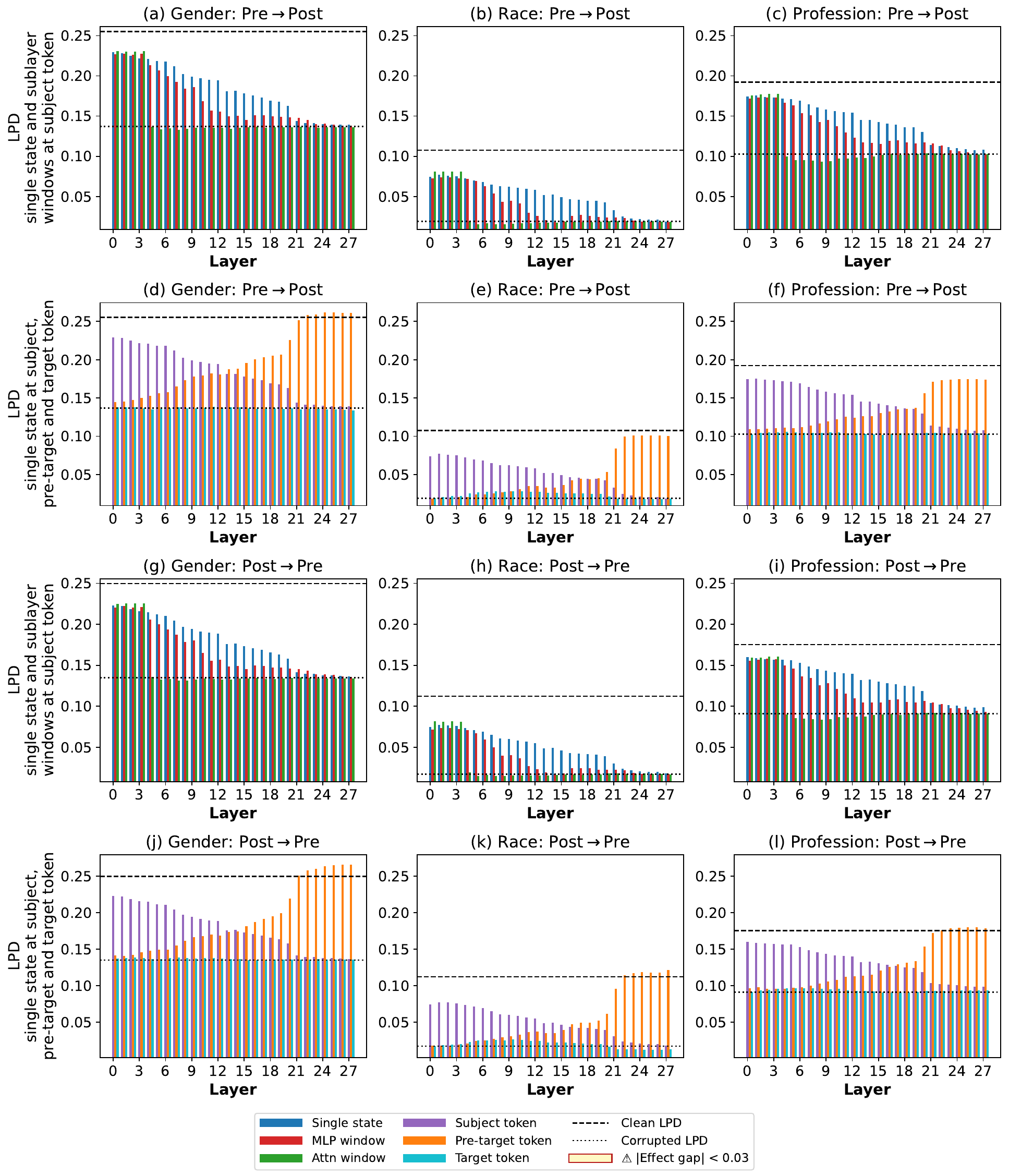}
  \caption{\textit{LPD} by layer for Qwen2.5 1.5B under cross-model patching. Rows 1--2 show Pre$\to$Post and rows 3--4 show Post$\to$Pre, each first restoring the single state, MLP window, and Attn window at the subject token, then the single state at the subject, pre-target, and target tokens.}
  \label{fig:rq3_qwen_crosslpd}
\end{figure*}

\begin{figure*}[p]
  \centering
  \includegraphics[width=\textwidth,height=0.78\textheight,keepaspectratio]{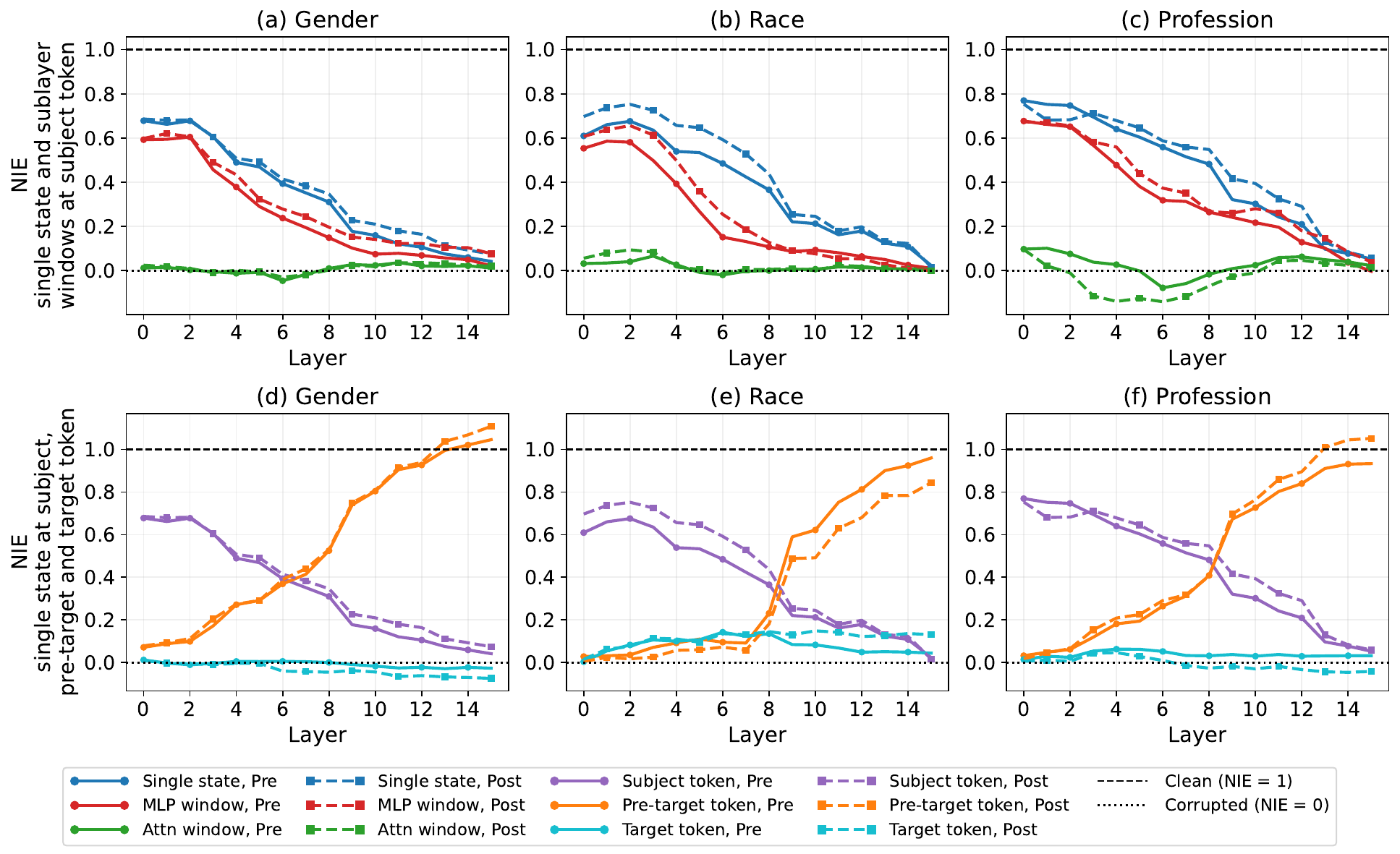}
  \caption{\textit{NIE} by layer for Llama~3.2 1B, Pre (\texttt{Llama-3.2-1B}) and Post (\texttt{Llama-3.2-1B-Instruct}), Pre (solid) vs.\ Post (dashed). The upper row restores the single state, MLP window, and Attn window at the subject token, and the lower row restores the single state at the subject, pre-target, and target tokens.}
  \label{fig:rq3_llama}
\end{figure*}

\begin{figure*}[p]
  \centering
  \includegraphics[width=\textwidth,height=0.78\textheight,keepaspectratio]{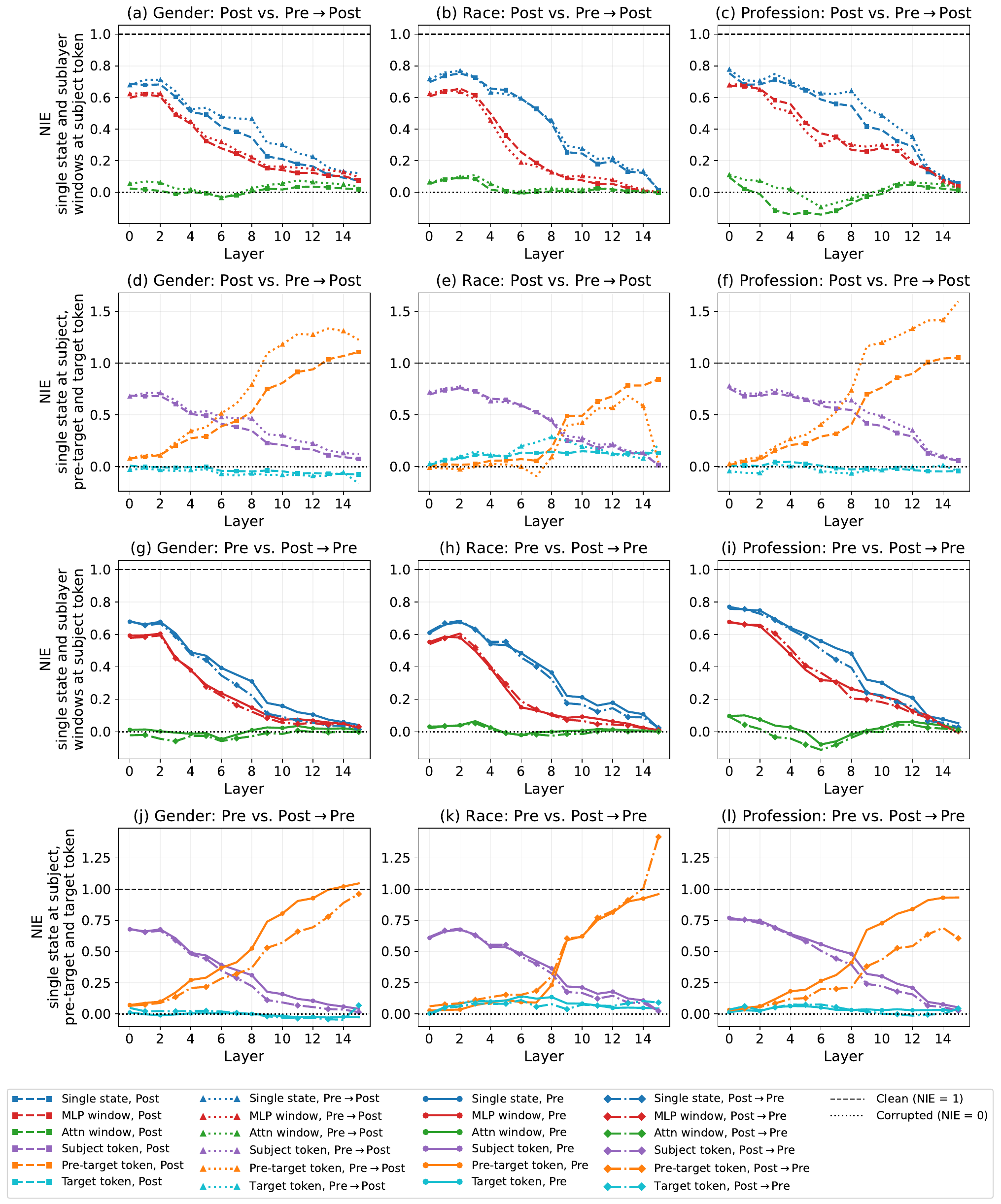}
  \caption{\textit{NIE} by layer for Llama~3.2 1B under within-model and cross-model patching. Rows 1--2 patch the Post model and rows 3--4 the Pre model, each first restoring the single state, MLP window, and Attn window at the subject token, then the single state at the subject, pre-target, and target tokens.}
  \label{fig:rq3_llama_cross}
\end{figure*}

\begin{figure*}[p]
  \centering
  \includegraphics[width=\textwidth,height=0.78\textheight,keepaspectratio]{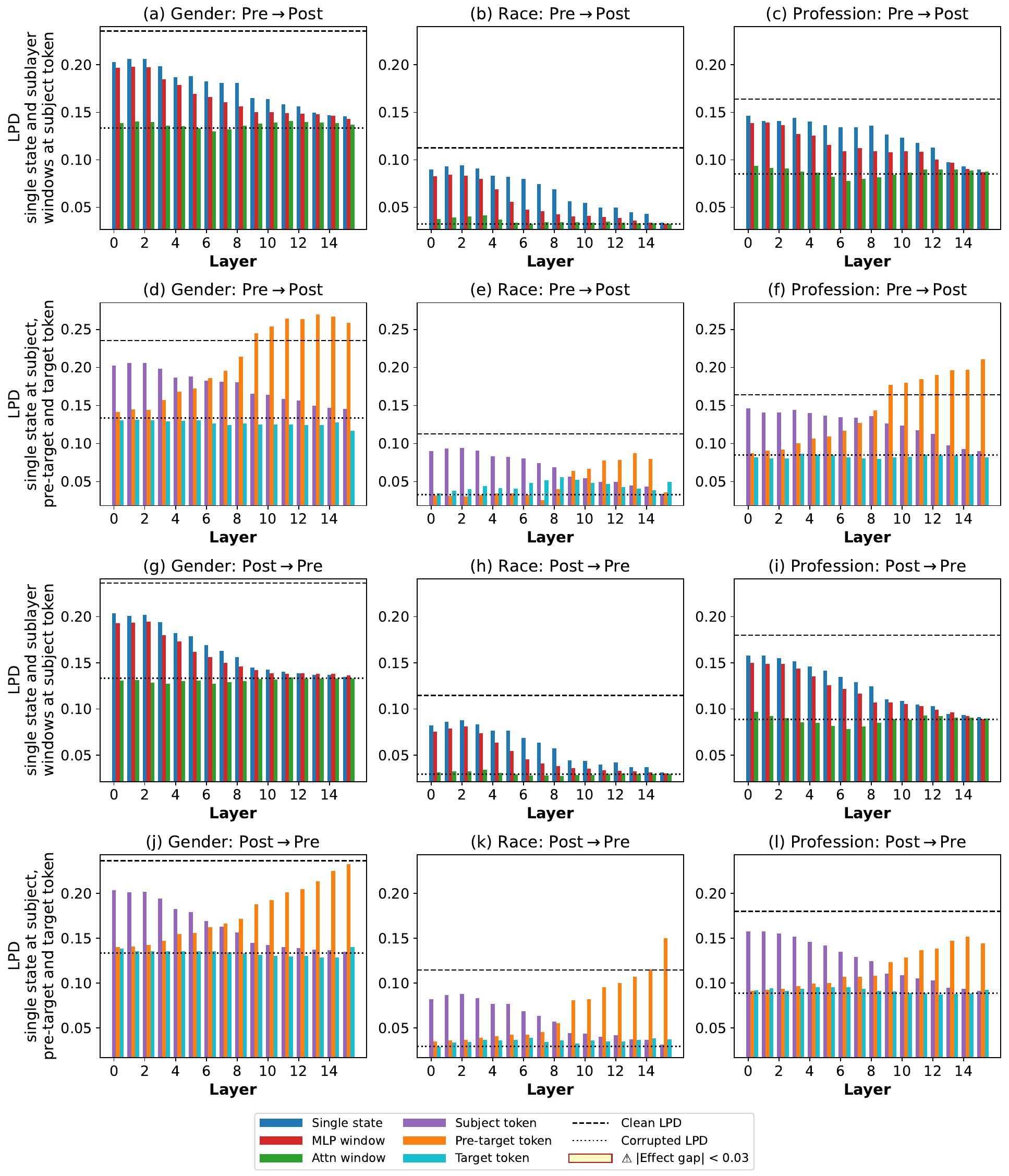}
  \caption{\textit{LPD} by layer for Llama~3.2 1B under cross-model patching. Rows 1--2 show Pre$\to$Post and rows 3--4 Post$\to$Pre, each first restoring the single state, MLP window, and Attn window at the subject token, then the single state at the subject, pre-target, and target tokens.}
  \label{fig:rq3_llama_crosslpd}
\end{figure*}

\begin{figure*}[p]
  \centering
  \includegraphics[width=\textwidth,height=0.78\textheight,keepaspectratio]{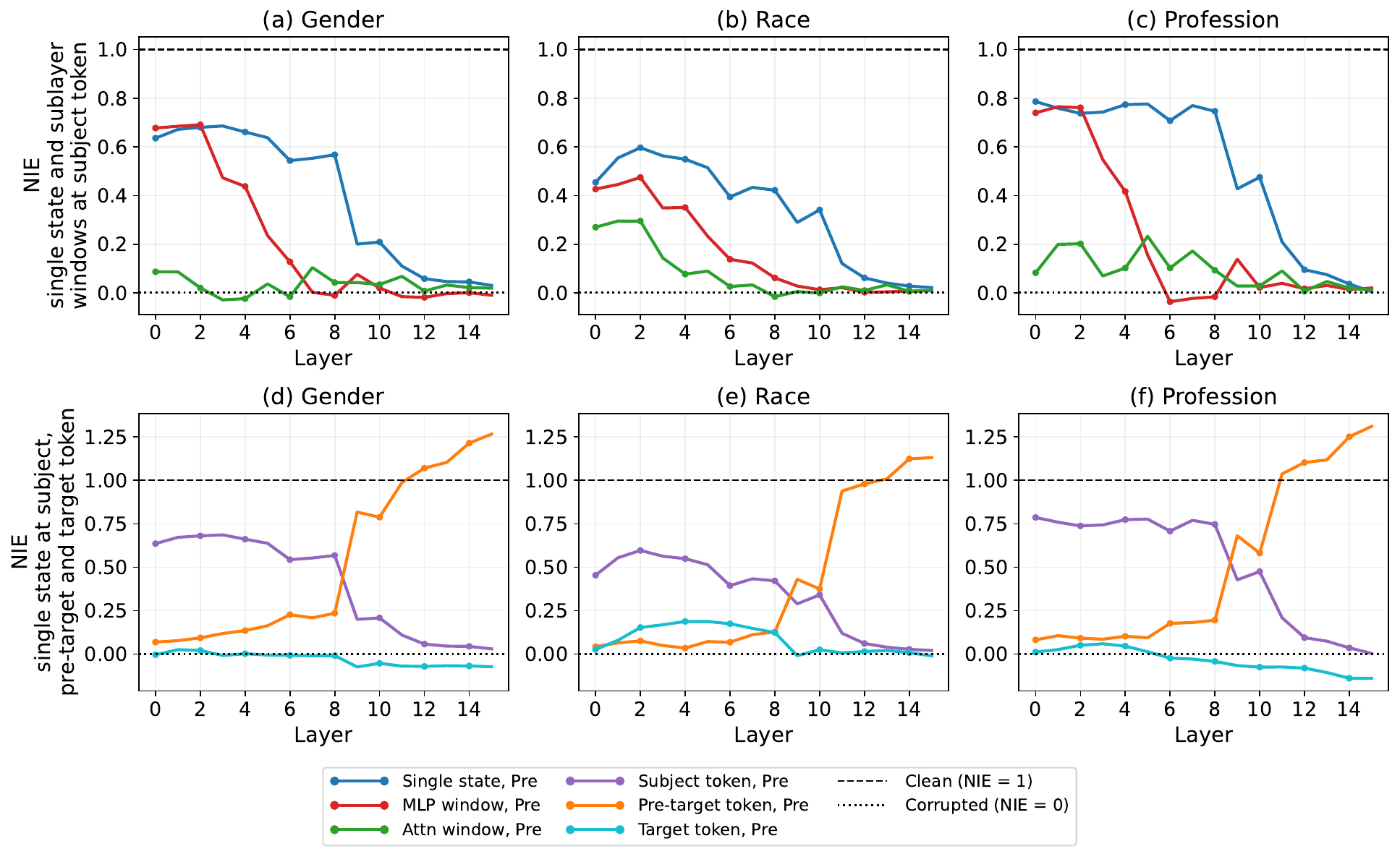}
  \caption{\textit{NIE} by layer for Pythia 1B (\texttt{pythia-1b} at \texttt{step143000}), which has no post-trained model. The upper row restores the single state, MLP window, and Attn window at the subject token, and the lower row restores the single state at the subject, pre-target, and target tokens.}
  \label{fig:rq3_pythia_nie}
\end{figure*}

\paragraph{Post-training Effects}
We present the \textit{NIE} curves of the three token positions for OLMo~2 1B, Qwen2.5 1.5B, Llama~3.2 1B and Pythia 1B in panels (d)--(f) of Figures~\ref{fig:rq3_olmo}, \ref{fig:rq3_qwen}, \ref{fig:rq3_llama} and~\ref{fig:rq3_pythia_nie} respectively.
As shown in Figures \ref{fig:rq3_olmo}, \ref{fig:rq3_qwen} and~\ref{fig:rq3_llama}, the Pre and Post models exhibit similar layer-wise \textit{NIE} patterns across token positions.
Across all nine bias type combinations, the maximum absolute difference between the Pre and Post \textit{NIE} curves is 0.17 (Table \ref{tab:rq3_shape}).
The 95\% bootstrap confidence intervals for the changes in subject \textit{NIE} at Layer 0 include zero in all cases except race bias in OLMo 2 and Llama 3.2, where the increases in \textit{NIE} are below 0.1.
These results suggest that post-training largely preserves the causal locations of LLM bias.
Meanwhile, as shown in Figure~\ref{fig:ct_domains} and detailed in Table \ref{tab:ct_posttraining}, post-training introduces relatively small changes in clean \textit{LPD}, with varying directions and magnitudes across model families and bias types. 
Table \ref{tab:ct_posttraining} shows the changes after post-training.
As shown in Figures~\ref{fig:rq3_olmo_cross}, \ref{fig:rq3_qwen_cross} and~\ref{fig:rq3_llama_cross}, cross-model patching yields similar layer-wise \textit{NIE} patterns to within-model patching across all three model families and in both patching directions.
The corresponding \textit{LPD} curves are presented in Figures~\ref{fig:rq3_olmo_crosslpd}, \ref{fig:rq3_qwen_crosslpd} and~\ref{fig:rq3_llama_crosslpd}
At the subject position in Layer 0, the absolute difference in \textit{NIE} between cross-model and within-model patching is at most 0.024, with all 95\% bootstrap confidence intervals for the differences including zero (Table \ref{tab:rq3_cross}). 
Figures~\ref{fig:rq3_olmo_ckpt} and~\ref{fig:rq3_pythia_gap} show the
\textit{Effect Gap} across training for OLMo~2 1B and Pythia 1B,
respectively. We report \textit{NIE}, the restored effect divided by
the \textit{Effect Gap}, only when the absolute \textit{Effect Gap} is
at least 0.03, since ratios with near-zero denominators are unreliable.
A warning sign marks every checkpoint with an absolute \textit{Effect
Gap} below 0.03.

\begin{table*}[t]
\centering
\small
\begin{tabular}{llcc}
\toprule
\textbf{Model} & \textbf{Bias Type} & \textbf{Subject \textit{NIE} at Layer 0, Post$\,-\,$Pre} & \textbf{Largest $|$\textit{NIE} difference$|$} \\
\midrule
OLMo~2 1B & Gender & -0.01 [-0.07, 0.06] & 0.06 [0.05, 0.19] \\
 & Race & 0.07 [0.01, 0.15] & 0.10 [0.06, 0.21] \\
 & Profession & -0.06 [-0.13, 0.01] & 0.07 [0.06, 0.19] \\
\midrule
Qwen2.5 1.5B & Gender & 0.01 [-0.02, 0.03] & 0.03 [0.03, 0.08] \\
 & Race  & 0.01 [-0.01, 0.03] & 0.06 [0.04, 0.11] \\
 & Profession  & -0.02 [-0.05, 0.01] & 0.04 [0.03, 0.09] \\
\midrule
Llama~3.2 1B & Gender  & 0.01 [-0.06, 0.07] & 0.07 [0.06, 0.20] \\
 & Race  & 0.09 [0.02, 0.16] & 0.14 [0.11, 0.26] \\
 & Profession  & -0.02 [-0.10, 0.06] & 0.17 [0.11, 0.32] \\
\bottomrule
\end{tabular}
  \caption{Similarity of the \textit{NIE} curves of the Pre and Post models under within-model patching: the change in subject \textit{NIE} at Layer 0 from Pre to Post, and the largest absolute difference between their \textit{NIE} curves, with 95\% bootstrap intervals.}
  \label{tab:rq3_shape}
\end{table*}

\begin{table*}[t]
  \centering\small
  \begin{tabular}{llccc}
\toprule
& & \multicolumn{2}{c}{\textbf{Clean \textit{LPD}}} &  
\\ \cmidrule(lr){3-4}
\textbf{Model} & \textbf{Bias Type} & \textbf{Pre} & \textbf{Post} & \textbf{Change (\%) [CI]} \\
\midrule
OLMo~2 1B & Gender & 0.241 [0.207, 0.277] & 0.261 [0.224, 0.297] & +8.1\% [+1.3\%, +15.6\%]  \\
 & Race & 0.130 [0.100, 0.160] & 0.127 [0.091, 0.161] & -2.6\% [-14.0\%, +7.4\%]  \\
 & Profession & 0.198 [0.161, 0.236] & 0.226 [0.185, 0.269] & +14.1\% [+3.1\%, +26.8\%]  \\
\midrule
Qwen2.5 1.5B & Gender & 0.250 [0.213, 0.286] & 0.255 [0.217, 0.294] & +2.2\% [-0.6\%, +5.0\%]  \\
 & Race & 0.112 [0.080, 0.144] & 0.108 [0.076, 0.139] & -4.0\% [-9.1\%, +0.4\%]  \\
 & Profession & 0.175 [0.139, 0.212] & 0.192 [0.155, 0.230] & +9.7\% [+5.6\%, +14.4\%]  \\
\midrule
Llama~3.2 1B & Gender & 0.236 [0.203, 0.270] & 0.235 [0.198, 0.273] & -0.4\% [-7.9\%, +7.1\%]  \\
 & Race & 0.115 [0.084, 0.145] & 0.113 [0.083, 0.143] & -1.8\% [-13.6\%, +11.3\%]  \\
 & Profession & 0.180 [0.145, 0.216] & 0.164 [0.128, 0.201] & -8.9\% [-19.6\%, +2.1\%]  \\
\midrule
Pythia 1B & Gender & 0.180 [0.146, 0.216] & -- & --  \\
 & Race & 0.111 [0.084, 0.140] & -- & -- \\
 & Profession & 0.140 [0.104, 0.176] & -- & --  \\
\bottomrule
\end{tabular}
  \caption{Clean \textit{LPD} of the Pre and Post model for each model family and bias type, and its change from Pre to Post relative to the Pre value with 95\% bootstrap intervals.}
  \label{tab:ct_posttraining}
\end{table*}

\begin{table*}[t]
  \centering\small
  \setlength{\tabcolsep}{4pt}
  \begin{tabular}{llllc}
\toprule
\textbf{Model} & \textbf{Cross-model} & \textbf{Within-model} & \textbf{Bias Type} & \textbf{NIE Difference (Cross $-$  Within)}  \\
\midrule
OLMo~2 1B & Pre$\to$Post & Post & Gender & 0.01 [-0.01, 0.03] \\
 &  &  & Race & -0.01 [-0.03, 0.01]  \\
 &  &  & Profession & -0.02 [-0.04, 0.00]  \\
 & Post$\to$Pre & Pre & Gender & 0.01 [-0.01, 0.03] \\
 &  &  & Race & -0.01 [-0.02, 0.00]  \\
 &  &  & Profession & 0.00 [-0.02, 0.02]  \\
\midrule
Qwen2.5 1.5B & Pre$\to$Post & Post & Gender & 0.00 [-0.00, 0.01]  \\
 &  &  & Race & 0.01 [-0.00, 0.01]  \\
 &  &  & Profession & 0.00 [-0.01, 0.01]  \\
 & Post$\to$Pre & Pre & Gender & -0.00 [-0.01, 0.01]\\
 &  &  & Race & -0.00 [-0.01, 0.00]  \\
 &  &  & Profession & -0.00 [-0.01, 0.01] \\
\midrule
Llama~3.2 1B & Pre$\to$Post & Post & Gender & -0.01 [-0.03, 0.02]  \\
 &  &  & Race & 0.02 [-0.01, 0.05]  \\
 &  &  & Profession & 0.02 [-0.01, 0.06]  \\
 & Post$\to$Pre & Pre & Gender & 0.00 [-0.02, 0.03]  \\
 &  &  & Race & 0.01 [-0.01, 0.03] \\
 &  &  & Profession & -0.01 [-0.04, 0.02] \\
\bottomrule
\end{tabular}
\caption{Differences in \textit{NIE} between cross-model and within-model patching at the subject position in Layer 0 across model families, bias types, and patching directions. Brackets indicate 95\% bootstrap confidence intervals.}
  \label{tab:rq3_cross}
\end{table*}

\begin{table*}[t]
  \centering\small
  \begin{tabular}{llcccc}
\toprule
& & & \multicolumn{3}{c}{Subject \textit{NIE} at Layer 0} \\ \cmidrule(lr){4-6}
\textbf{Model} & \textbf{Bias Type} & \textbf{\textit{Effect Gap}} & \textbf{Single State} & \textbf{MLP Window} & \textbf{Attn Window} \\
\midrule
OLMo~2 1B Pre & Gender & 0.096 [0.076, 0.115] & 0.70 [0.62, 0.77] & 0.14 [0.10, 0.18] & 0.08 [0.06, 0.11] \\
 & Race & 0.070 [0.055, 0.085] & 0.58 [0.52, 0.65] & 0.11 [0.08, 0.14] & 0.10 [0.07, 0.12] \\
 & Profession & 0.101 [0.080, 0.123] & 0.76 [0.69, 0.83] & 0.11 [0.07, 0.14] & 0.10 [0.07, 0.13] \\
\midrule
OLMo~2 1B Post & Gender & 0.100 [0.078, 0.123] & 0.69 [0.60, 0.78] & 0.10 [0.06, 0.14] & 0.08 [0.05, 0.11] \\
 & Race & 0.070 [0.052, 0.088] & 0.66 [0.57, 0.77] & 0.09 [0.06, 0.12] & 0.10 [0.07, 0.13] \\
 & Profession & 0.105 [0.080, 0.130] & 0.70 [0.61, 0.78] & 0.07 [0.03, 0.10] & 0.07 [0.04, 0.10] \\
\midrule
Qwen2.5 1.5B Pre & Gender & 0.115 [0.094, 0.135] & 0.77 [0.70, 0.83] & 0.75 [0.68, 0.82] & 0.78 [0.72, 0.85] \\
 & Race & 0.095 [0.079, 0.112] & 0.61 [0.55, 0.67] & 0.58 [0.53, 0.64] & 0.69 [0.64, 0.74] \\
 & Profession & 0.084 [0.062, 0.107] & 0.82 [0.74, 0.90] & 0.76 [0.68, 0.85] & 0.82 [0.74, 0.89] \\
\midrule
Qwen2.5 1.5B Post & Gender & 0.118 [0.097, 0.140] & 0.78 [0.71, 0.84] & 0.75 [0.68, 0.83] & 0.79 [0.73, 0.85] \\
 & Race & 0.089 [0.072, 0.106] & 0.62 [0.55, 0.68] & 0.59 [0.54, 0.65] & 0.69 [0.64, 0.75] \\
 & Profession & 0.090 [0.067, 0.113] & 0.80 [0.72, 0.87] & 0.77 [0.69, 0.85] & 0.81 [0.74, 0.89] \\
\midrule
Llama~3.2 1B Pre & Gender & 0.103 [0.083, 0.124] & 0.68 [0.60, 0.75] & 0.59 [0.51, 0.68] & 0.01 [-0.02, 0.04] \\
 & Race & 0.085 [0.068, 0.103] & 0.61 [0.54, 0.69] & 0.55 [0.48, 0.63] & 0.03 [0.01, 0.05] \\
 & Profession & 0.091 [0.070, 0.114] & 0.77 [0.68, 0.86] & 0.68 [0.59, 0.78] & 0.10 [0.05, 0.15] \\
\midrule
Llama~3.2 1B Post & Gender & 0.102 [0.080, 0.126] & 0.69 [0.60, 0.77] & 0.60 [0.51, 0.69] & 0.02 [-0.01, 0.06] \\
 & Race & 0.080 [0.063, 0.097] & 0.70 [0.62, 0.79] & 0.61 [0.53, 0.70] & 0.06 [0.03, 0.08] \\
 & Profession & 0.079 [0.056, 0.102] & 0.75 [0.65, 0.86] & 0.67 [0.58, 0.80] & 0.09 [0.04, 0.16] \\
\midrule
Pythia 1B & Gender & 0.052 [0.035, 0.070] & 0.64 [0.51, 0.76] & 0.68 [0.55, 0.81] & 0.09 [-0.01, 0.19] \\
 & Race & 0.063 [0.049, 0.077] & 0.45 [0.37, 0.54] & 0.43 [0.35, 0.50] & 0.27 [0.22, 0.33] \\
 & Profession & 0.042 [0.025, 0.059] & 0.79 [0.65, 0.95] & 0.74 [0.59, 0.93] & 0.08 [-0.06, 0.20] \\
\bottomrule
\end{tabular}
  \caption{\textit{Effect Gap} and \textit{NIE} under single-state, MLP, and attention patching at the subject position, with Layer 0 as the patched layer or window center. Brackets indicate 95\% bootstrap confidence intervals.}
  \label{tab:rq3_loc}
\end{table*}

\paragraph{Model Families.}
\label{app:rq3_sublayers}
The causal contributions to LLM bias vary across model families but exhibit similar patterns before and after post-training.
Figures~\ref{fig:rq3_olmo}, \ref{fig:rq3_qwen}, \ref{fig:rq3_llama} and~\ref{fig:rq3_pythia_nie} (panels (a)–(c)) present the layer-wise \textit{NIE} for single-state, MLP, and attention patching at the subject position, with detailed results at layer 0 in Table~\ref{tab:rq3_loc}.

\end{document}